\documentclass[preprint,12pt]{elsarticle}
\usepackage[top=1in,bottom=1in,left=1in,right=1in]{geometry}
\usepackage{amssymb}
\usepackage{amsmath,bm}
\usepackage{mathtools}
\usepackage{xcolor}
\usepackage{subfigure}
\usepackage{hyperref}
\usepackage{pdfpages}

\biboptions{sort&compress}

\begin{document}

\begin{frontmatter}

\title{Correcting model misspecification in physics-informed neural networks (PINNs)}

\author[brown]{Zongren Zou}
\author[hust]{Xuhui Meng\fnref{1}}
\author[brown]{George Em Karniadakis}

\fntext[1]{Corresponding author: xuhui\_meng@hust.edu.cn (Xuhui Meng).}
\address[brown]{Division of Applied Mathematics, Brown University, Providence, RI 02906, USA}
\address[hust]{Institute of Interdisciplinary Research for Mathematics and Applied Science, School of Mathematics and Statistics, Huazhong University of Science and Technology, Wuhan 430074, China}

\begin{abstract}
Data-driven discovery of governing equations in computational science has emerged as a new paradigm for obtaining accurate physical models and as a possible alternative to theoretical derivations. The recently developed physics-informed neural networks (PINNs) have also been employed to learn governing equations given data across diverse scientific disciplines, e.g., in biology and fluid dynamics. Despite the effectiveness of PINNs for discovering governing equations, the physical models encoded in PINNs may be misspecified in complex systems as some of the physical processes may not be fully understood, leading to the poor accuracy of PINN predictions. In this work, we present a general approach to correct the misspecified physical models in PINNs for discovering governing equations, given some sparse and/or noisy data. Specifically, we first encode the assumed physical models, which may be misspecified in PINNs, and then employ other deep neural networks (DNNs) to model the discrepancy between the imperfect models and the observational data. Due to the expressivity of DNNs, the proposed method is capable of reducing the computational errors caused by the model misspecification and thus enables the applications of PINNs in complex systems where the physical processes are not exactly known. Furthermore, we utilize the Bayesian physics-informed neural networks (B-PINNs) and/or ensemble PINNs to quantify uncertainties arising from noisy and/or gappy data in the discovered governing equations. A series of numerical examples including reaction-diffusion systems and non-Newtonian channel and cavity flows demonstrate that the added DNNs are capable of correcting the model misspecification in PINNs and thus reduce the discrepancy between the physical models encoded in PINNs and the observational data.  In addition, the B-PINNs and ensemble PINNs can provide reasonable uncertainty bounds in the discovered physical models, which makes the predictions more reliable.  We also demonstrate  that we can seamlessly combine the present approach with the symbolic regression to obtain the explicit governing equations upon the training of PINNs. We envision that the proposed approach will extend the applications of PINNs for discovering governing equations in problems where the physico-chemical or biological processes are not well understood.  
\end{abstract}

\begin{keyword}
model uncertainty\sep physics-informed neural networks \sep model misspecification \sep uncertainty quantification \sep non-Newtonian flows \sep symbolic regression
\end{keyword}

\end{frontmatter}

\section{Introduction}\label{sec:1}

Accurate governing equations or physical models are essential for understanding and quantifying physical processes across diverse scientific disciplines, such as the Navier-Stokes equations in fluid dynamics and the transport equations in geosciences. Generally, the governing equations are derived based on established theories such as the conservation of mass and momentum, thermodynamic laws, and so on.  However, there are also complex systems in which the governing equations are difficult to obtain theoretically due to the lack of understanding for certain physical processes, e.g.,  the constitutive relation in highly nonequilibrium flows \cite{zhang2020data}, the forcing caused by solar and volcanic variability in climate problems \cite{fyfe2021significant}, non-equilibrium reactions, etc. How to obtain analytical expressions that represent physical phenomena in nature and engineering remains an open question. 

With the rapid growth of available data and computing power, 
data-driven discovery of governing equations based on machine learning algorithms has emerged as a new paradigm for obtaining accurate physical models as the alternative to theoretical derivations \cite{schmidt2009distilling,brunton2016discovering,rudy2017data,chen2021physics,zhang2022gfinns,lee2022structure, gonzalez1998identification, rico1992discrete, zhang2018robust}. To name a few examples, Schmidt {\sl et al.} presented a symbolic regression approach to discover equations from experimental data \cite{schmidt2009distilling};  Brunton {\sl et al.} proposed the sparse identification of nonlinear dynamical systems (SINDy) to learn both ordinary and partial differential equations from data \cite{brunton2016discovering,rudy2017data}. In addition, Chen {\sl et al.} employed deep neural networks (DNNs), more specifically the physics-informed neural networks (PINNs), which encode the physical laws into DNNs via automatic differentiation \cite{raissi2019physics}, to discover governing equations from scarce data \cite{chen2021physics}.  
In most of the aforementioned approaches, a library of candidate terms that describe all the possible physical processes is required as prior knowledge \cite{schmidt2009distilling,brunton2016discovering,rudy2017data,chen2021physics}. 
However, mathematical expressions or models for certain physical processes are quite challenging to obtain in some complex real-world applications, e.g., chemical reactions in combustion, the constitutive relation in highly partially ionized plasmas, etc. It is probable that we miss some details of the physical process, hence misspecifying the physical models in such scenarios \cite{berger2020model}. An incomplete or non-comprehensive library leads to  the discrepancy between the observational data and the assumed governing equations. In other words, the discovered governing equations may not be accurate enough to describe the physical processes represented by the observational data. 

Several recent approaches have been developed to reduce the discrepancy between the observational data and the assumed physical models. For instance, Ebers {\sl et al.} \cite{ebers2022discrepancy} proposed to learn the {\emph{missing physics}} or the discrepancy between the imperfect models and the measurements using a plurality of methods, e.g. Gaussian process regression (GPR), DNNs, etc.; Chen {\sl et al.} also employed DNNs to recover the {\emph{missing physics}} in prior models \cite{chen2021generalized};  in \cite{zhang2023discovering, eastman2022pinn}, combinations of PINNs and symbolic regression were employed to identify the missing physics in biological problems, e.g. Alzheimer's disease; Saurabh {\sl et al.} \cite{malani2023some} and Zhu {\sl et al.} \cite{zhu2023implementation} discovered source terms of dynamic systems in a ``gray-box'' fashion with NN-based numerical integrators.
However, most of the existing work focuses on resolving the missing physics. Investigations on scenarios that we {\emph{misspecify the physical models}} are rare to the best of our knowledge. Further, the observational data from sensors in real-world applications are generally noisy, and may also be incomplete due to the limitation of data collection techniques. Such noisy and incomplete or gappy data lead to uncertainties in the predictions of machine learning algorithms or more specifically the scientific machine learning algorithms \cite{psaros2023uncertainty,zou2022neuraluq}. In addition, as widely discussed in the climate change modeling \cite{parker2013ensemble,gosling2012benefits,berger2020model}, the incomplete understanding of physical processes will also result in uncertainties in simulations.  We point out that quantifying uncertainties arsing from noisy/gappy data as well as the incomplete understanding of the physical processes is an important issue but has been largely ignored in the existing literature on data-driven discovery of governing equations.  In the current study, we refer to the uncertainties in the discovered physical models as {\emph{model uncertainty}}.

In the present work, we aim to address the issue of misspecifying physical models as well as quantifying model uncertainty in the data-driven discovery of governing equations. Specifically, PINNs are employed as the backbone to discover the governing equations or physical models from data due to their effectiveness as well as easy implementations. 
In particular, we first encode the assumed physical models that may be misspecified in PINNs, and add another DNN as the correction to the misspecified model to alleviate the discrepancy between the imperfect models and the observational data. We note that the library of candidate terms used in \cite{schmidt2009distilling,brunton2016discovering,rudy2017data,chen2021physics} is not required here since the added DNN can serve as a universal approximator to model the discrepancy. 
The Bayesian physics-informed neural networks (B-PINNs) \cite{yang2021b} and/or ensemble PINNs \cite{psaros2023uncertainty,zou2022neuraluq} are used to quantify the uncertainties in the discovered governing equations or physical models arising from the noisy and/or gappy data. Further, the symbolic regression is utilized to obtain the analytical expression for the unknown physics based on the trained DNNs that are used to correct the model misspecifications in PINNs.

The rest of this paper is organized as follows. In Sec. \ref{sec:2}, we introduce the model misspecification in PINNs as well as our approach to correct it, and also the methods for quantifying uncertainties in PINN predictions. In Sec. \ref{sec:3}, we conduct four numerical experiments to demonstrate the effectiveness of the proposed  approach in addressing the model misspecification issue in PINNs. We summarize the present work in Sec. \ref{sec:4}.

\section{Model misspecification in PINNs for learning governing equations}\label{sec:2}

\subsection{Learning governing equations from data using PINNs}

\begin{figure}[ht]
    \centering
    \includegraphics[scale=.6]{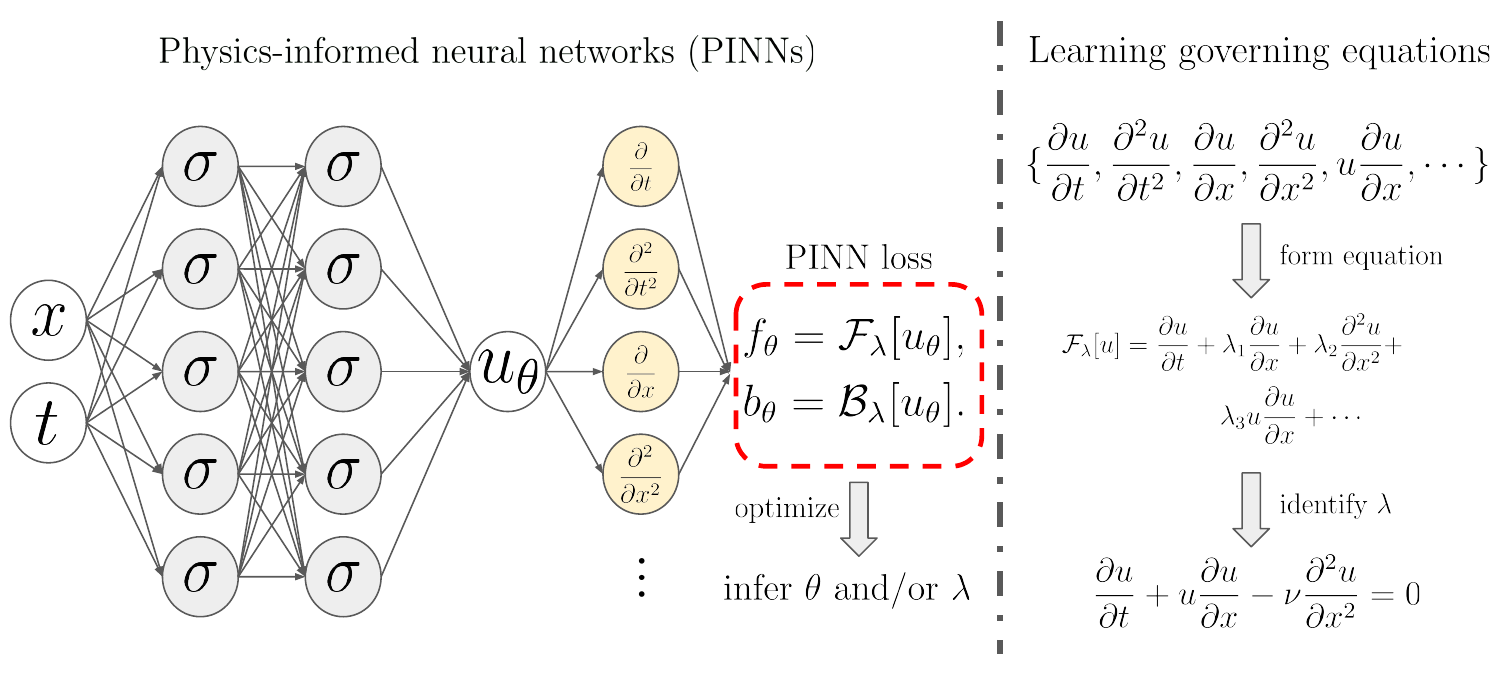}
    \caption{A schematic view of PINNs for learning governing equations from data. On the left, a general framework of PINNs is displayed where $\sigma$ denotes the activation function, $\theta$ denotes the DNN parameters, $u_\theta$ denotes the output of the DNN that is used to approximate $u$, and $f_\theta/b_\theta$ are computed via automatic differentiation \cite{raissi2019physics}. An example on learning governing equations with PINNs is illustrated on the right panel. Here a 1D Burgers equation is learned ($\lambda_1=0, \lambda_2=1, \lambda_3=-\nu$ where $\nu$ denotes the viscosity) given data on $u$. }
    \label{fig:pinn}
\end{figure}

Consider the following nonlinear ODE/PDE describing a physical system:
\begin{subequations}\label{eq:problem}
    \begin{align}
        \mathcal{F}_{\lambda}[u](x) &= f(x), x\in\Omega, \label{eq:problem_1}\\
        \mathcal{B}_{\lambda}[u](x) &= b(x), x\in \partial\Omega, \label{eq:problem_2}
    \end{align}
\end{subequations}
where $\Omega$ is the domain, $\lambda$ is the model parameter, $u$ is the sought solution, $f$ is the source term, $b$ is the boundary term, $\mathcal{F}_\lambda$ and $\mathcal{B}_\lambda$ are operators parameterized by $\lambda$, defining the equation and the boundary condition. 
The PINN method is capable of solving ODEs/PDEs given initial/boundary conditions as well as learning governing equations and identifying model parameters from data \cite{ yang2021b, zou2022neuraluq, lu2021deepxde, cai2021physics, pang2019fpinns, kharazmi2021hp, jagtap2021extended, jagtap2020conservative, zou2023hydra, linka2022bayesian, chen2023leveraging, lin2022multi, leung2022nh, tang2023pinns, tang2023adversarial, meng2021multi, sarabian2022physics, guo2022monte, gao2023failure, YIN2023105424}. A schematic view of the PINN method is presented in Fig.~\ref{fig:pinn}, in which a DNN with $\sigma$ as the activation function and $\theta$ as the parameter is built to approximate $u$ with $u_\theta$ and $f, b$ with $f_\theta, b_\theta$, respectively, via automatic differentiation \cite{raissi2019physics}. 

In the present study, we focus on using PINNs to learn governing equations from data. The dataset for learning governing equations with PINNs is denoted by $\mathcal{D}$, which can be expressed as $\mathcal{D} = \mathcal{D}_u \cup \mathcal{D}_f \cup \mathcal{D}_b$, where $\mathcal{D}_u = \{x^u_i, u_i\}_{i=1}^{N_u}$ is the set of data for $u$, and $\mathcal{D}_f = \{x^f_i, f_i\}_{i=1}^{N_f}$ and $\mathcal{D}_b = \{x^b_i, b_i\}_{i=1}^{N_b}$ are sets of data for the physics. The model parameter $\lambda$, which defines the equation, is then obtained by optimizing the following PINN loss function with the gradient-descent method and its variants \cite{ruder2016overview}:
\begin{equation}\label{eq:loss_PINN}
    \mathcal{L}(\theta) = w_u\mathcal{L}_{\mathcal{D}_u}(\theta) + \mathcal{L}_{PDE}(\theta),
\end{equation}
where 
\begin{equation}\nonumber
\begin{aligned}
    \mathcal{L}_{\mathcal{D}_u}(\theta) &= \frac{1}{N_u}\sum_{i=1}^{N_u} ||u_\theta(x^u_i) - u_i||_2^2,\\
    \mathcal{L}_{PDE}(\theta) &= \frac{w_f}{N_f}\sum_{i=1}^{N_u} ||\mathcal{F}[u_\theta](x^f_i) - f_i||_2^2 + \frac{w_b}{N_b}\sum_{i=1}^{N_b} ||\mathcal{B}[u_\theta](x^b_i) - b_i||_2^2,
\end{aligned}
\end{equation}
$w_u, w_f, w_b$ are belief weights for balancing different terms, and $||\cdot||_2$ is the $\ell^2$-norm for finite-dimensional vector. We note that 
in certain cases we may not have data corresponding to the boundary/initial conditions, i.e., the last term in $\mathcal{L}_{PDE}(\theta)$, and we can directly drop this term in the computations. 

\subsection{Correcting the model misspecification in PINNs}
\label{sec:mis_pinn}
\begin{figure}[ht!]
    \centering
    \subfigure[Uncertainty from physical model misspecifications.]
    {
     \includegraphics[scale=.6]{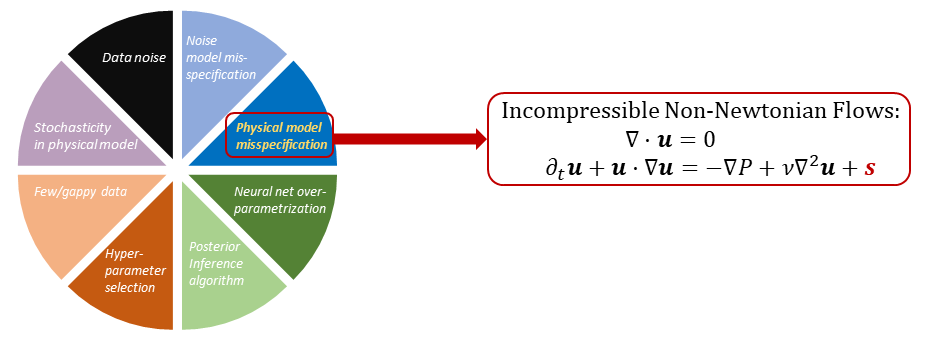}
    }
    \subfigure[When PINNs encounter model misspecification.]{
        \includegraphics[scale=.5]{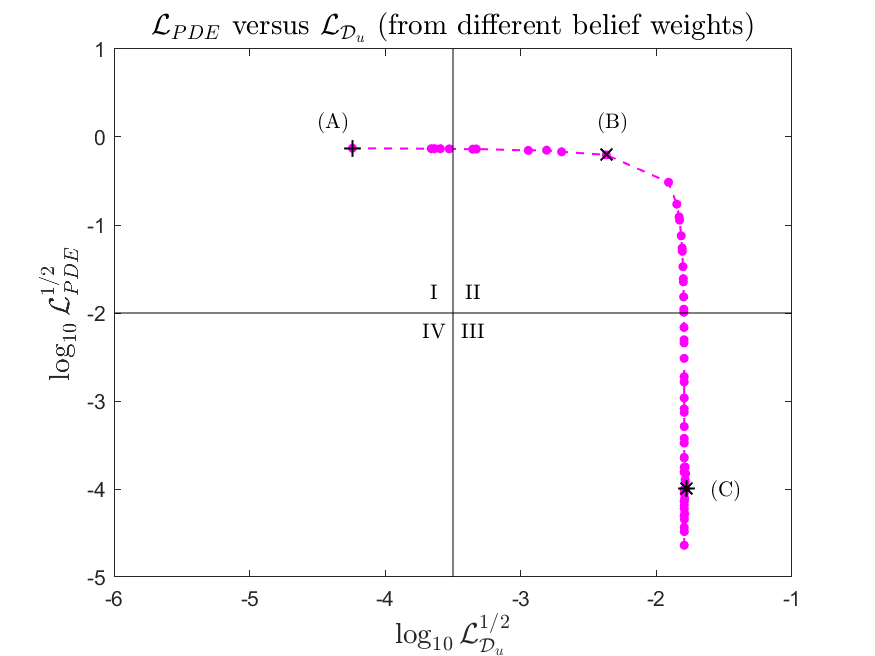}
        \includegraphics[scale=.27]{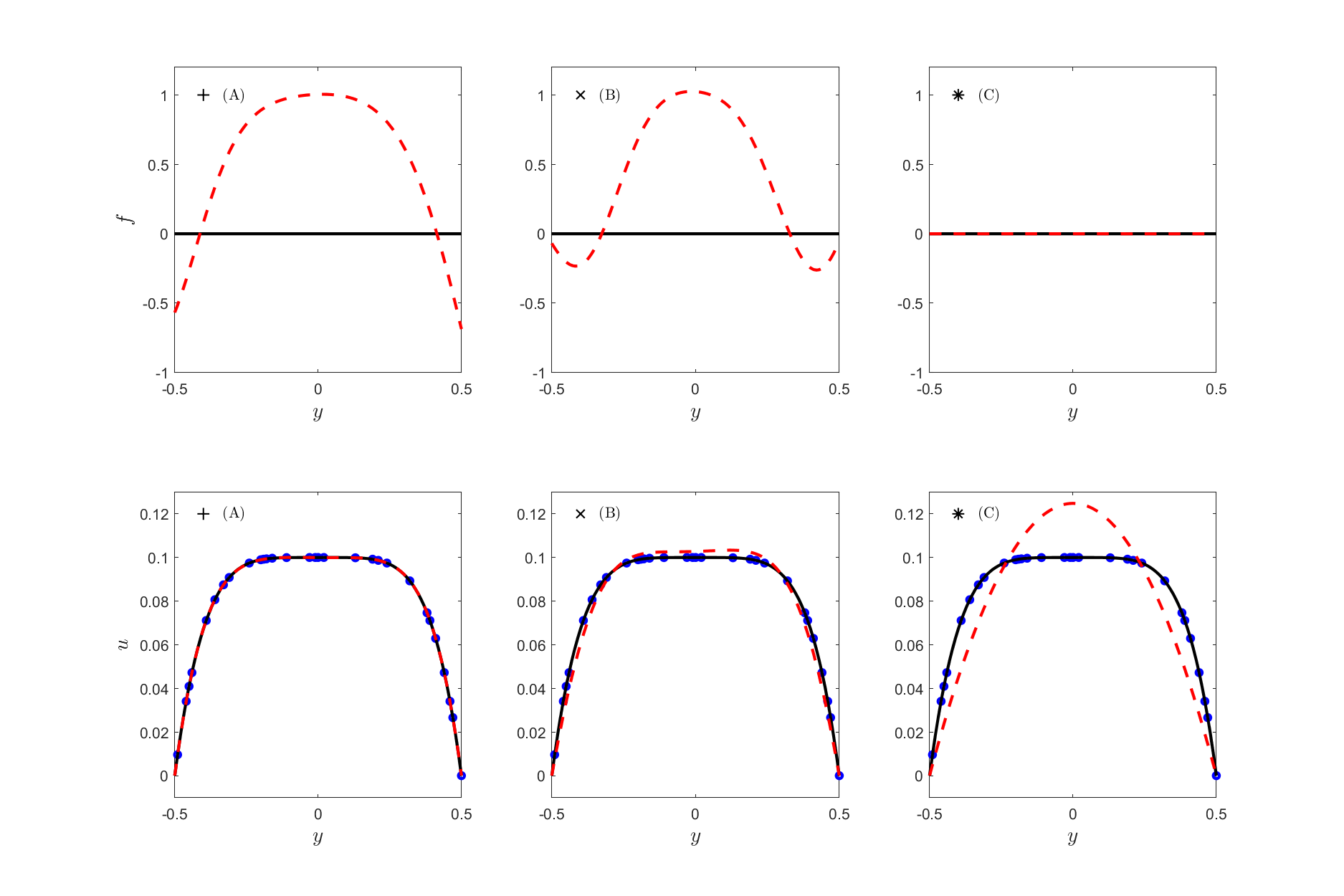}
    }
    \caption{
    In (a), a breakdown of total uncertainty in PINNs is presented. 
    In (b), results from PINNs encountering model misspecifications are displayed, where a physical model of Newtonian fluid is assumed but the real physics, which generates the data, is non-Newtonian. The PINN method is employed to identify the homogeneous viscosity with different belief weights in the loss function, i.e. pairs of $w_{PDE}$ and $w_{\mathcal{D}_u}$ in \eqref{eq:loss_PINN}, resulting in different pairs of PDE loss $\mathcal{L}_{PDE}$ and data loss $\mathcal{L}_{\mathcal{D}_u}$, shown in the left of (b). Due to the model misspecification, the PDE loss $\mathcal{L}_{PDE}$ and the data loss $\mathcal{L}_{\mathcal{D}_u}$ cannot be minimized simultaneously. Three results are presented in the right of (b), corresponding to three cases in the left, + (data are fitted well but the physics is not satisfied), * (the physics is satisfied but data are not fitted), and x (neither data nor physics are fitted). We note that the boundary condition is hard-encoded in the modeling. Details can be found in Sec. \ref{subsec:channel_flow}.
    }
    \label{fig:1}
\end{figure}

The computational accuracy of PINNs in learning governing equations strongly depends on the encoded physical models, i.e. $\mathcal{F}_\lambda$ and/or $\mathcal{B}_\lambda$ in Eq.~\eqref{eq:problem}. However, the correct specification of the physics is not always guaranteed, leading to significant discrepancy between the observational data and the assumed governing equations.
An example is demonstrated in Fig.~\ref{fig:1} where the flow is non-Newtonian but the physical model is misspecified as the one for Newtonian flows (see Sec. \ref{subsec:channel_flow} for details). 
As a result, the physics and the data cannot be fitted well at the same time: the training does not land on the region ``IV'', where both the losses for the data as well as the residual of the equations are close to zero, regardless of the choice of belief weights in the loss function Eq. \eqref{eq:loss_PINN}, i.e. $w_{f}$ for the physics and $w_u$ for the data. 
Three results from PINNs are sampled and displayed in Fig.~\ref{fig:1}(b): when $u_\theta$ fits the data (which come from a non-Newtonian flow), it does not satisfy the governing equations for Newtonian flows; when $u_\theta$ is forced to follow the Newtonian flow by setting large $w_{PDE}$, it presents Newtonian behavior and thus cannot fit the data from the non-Newtonian flow well.

\begin{figure}[ht]
    \centering
    \includegraphics[scale=.6]{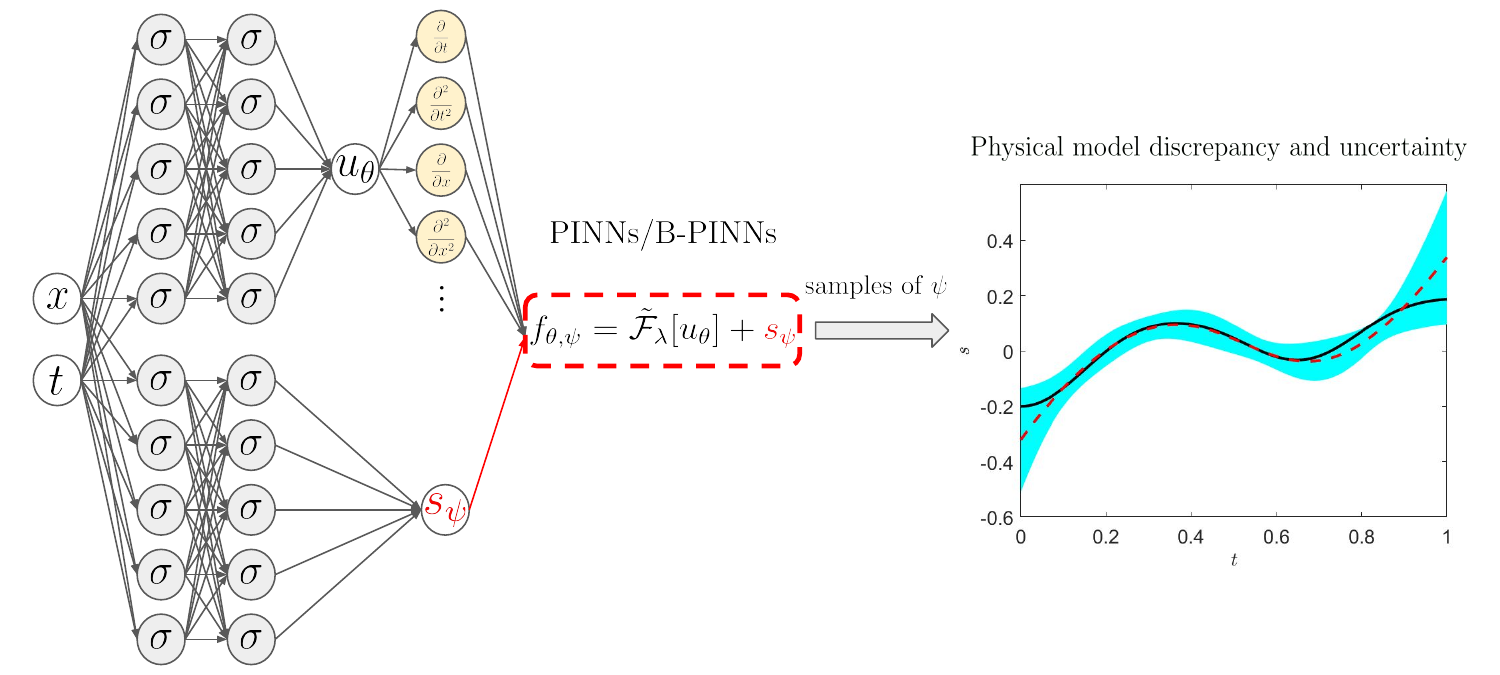}
    \caption{
    An overview of the proposed approach for correcting the misspecified physics. Here we assume that the nonlinear differential operator $\mathcal{F}$ is misspecified as $\tilde{\mathcal{F}}_\lambda$ such that $\tilde{\mathcal{F}}_\lambda[u]\neq f$, which leads to poor performance of PINNs in learning governing equations. Compared to the conventional PINN method shown in Fig.~\ref{fig:pinn}, an additional DNN, parameterized by $\psi$, is used to model the discrepancy. We note that $s_\psi$ serves two roles: (1) correcting the equation such that data of $u$ can be explained, and (2) modeling the discrepancy of the physics. With posterior samples of $\psi$, we obtain the prediction as well as the model uncertainty of the discrepancy.  We also note that DNNs in this figure, namely $u_\theta$ and $s_\psi$, are deterministic NNs in ensemble PINNs and Bayesian NNs (BNNs) in B-PINNs. 
    }
    \label{fig:method}
\end{figure}
The discrepancy between observational data and the assumed governing equations is caused by the physics/model misspecification. In this regard, we propose a general approach developed upon the original PINN framework to quantify the discrepancy as well as to correct the model misspecification. In this work, we focus on the misspecification of the differential operator $\mathcal{F}_\lambda$, but the proposed approach can be easily generalized to the boundary/initial operator. Here, we denote the misspecified differential operator by $\tilde{\mathcal{F}}_\lambda$ and then the discrepancy can be quantified as follows:
\begin{equation}
    s(x) = f(x) - \tilde{\mathcal{F}}_{\lambda}[u](x).
\end{equation}
Inspired by the PINN method, in which the sought solution $u$ is modeled with a DNN parameterized by $\theta$, we employ another DNN parameterized by $\psi$ to model $s(x)$, denoted as $s_\psi(x)$.
The source term $f(x)$ is approximated by $\tilde{\mathcal{F}}_\lambda[u_\theta](x) + s_\psi(x)$, and hence the term penalizing the violation of the physics in the loss function \eqref{eq:loss_PINN} is reformulated as:
\begin{equation}\label{eq:new_pde_loss}
    \mathcal{L}_{PDE}(\theta, \psi) = \frac{w_f}{N_f}\sum_{i=1}^{N_u} ||\tilde{\mathcal{F}}_\lambda[u_\theta](x^f_i) + s_\psi(x^f_i) - f_i||_2^2 + \frac{w_b}{N_b}\sum_{i=1}^{N_b} ||\mathcal{B}_\lambda[u_\theta](x^b_i) - b_i||_2^2.
\end{equation}
We note that $s_\psi$ plays two important roles in our approach handling model misspecification: (1) it approximates the function quantifying the model discrepancy, i.e. $s$, and (2) it allows both the data term $\mathcal{L}_{\mathcal{D}_u}(\theta)$ and the PDE term $\mathcal{L}_{PDE}(\theta, \psi)$ to be minimized to very small values simultaneously in the training of PINNs such that data are fitted by $u_\theta$ and the (corrected) physics are satisfied. 

\subsection{Uncertainty quantification}
As mentioned in Sec. \ref{sec:mis_pinn}, a key step in the present work is to learn the correction term, i.e., $s_{\psi}$, that is used to fix the issue of model misspecifications in PINNs. Generally, the measurements in real-world applications are noisy and can also be incomplete, leading to the uncertainties in predictions from PINNs \cite{psaros2023uncertainty,zou2022neuraluq}. We refer to the uncertainties in $s_{\psi}$ as \emph{model uncertainty}. Here, two typical approaches are employed for quantifying the uncertainties in $s_{\psi}$, i.e., ensemble PINNs \cite{psaros2023uncertainty, zou2022neuraluq} and B-PINNs \cite{yang2021b}. The former is computationally efficient but has difficulties preventing overfitting when the data are noisy, while the latter is able to handle the noisy data well but is computationally more expensive. We therefore utilize the ensemble PINNs and B-PINNs for cases with clean and noisy data, respectively. 

We now briefly review the ensemble PINNs and B-PINNs. The ensemble PINN method follows the deep ensemble method, proposed in \cite{lakshminarayanan2017simple}, and employs multiple standard training of PINNs independently. That is, PINNs with the PDE loss term in Eq. \eqref{eq:new_pde_loss} are trained multiple times with random initialization of the DNN parameter. Each random initialization corresponds to one initial guess for minimizing the PINN loss function \eqref{eq:loss_PINN} and each training is finding one global minimum. As a result, ensemble PINNs can be considered as identifying different maximum a posteriori (MAP) estimates of the PINN loss function \cite{psaros2023uncertainty}. 
On the other hand, the B-PINN method focuses on estimating the posteriors for $\theta$ and $\psi$ based on the Bayes' rule as follows:
\begin{equation}\label{eq:posterior}
    p(\theta, \psi|\mathcal{D}) \propto p(\mathcal{D}|\theta)p(\theta).
\end{equation}
Here $p(\theta|\mathcal{D})$ is the density function for the posterior distribution, $p(\mathcal{D}|\theta)$ for the likelihood, and $p(\theta)$ for the prior. 
Given independent and identically distributed (i.i.d.) data, the density function for the likelihood distribution can be written as follows:
\begin{equation}
    p(\mathcal{D}|\theta) = p(\mathcal{D}_u|\theta) p(\mathcal{D}_f|\theta) p(\mathcal{D}_b|\theta) = \prod_{i=1}^{N_u} p(u_i|x^u_i, \theta)\prod_{i=1}^{N_f} p(f_i|x^f_i, \theta)\prod_{i=1}^{N_b} p(b_i|x^b_i, \theta).
\end{equation}
If we further assume that the data for $u$, $f$ and $b$ follow Gaussian distribution with mean zero and standard deviation, $\sigma_u$, $\sigma_f$ and $\sigma_b$, respectively, then correcting it follows that:
\begin{subequations}
    \begin{align}
        p(u_i|x^u_i, \theta) &= \frac{1}{\sqrt{2\pi}\sigma_u}\exp(-\frac{||u_\theta(x_i) - u_i||_2^2}{2\sigma_u^2}), \label{eq:likelihood_u}\\
        p(f_i|x^f_i, \theta) &= \frac{1}{\sqrt{2\pi}\sigma_f}\exp(-\frac{||\mathcal{F}_\lambda[u_\theta](x^f_i) + s_\psi(x^f_i) - f_i||_2^2}{2\sigma_f^2}), \label{eq:likelihood_f}\\
        p(b_i|x^b_i, \theta) &= \frac{1}{\sqrt{2\pi}\sigma_b}\exp(-\frac{||\mathcal{B}_\lambda[u_\theta](x^b_i) - b_i||_2^2}{2\sigma_b^2}),.\label{eq:likelihood_f}
    \end{align}
\end{subequations}
We note that $\sigma_u$, $\sigma_f$ and $\sigma_b$ are often treated as scales of the additive Gaussian noise in measuring data of $u$, $f$and $b$, respectively. Various methods, e.g. Markov Chain Monte Carlo (MCMC) and variational inference, have been proposed to tackle the posterior distribution \eqref{eq:posterior}. 
In this work, we choose to use Hamiltonian Monte Carlo (HMC) method to estimate the posterior distribution by sampling $\theta$ and $\psi$ from it. We also note that here we only consider the homogeneous noise, while more complicated noise models, e.g. the heteroscedastic noise \cite{psaros2023uncertainty}, are also compatible with the framework.

In both the ensemble PINN and B-PINN methods, we obtain samples of the NN parameters, from which statistics of the prediction are estimated. In general, the mean and the standard deviation are often chosen to represent the prediction and the uncertainty \cite{psaros2023uncertainty}, respectively. Specifically in this work, we obtain samples of $\theta$ and $\psi$ from ensemble PINNs and/or B-PINNs, denoted by $\{\theta_i\}_{i=1}^M$ and $\{\psi_i\}_{i=1}^M$, respectively, where $M$ denotes the number of samples. 
Then, the predicted means of the sought solution $u$ and the discrepancy $s$ are estimated as follows:
\begin{equation}
    \hat{u}(x) \approx \frac{1}{M}\sum_{i=1}^M u_{\theta_i}(x), ~\hat{s}(x) \approx \frac{1}{M}\sum_{i=1}^M s_{\psi_i}(x),
\end{equation}
where $\hat{u}$ and $\hat{s}$ denote the mean of $u$ and $s$, respectively.
Similarly, the predicted uncertainties are:
\begin{equation}
    Var[u](x) \approx \frac{1}{M}\sum_{i=1}^M (u_{\theta_i}(x) - \hat{u}(x))^2, ~Var[s](x) \approx \frac{1}{M}\sum_{i=1}^M (s_{\psi_i}(x) - \hat{s}(x))^2,
\end{equation}
where $Var[u]$ and $Var[s]$ denote the variance of $u$ and $s$, respectively.
Interested readers are directed to \cite{psaros2023uncertainty} for a review and \cite{zou2022neuraluq} for a comprehensive Python library termed NeuralUQ.

We note that upon the computation of $s_\psi(x)$, it can be used in various way: (1) its value represents the model discrepancy and its uncetainty represents the uncertainty in the physical model which we refer to as \emph{model uncertainty}; (2) it corrects the misspecified physical models and identifies a system which explains the data; and (3) it could also be used to obtain mathematical expressions for the missing details of certain physical process via the symbolic regression \cite{zhang2023discovering, eastman2022pinn, kiyani2023framework}.  Specifically in the present work, we employ the symbolic regression method developed in \cite{cranmer2023interpretable} and the associated Python library PySR to regress analytic mathematical expressions for the unknown  physical processes (See Sec. \ref{subsec:ode} for a demonstration example).

\section{Results and discussion}\label{sec:3}

In this section, we conduct four numerical experiments, i.e., (1) an ODE system, (2) a one-dimensional (1D) reaction-diffusion equation, (3) a two-dimensional (2D) channel flow, and (4) a 2D cavity flow. In the first two examples, reaction models are possibly misspecified, and in the last two examples, we assume that we misspecify the constitutive relation in non-Newtonian flows as Newtonian ones. In all the test cases, we employ DNNs to correct the misspecified models as discussed in Sec. \ref{sec:2}. Details for the computations (e.g., training steps, NN architectures, etc.) can be found  in \ref{subsec:hyperparameter}.

\subsection{Pedagogical example: An ODE system}\label{subsec:ode}
We first consider the following one-dimensional ODE:
\begin{equation}\label{eq:ode}
    \begin{split}
        &\frac{du}{dt} = f(t) + \lambda u (1 - u), ~ t \in [0, 1],\\
        & u(t = 0) = u_0,
    \end{split}
\end{equation}
where $\lambda$ is a non-negative constant, $u_0$ is the initial condition, and $f$ is the source term. In certain real-world applications, the reaction term, i.e., $\lambda u(1-u)$ in the right hand side of Eq.~\eqref{eq:ode}, may not be exactly known and herein it is misspecified as an empirical model. In this regard, we study the effect of different prior knowledge about the reaction model in the discovery of governing equation. In particular, the following cases are considered:
\begin{enumerate}
    \item Case (A): The reaction term is correctly specified as $\lambda u (1-u)$, where $\lambda\geq 0$ is unknown.
    \item Case (B): The reaction term is misspecified as $\lambda \cos(u)$, where $\lambda\geq 0$ is unknown.
    \item Case (C): The reaction term is misspecified as $\lambda \cos(u)$ (with known $\lambda=0.2$) but we add $s(t)$ to correct it, i.e., $\lambda \cos(u) + s(t)$. 
\end{enumerate}
We denote the reaction term as $\phi$, and the target here is to identify $\phi$ given data on $u/f$ as well as differently specified models. In Case (A) and (B), the problem degenerates to identifying the model parameter $\lambda$, and $\phi$ in these two cases can then be computed  by $\tilde{\lambda}u_\theta(1-u_\theta)$ and $\tilde{\lambda}\cos(u_\theta)$, respectively, where $\tilde{\lambda}$ is the PINN estimate of $\lambda$. In Case(C), $\phi$ is computed by $\tilde{\lambda} \cos(u_\theta) + s_\psi(t)$ where $s_\psi$ is the estimation of the discrepancy induced by the model misspecification. We note that in Case (C) we assume $\lambda$ is known and set $\lambda=0.2$ for simplicity. This is because the discrepancy is already modeled by $s(t)$ and hence treating $\lambda$ as unknown does not make any difference in correcting the misspecified model.

To reduce the effect of the noisy/gappy data on the predicted accuracy, 
we first test the case with clean and sufficient data. The training data are generated by solving Eq.~\eqref{eq:ode} with $f(t) = \sin(2\pi t)$ and $u_0=0$ using \textit{MATLAB ode45} \cite{MATLAB}. We assume that we have 101 data for $u$ and $f$, uniformly sampled from $t\in[0, 1]$, and we employ ensemble PINNs \cite{zou2022neuraluq, psaros2023uncertainty} to learn the governing equation and also quantify the uncertainties in predictions. 
In all the test cases here, we choose a DNN with two hidden layers, each of which has 50 neurons and is equipped with hyperbolic tangent as the activation function, to approximate $u$. The equations are then encoded in PINNs via the automatic differentiation \cite{raissi2019physics}. Further, we utilize an additional DNN to correct the model misspecification in Case (C), which has the same architecture as the one used for approximating $u$. 
The Adam optimizer \cite{kingma2014adam} is applied to train the ensemble PINNs.

\begin{figure}[ht!]
    \centering
    \subfigure[PINN with the correct physical model.]{
        \includegraphics[width=0.3\textwidth]{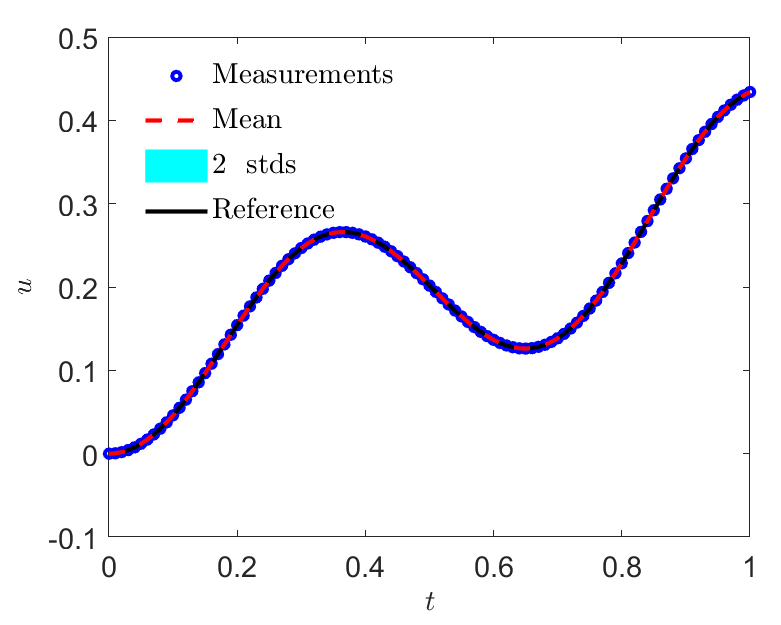}
        \includegraphics[width=0.3\textwidth]{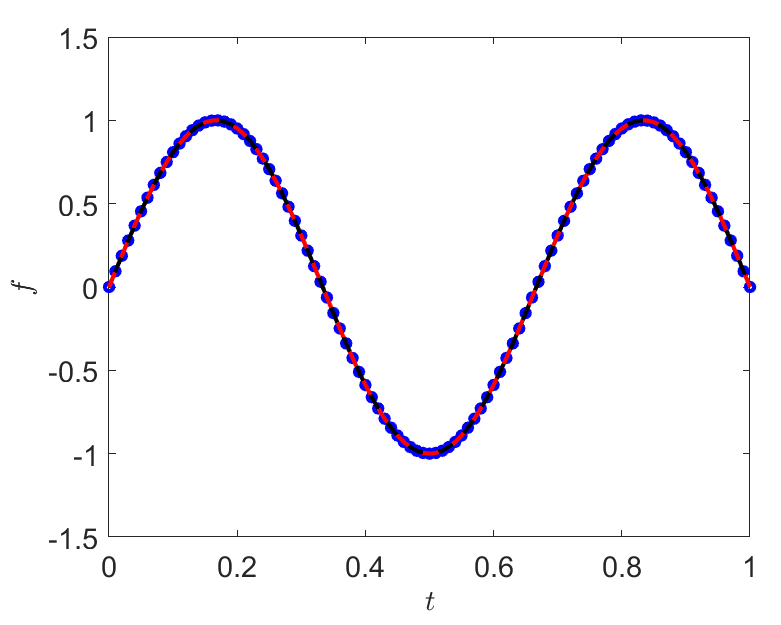} 
        \includegraphics[width=0.3\textwidth]{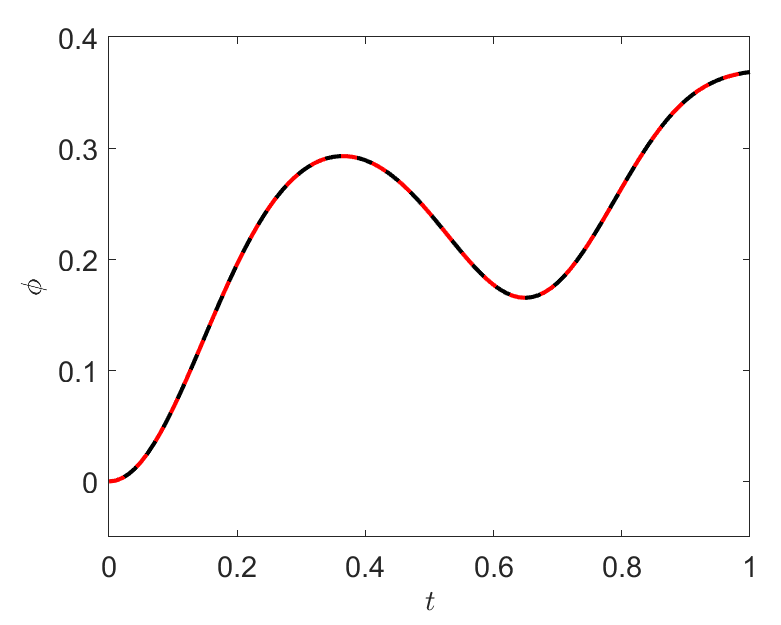}
    }
    \subfigure[PINN with a misspecified physical model.]{
        \includegraphics[width=0.3\textwidth]{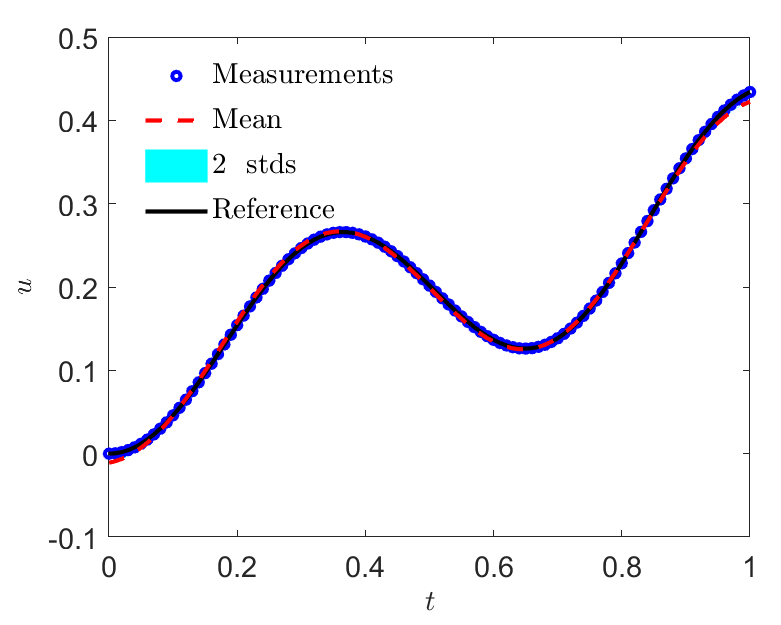}
        \includegraphics[width=0.3\textwidth]{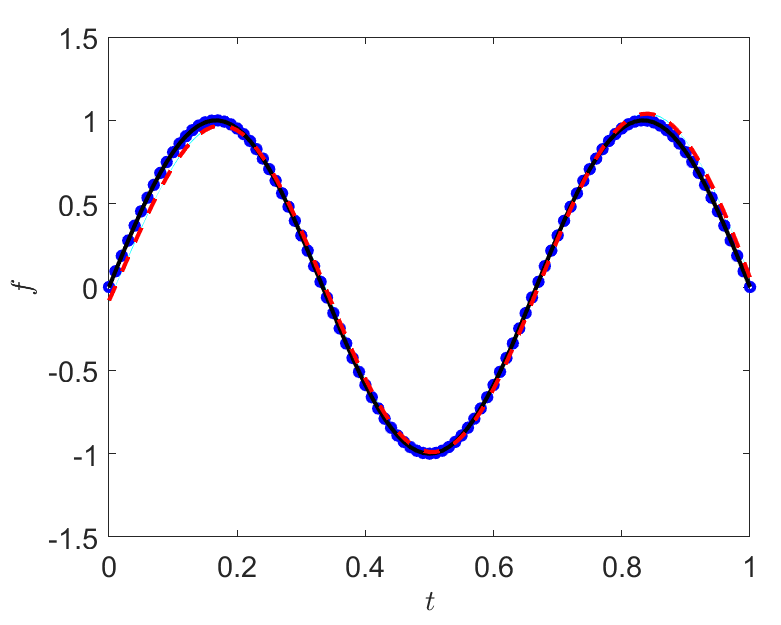}
        \includegraphics[width=0.3\textwidth]{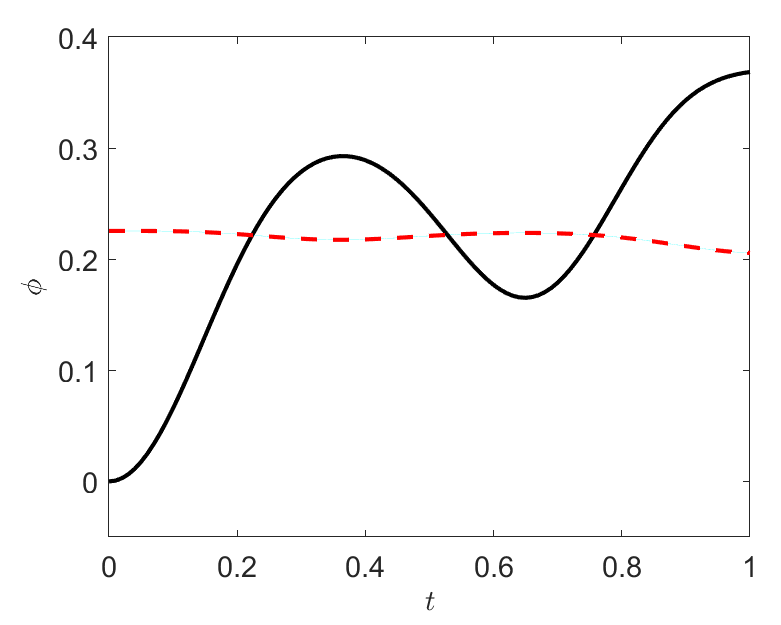}
    }
    \subfigure[Correcting the misspecified physical model in PINNs using an NN.]{
        \includegraphics[width=0.3\textwidth]{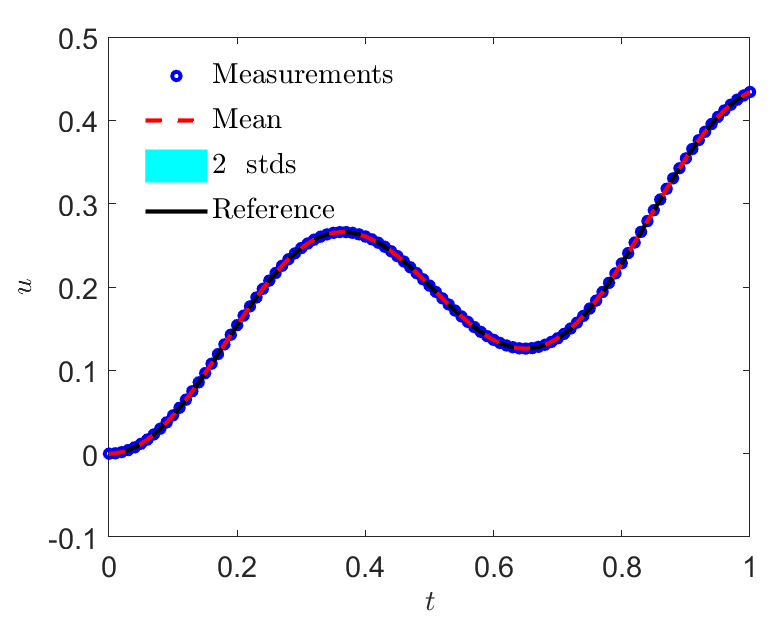}
        \includegraphics[width=0.3\textwidth]{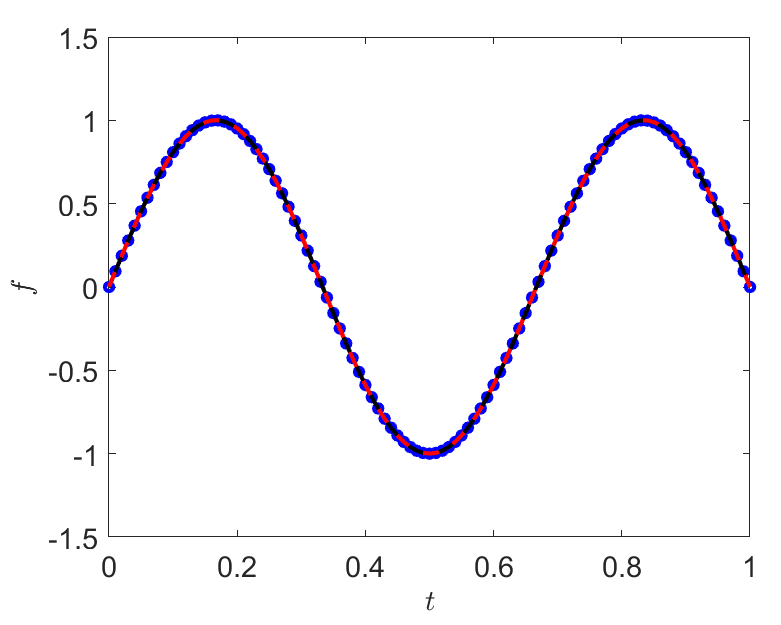}  
        \includegraphics[width=0.3\textwidth]{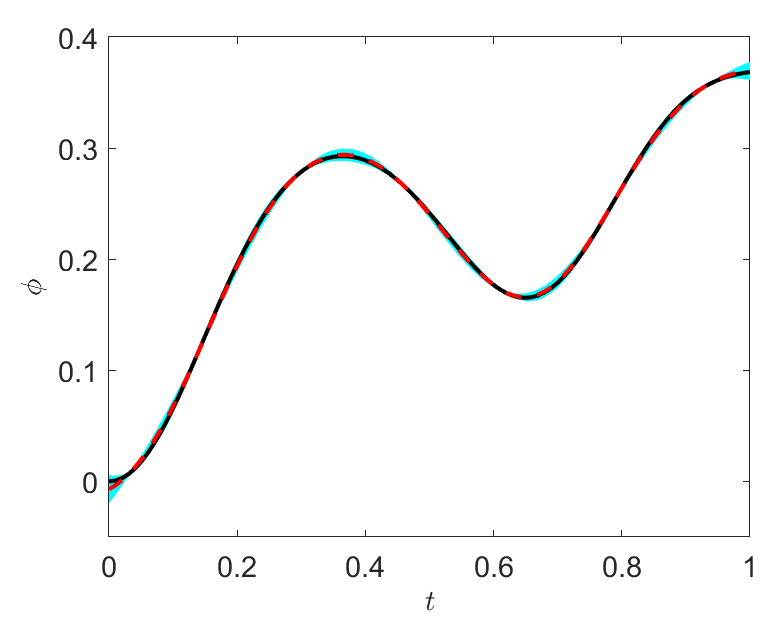}
    }
    \caption{
    {ODE system: Predictions for $u$, $f$ and $\phi$ from ensemble PINNs with clean and sufficient data. Ensemble PINNs with (a) the correct physical model, (b) a misspecified physical model, and (c) the misspecified physical model plus a correction. Blue circles: Representative measurements for $u$ and $f$; Red dashed line: Predicted mean; Black solid line: Reference solution.
    }}
    \label{fig:example_1_1}
\end{figure}

\begin{table}[ht]
    \footnotesize
    \centering
    \begin{tabular}{c|c|c|c|c|c}
    \hline\hline
    & $\tilde{\lambda}$ & Error of $\phi$ & Error of $u$ & Error of $f$ & Error of $\tilde{u}$\\
    \hline
    Case (A): Known model & $1.5000\pm 0.0001$ & $0.00\%$ & $0.00\%$ & $0.06\%$ & $0.01\%$\\
    \hline
    Case (B): Misspecified model & $0.2254\pm0.0001$ & $41.46\%$ & $1.73\%$ & $6.87\%$ & $4.58\%$\\
    \hline
    Case (C): Corrected model & $0.2$ & $0.55\%$ & $0.01\%$ & $0.10\%$ & $0.02\%$\\
    \hline\hline
    \end{tabular}
    \caption{ODE system: Predictions for $\phi$, $u$ and $f$ and the reconstruction $\tilde{u}$. The metric is relative $L_2$ error, and errors are computed between the reference and the mean of 20 PINNs trained independently. The exact value of $\lambda$ is $1.5$ and $\tilde{\lambda}$ denotes the PINN estimate of $\lambda$ ($\tilde{\lambda}=0.2$ is fixed in Case (C)).
    We note that $\tilde{u}$ is reconstructed by solving the identified ODE system with the identified $\phi$, the inferred $f$ and the initial condition $u_0$. Cubic spline data interpolation is employed to approximate the function $f$ from the prediction of $f$ on a uniform grid.}
    \label{tab:example_1_1}
\end{table}

As shown in Fig.~\ref{fig:example_1_1}(a), PINNs with correct physical model (Case (A)) enable accurate inference of $\phi$ and the surrogate model $u_\theta$ fits the data and satisfies the ODE simultaneously. In contrast, PINNs with misspecified physical model (Case (B)) yield completely wrong estimate of $\phi$ as displayed in Fig.~\ref{fig:example_1_1}(b). In Case (C), we add a DNN to correct the misspecified reaction model used in Case (B). As illustrated in Fig.~\ref{fig:example_1_1}(c) as well as Table~\ref{tab:example_1_1}, the predicted accuracy for $\phi$ improves significantly as compared to the result in Case (B). In addition, the computational accuracy for $u$, $f$ and $\phi$ in Case (C) is quite similar as in the Case (A), where the physical model is correctly specified. The above results demonstrate the effectiveness of the present approach. We note that the predicted uncertainties in all cases are quite small, which is reasonable since we use sufficient data for both $u$ and $f$ to determine the unknown reaction term here. Furthermore, we present further results as well as the related discussion for the case with clean but gappy data in \ref{sec:appendix_ode:1}.  

\begin{figure}[ht!]
    \centering
    \subfigure[B-PINN with the correct physical model.]{
        \includegraphics[width=0.3\textwidth]{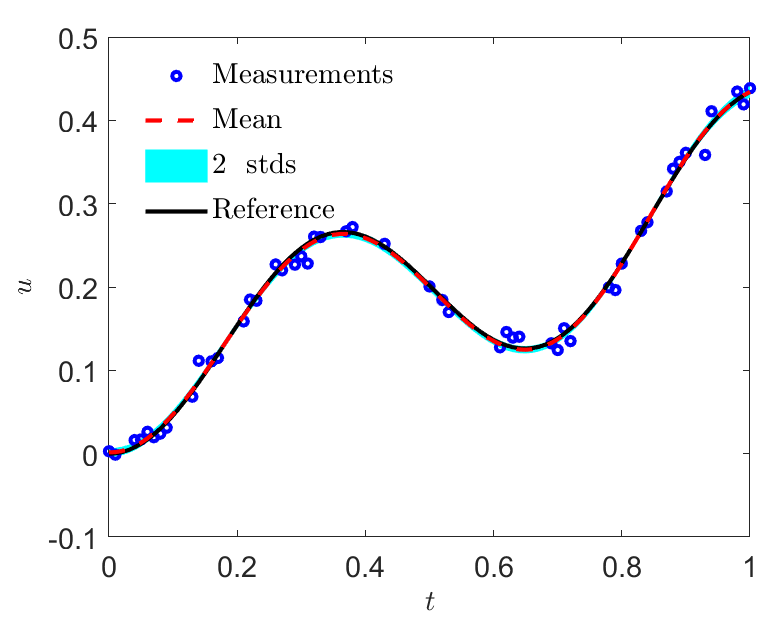}
        \includegraphics[width=0.3\textwidth]{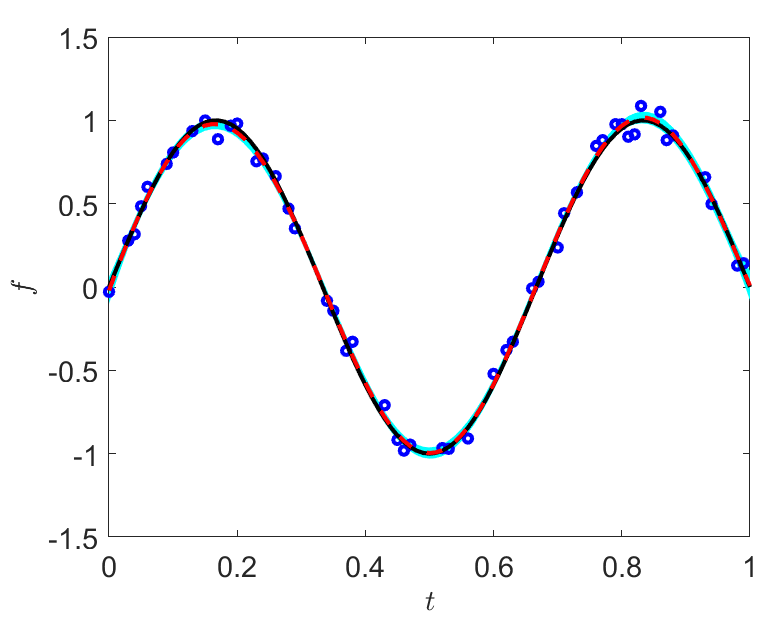}
        \includegraphics[width=0.3\textwidth]{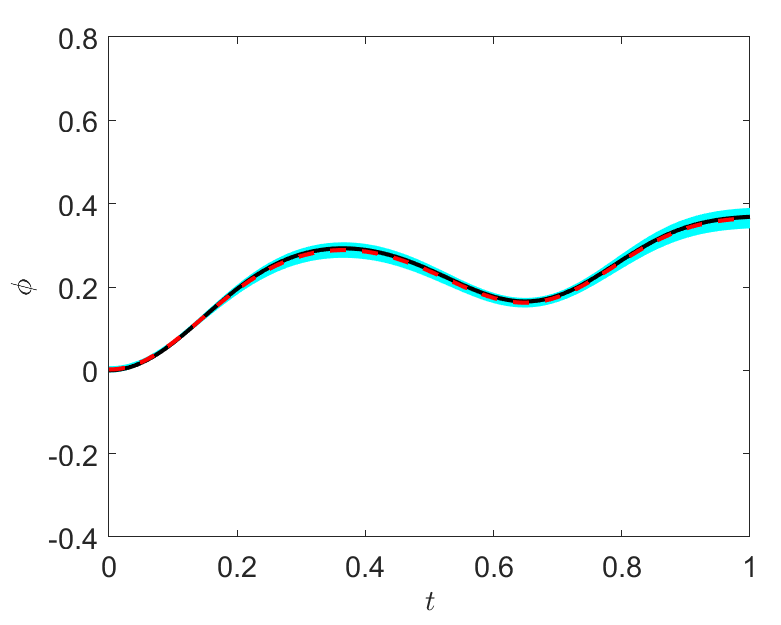}
    }
    \subfigure[B-PINN with a misspecified physical model.]{
        \includegraphics[width=0.3\textwidth]{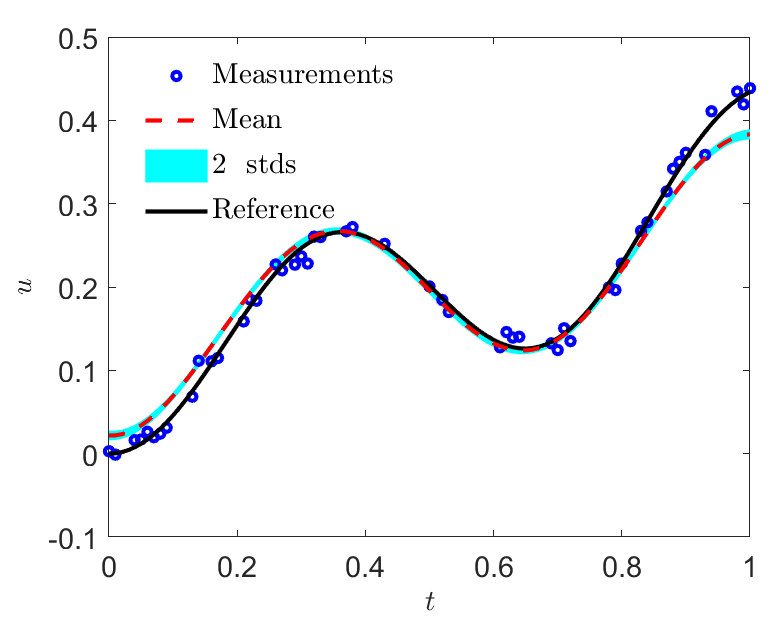}
        \includegraphics[width=0.3\textwidth]{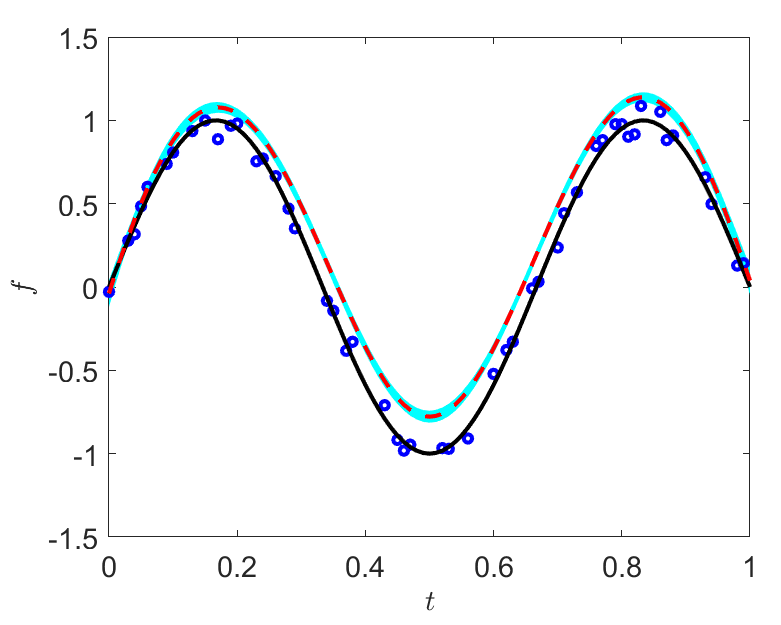}
        \includegraphics[width=0.3\textwidth]{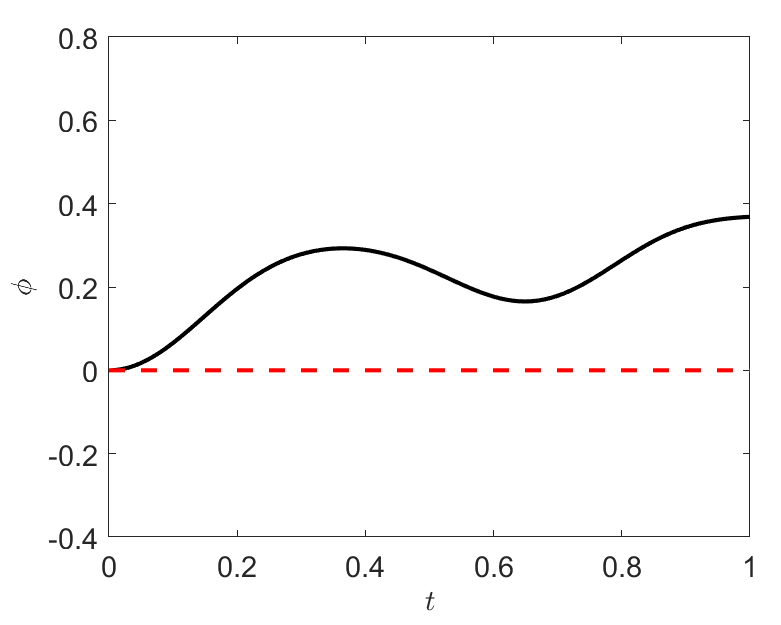}
        
    }
    \subfigure[Correcting the misspecified physical mdoel]{
        \includegraphics[width=0.3\textwidth]{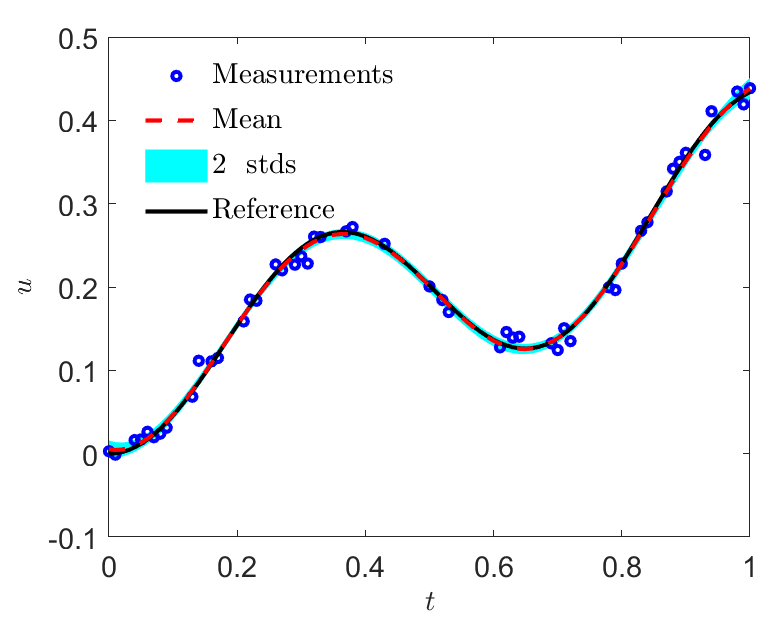}
        \includegraphics[width=0.3\textwidth]{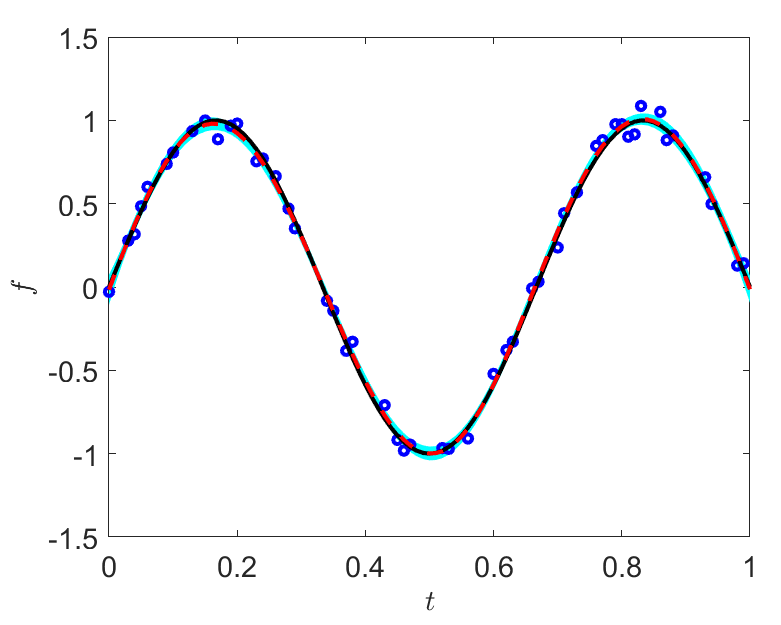}
        \includegraphics[width=0.3\textwidth]{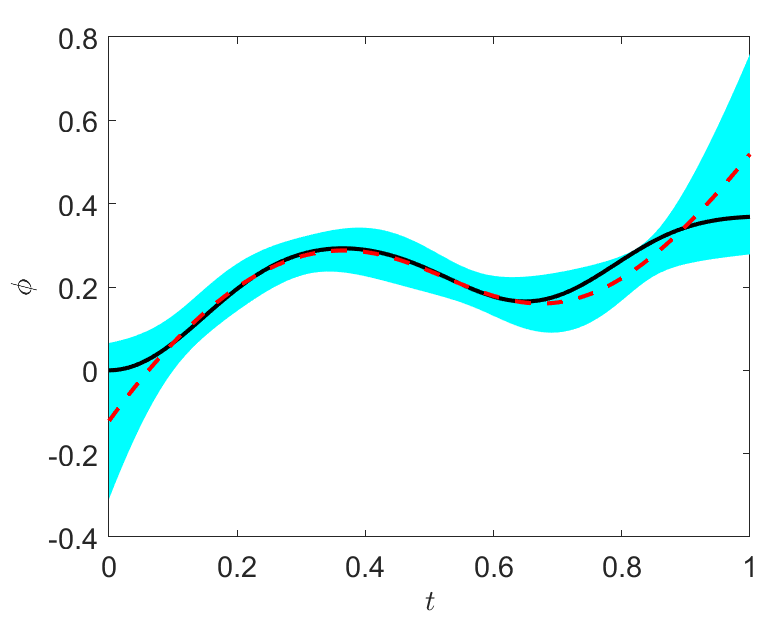}
    }
    \caption{
    ODE system: Predictions for $u$, $f$ and $\phi$ from B-PINNs with noisy and gappy data. B-PINNs with (a) correct physical model, (b) misspecified physical model, and (c) misspecified physical model plus a correction. Blue circles: Noisy measurements for $u$ and $f$; Red dashed line: Predicted mean from B-PINNs; Black solid line: Reference solution. Our approach alleviates the model misspecification issue significantly, providing more accurate predictions. Compared to correctly specified model, the increase in the predicted uncertainty is caused by the lack of knowledge in the physical model and could be considered as model uncertainty.
    }
    \label{fig:example_1_2}
\end{figure}

\begin{figure}[ht!]
    \centering
    \subfigure[]{
        \includegraphics[width=0.3\textwidth]{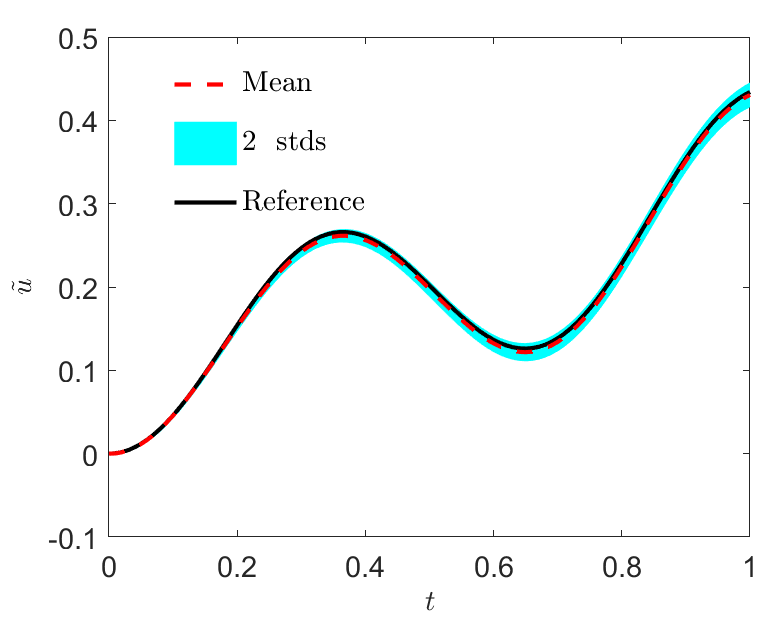}
        }
        \subfigure[]{
        \includegraphics[width=0.3\textwidth]{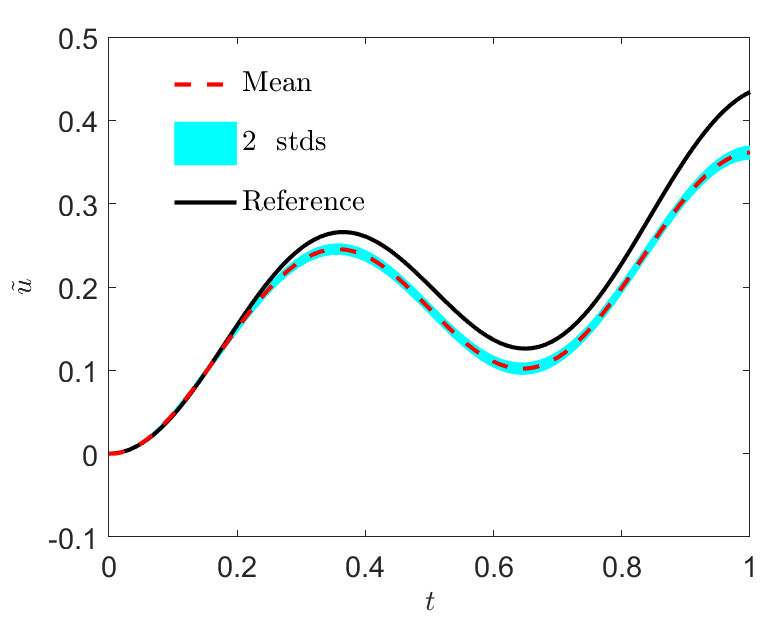}
        }
        \subfigure[]{
        \includegraphics[width=0.3\textwidth]{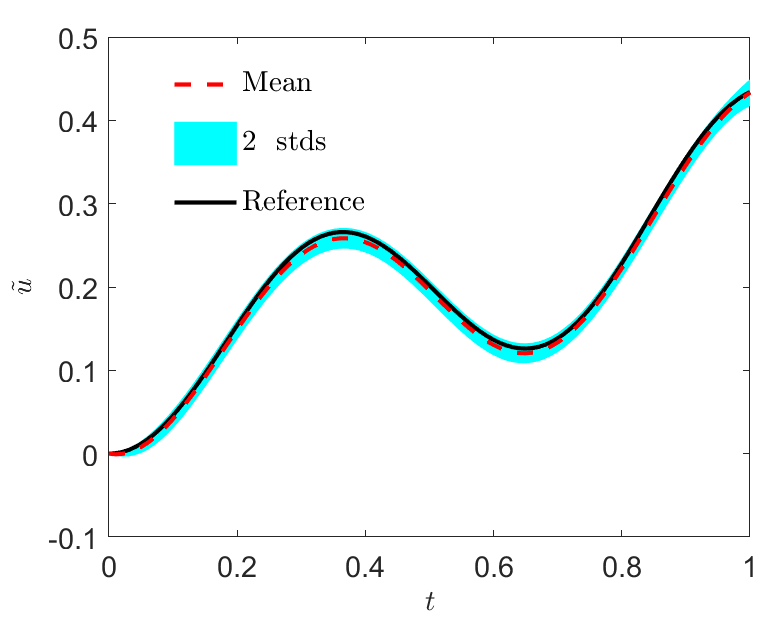}
        }
    \caption{
    ODE system: Solving the identified system with the inferred $f$ and $\phi$ from B-PINNs with the initial condition $u_0=0$. The inferred $f$ and $\phi$ come from (a) correct physical model, (b) misspecified physical model, and (c) our approach. Red dashed line: Predicted mean; Black solid line: Reference solution. It shows that the misspecified physical model leads to wrong reconstruction $\tilde{u}$ while our approach is able to correct it. Due to the lack of knowledge of the model, our approach yields larger uncertainty than the correct physical model.
    }
    \label{fig:example_1_3}
\end{figure}

\begin{table}[ht]
    \footnotesize
    \centering
    \begin{tabular}{c|c|c}
    \hline\hline
    & $\tilde{\lambda}$ & Error of $\tilde{u}$\\
    \hline
    Case (A): Known model & $1.4867\pm 0.0488$  & $1.56\%$\\
    \hline
    Case (B): Misspecified model & $0.0000\pm0.0000$ &   $12.73\%$\\
    \hline
    Case (C): Corrected model & 0.2 & $2.61\%$\\
    \hline\hline
    \end{tabular}
    \caption{ODE system: Relative $L_2$ error for the reconstructed $u$ from B-PINNs. Here the computational errors are calculated using the predicted mean from B-PINNs.}
    \label{tab:example_1_2}
\end{table}

We now keep the same setup as in Case (A) - (C) but assume that the training data are both noisy and gappy. In these cases, $50$ data of $f$ and $u$ are randomly sampled from $t\in[0, 1]$ and corrupted by independent additive Gaussian noises with known noise scales ($0.05$ for $f$ and $0.01$ for $u$). As discussed in Sec. \ref{sec:2}, we will employ the B-PINNs to quantify the uncertainties arsing from the noise as well as the incompleteness of the training data. As shown in Fig. \ref{fig:example_1_2}, (1) B-PINNs with correct physical model can obtain accurate predictions for both $u$, $f$, and $\phi$, and the computational errors for $\phi$ are generally bounded by the predicted uncertainties (Case (A), Fig. \ref{fig:example_1_2}(a)); (2) 
B-PINNs with the misspecified physical model fails to provide accurate inference for $u$, $f$, or $\phi$ (Case (B), Fig. \ref{fig:example_1_2}(b)); and (3) B-PINNs with a DNN as correction to the misspecified model is able to improve the predicted accuracy for $u$, $f$, and $\phi$, as compared to the results in Case (B) (Case (C), Fig. \ref{fig:example_1_2}(c)). Similar as in Case (A), the computational errors for $\phi$ are bounded by the predicted uncertainties here. 

In addition to the relation between the computational errors and the predicted uncertainties for $\phi$ discussed above, we would like to further discuss that: (1) it is interesting to observe in Case (A) and Case (C) that the predicted uncertainty near the right end is larger than that near the left end. We note that the results here are  reasonable since we do not have data for $f$ on the right end; (2)  the predicted uncertainties of $\phi$ for $t \in [0, 1]$ in Case (C) are larger than those in Case (A) although we use the same training data as well as prior for the BNN that is used to approximate $u$.  We remark that the difference between the uncertainties in these two cases is caused by the lack of knowledge on the physical model and reflects the uncertainty in physical models. 

We further verify the learned governing equation from the proposed approach. Specifically, we solve the equation ${du}/{dt} = f(t) + \phi(t)$ with the inferred $f$, $\phi$, and the initial condition $u_0=0$, using MATLAB ode45 to see if the solution for $u$ agrees with the reference solution. We denote the solution to the identified system by $\tilde{u}$. In particular, 1000 posterior samples of $f$ and $\phi$ from the B-PINNs are used to compute $\tilde{u}$, as displayed in  Fig.~\ref{fig:example_1_3}. We further illustrate the relative $L_2$ errors between the mean for $\tilde{u}$ and the reference solution in Table ~\ref{tab:example_1_2}. As we can see, PINNs with correct physical model leads to the lowest error in $\tilde{u}$. Adding a DNN to correct the misspecified model can significantly improve the accuracy in $\tilde{u}$ as compared to the case with misspecified model, and the computational errors are slightly larger than the ones from the correct physical model. We also present the computational errors for $\tilde{u}$ from the ensemble PINNs in Table~ \ref{tab:example_1_1}, which will not be discussed again since the results are similar as the above ones.  Finally, we would like to point out that (1) the uncertainties in the reconstructed $u$ (i.e., $\tilde{u}$) are caused by the predicted uncertainties for $f$ and $\phi$ from B-PINNs, and (2) the computational errors for the mean of $\tilde{u}$ in Case (A) and (C) are of the same order as the noise in the measurements, suggesting the reasonableness of the current results.

As discussed in Sec. \ref{sec:2}, the discovered physical models or governing equations up to now are expressed using the trained DNNs (i.e., $s_{\psi}$), which are not explicitly presented. We can then combine the proposed method with the symbolic regression as the further step to obtain the explicit governing equations. Here we take the results in Case (C) from both ensemble PINNs and B-PINNs to demonstrate how to get the analytical expressions for the trained NNs, i.e., $s_{\psi}$. Recall that with the present methodology the reaction term is now expressed as $\phi(t) = \tilde{\lambda} \cos(u_\theta(t)) + s_\psi(t)$. Assume that we now have the knowledge that the reaction term is autonomous with respect to the solution, i.e. $\phi = s(u)$. We can then adopt the symbolic regression developed in \cite{cranmer2023interpretable} to find the analytical expression of $s(u)$ from $u_\theta$ and $s_\psi$. The data for the symbolic regression are generated as $\{u_\theta(t_i), s_\psi(t_i) + \tilde{\lambda} \cos(u_\theta(t_i))\}_{i=1}^N$, where $t_i, i=1,...,N$ are $101$ uniform grids points on $[0, 1]$, and $u_\theta$ and $s_\psi$ are the predicted means from the trained ensemble PINNs/B-PINNs.  As displayed in Table ~\ref{tab:example_1_3}, the recovered reaction model via symbolic regression agrees well with the reference solution as we have sufficient clean data. However, for the case with noisy and gappy data, the expression for the reaction model from symbolic regression does not match the reference solution. The result is expected since we can see in Fig. \ref{fig:example_1_2}(c) that the predicted mean for $\phi$ does not agree well with the reference solution well and the predicted uncertainties are generally large. 
Generally, increasing the number of training data is able to reduce the predicted uncertainty in B-PINNs \cite{psaros2023uncertainty,zou2022neuraluq}. Here we also show that the learned reaction model agrees better with the reference solution as we increase the training data from 50 to 101 in B-PINNs. A recommendation here is  to use the predicted uncertainty as guidance, i.e., we use the symbolic regression for discovering equation when the predicted uncertainty is small to achieve better accuracy.
More results for the symbolic regression can be found in \ref{sec:appendix_ode:2}.

\begin{table}[ht]
    \footnotesize
    \centering
    \begin{tabular}{c|c|c|c}
    \hline\hline
    Reference  & Clean and sufficient data & Noisy and gappy data & Noisy and sufficient data\\
    \hline
    $1.5u(1-u)$ & $1.4980u - 1.4929u^2$ & $u/(u + 0.6213)$ & $1.3540 u - u^2$\\
    \hline\hline
    \end{tabular}
    \caption{
    ODE system: Discovered equations from $\phi(u)$ using a symbolic regression method \cite{cranmer2023interpretable} (see Fig~\ref{fig:appendix_2} for plots of the discovered models). In all three cases, the symbolic regression is performed on the predicted mean from ensemble PINNs and B-PINNs.
    }
    \label{tab:example_1_3}
\end{table}

\subsection{Reaction-diffusion equation}\label{subsec:reaction_diffusion}
We now consider the following time-dependent PDE that describes a reaction-diffusion system:
\begin{equation}\label{eq:reaction_diffusion}
    \begin{aligned}
        & \frac{\partial u}{\partial t} = D\frac{\partial^2 u}{\partial x^2} + \lambda g(u) + f(x, t),~ x\in [0, 1],~ t\in[0, 1],\\
        & u(x, 0) = 0.5\sin^2(\pi x),~ x\in[0, 1],\\ 
        & u(0, t) = u(1, t) = 0, ~t\in[0, 1],
    \end{aligned}
\end{equation}
where $D=0.01$ denotes the diffusion coefficient, $g(u) = u(1 - u)$ is the reaction model, $\lambda$ is the reaction rate, and $f(x, t)=0$ is the source term. 
Similar as in Sec. \ref{subsec:ode}, we test the following three cases:
\begin{enumerate}
    \item Case (A): The reaction model is correctly specified and the reaction rate $\lambda \geq 0$ is unknown.
    \item Case (B): The reaction model is misspecified as $u^2$ and the reaction rate $\lambda \geq 0$ is unknown.
    \item Case (C): The reaction model is misspecified as $u^2$ and we correct it with a DNN, i.e., $\lambda u^2 + s$ (with known $\lambda=1$). 
\end{enumerate}
We denote the reaction term as $\phi$ and we would like to identify $\phi$ from data of $u$, $f$ and different physical models. In Case (A)/(B)/(C), the estimate of $\phi$ is $\tilde{\lambda} u_\theta (1 - u_\theta)$/$\tilde{\lambda} u_\theta^2$/$\tilde{\lambda} u_\theta^2 + s_\psi$, where $\tilde{\lambda}$ is the estimate of $\lambda$. We note that in Case (C) we assume $\lambda=1$ is known because the model misspecification induced by wrong $\lambda$ can be fully covered by $s$.

\begin{figure}[ht!]
    \centering
    \subfigure[Predictions for $u$.]
    {
        \includegraphics[width=0.23\textwidth]
        {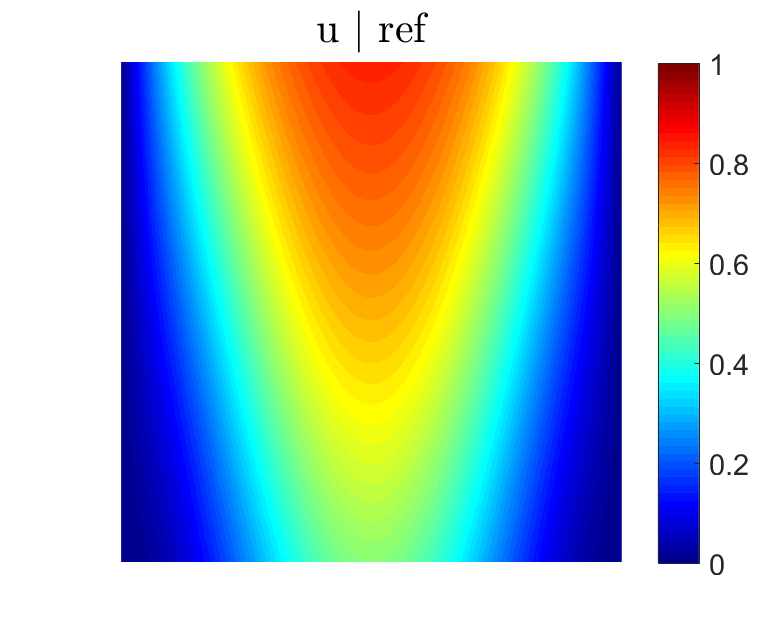}
        \includegraphics[width=0.23\textwidth]{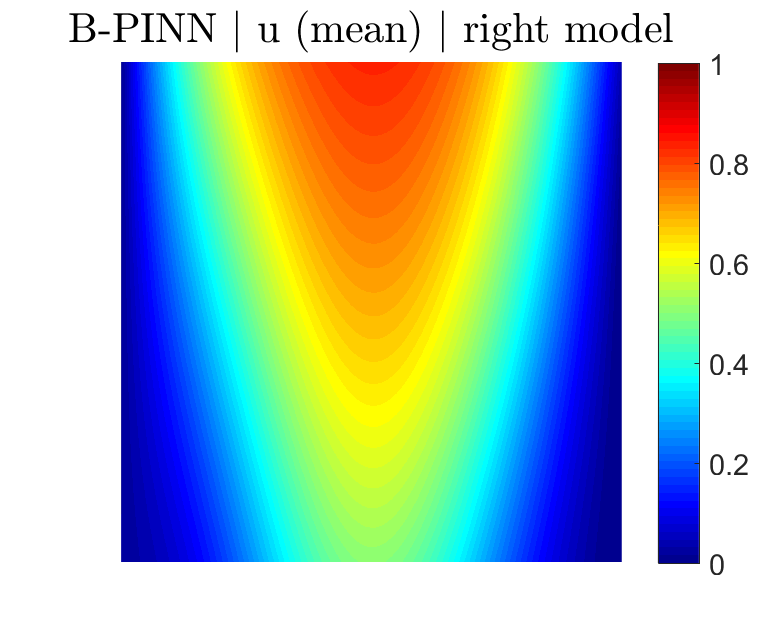}
        \includegraphics[width=0.23\textwidth]{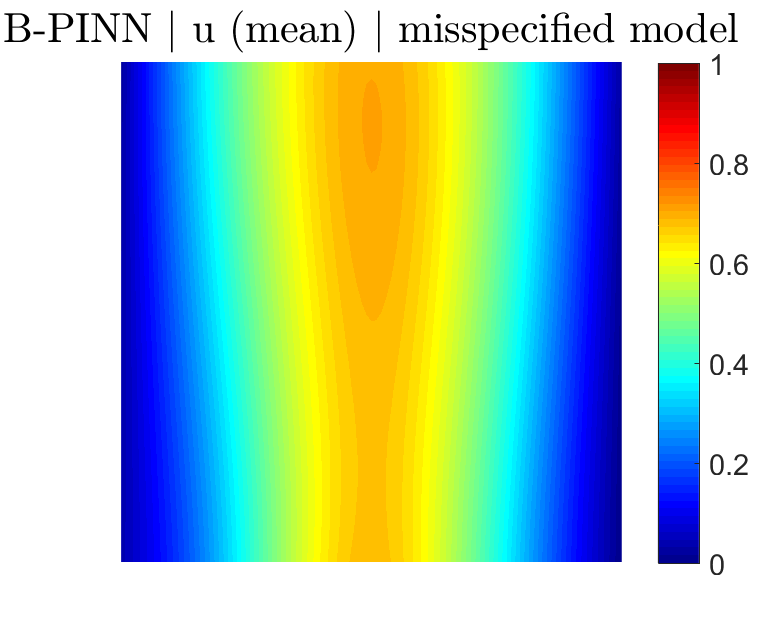}
        \includegraphics[width=0.23\textwidth]{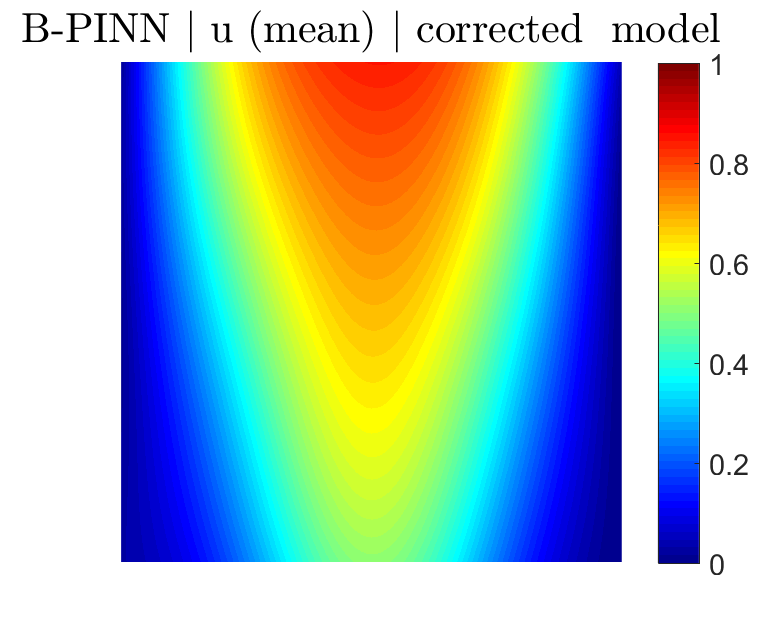}
    }
    \subfigure[Predictions for $\phi$.]
    {
        \includegraphics[width=0.23\textwidth]
        {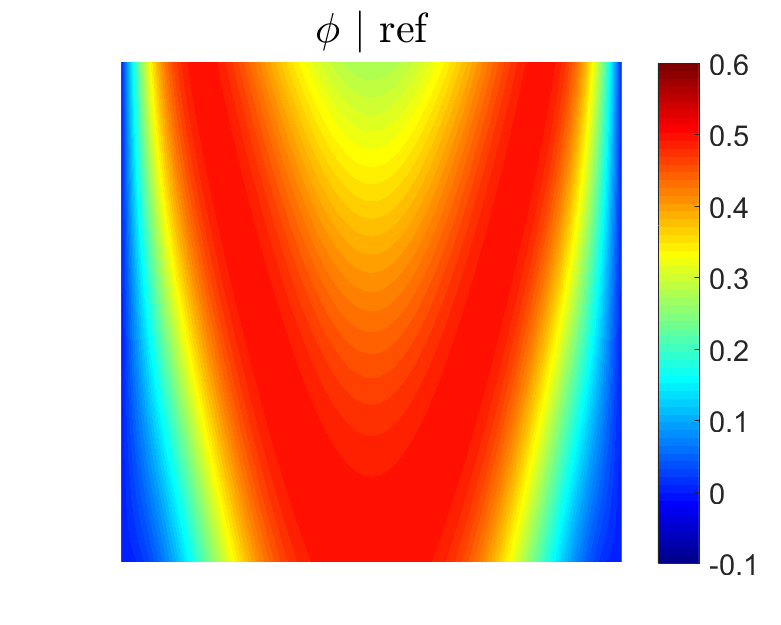}
        \includegraphics[width=0.23\textwidth]{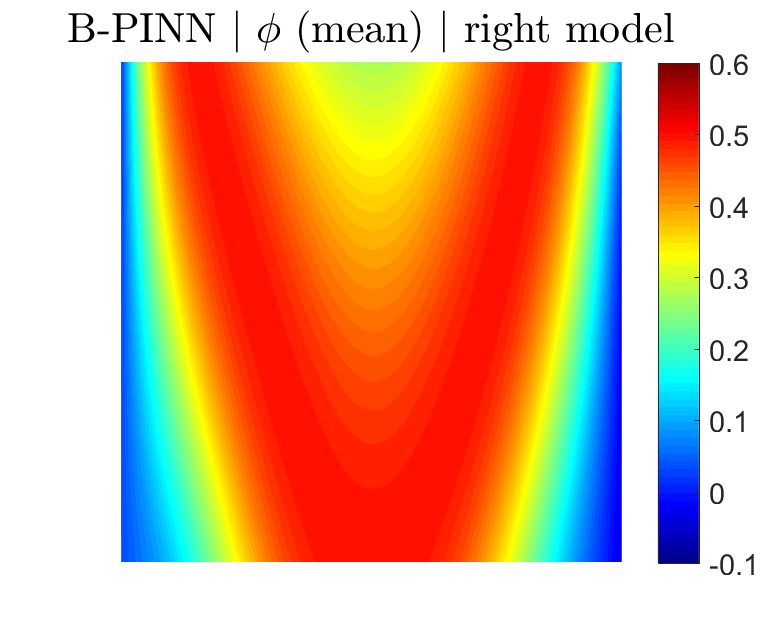}
        \includegraphics[width=0.23\textwidth]{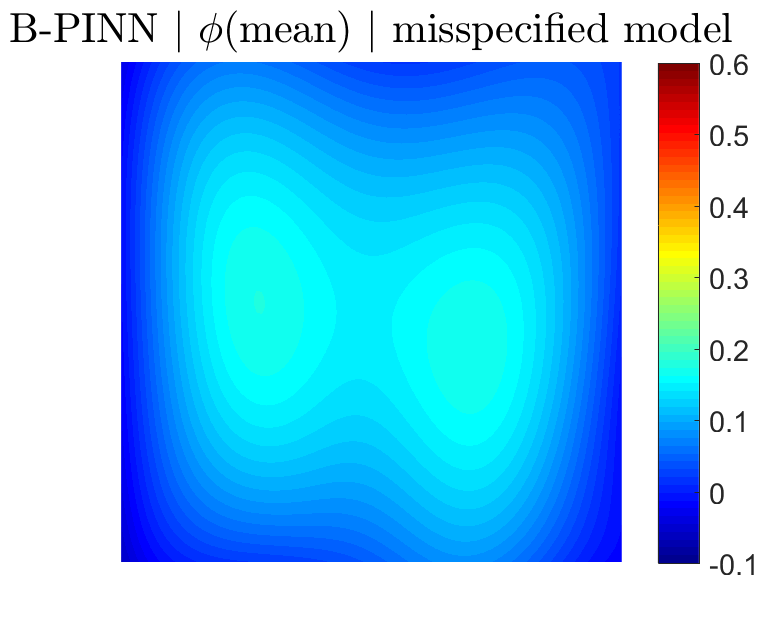}
        \includegraphics[width=0.23\textwidth]{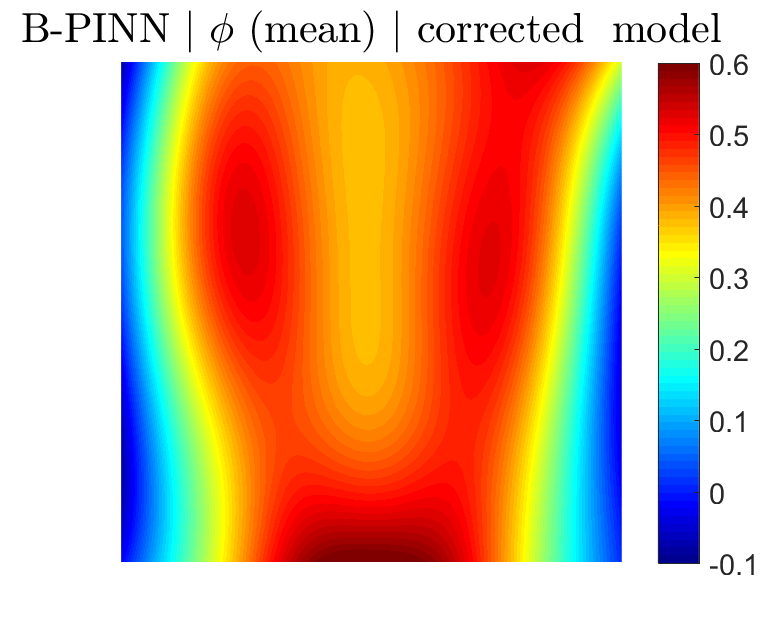}
    }
    \caption{
    Time-dependent reaction-diffusion system: Predicted mean for $u$ and $\phi$ from B-PINNs with noisy data of $u$ and $f$ and differently specified reaction models. Misspecified model ($u^2$) impairs the performance of B-PINNs significantly in terms of the accuracy in predicting $u$ and $\phi$. and the proposed approach is able to alleviate the issue caused by model misspecification; see Fig.~\ref{fig:example_2_2} for UQ. 
    }
    \label{fig:example_2_1}
\end{figure}

\begin{figure}[ht!]
    \centering
    \subfigure[Predictions for $u$.]
    {
        \includegraphics[width=0.3\textwidth]
        {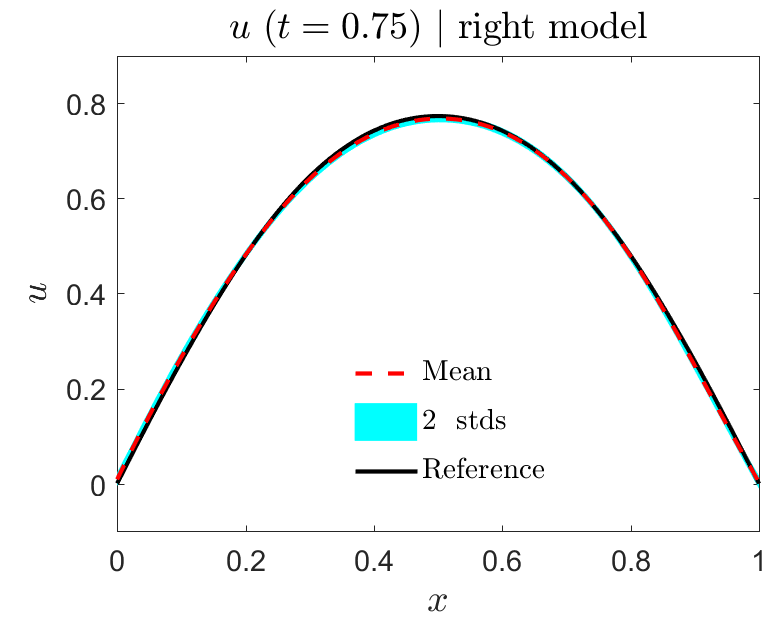}
        \includegraphics[width=0.3\textwidth]{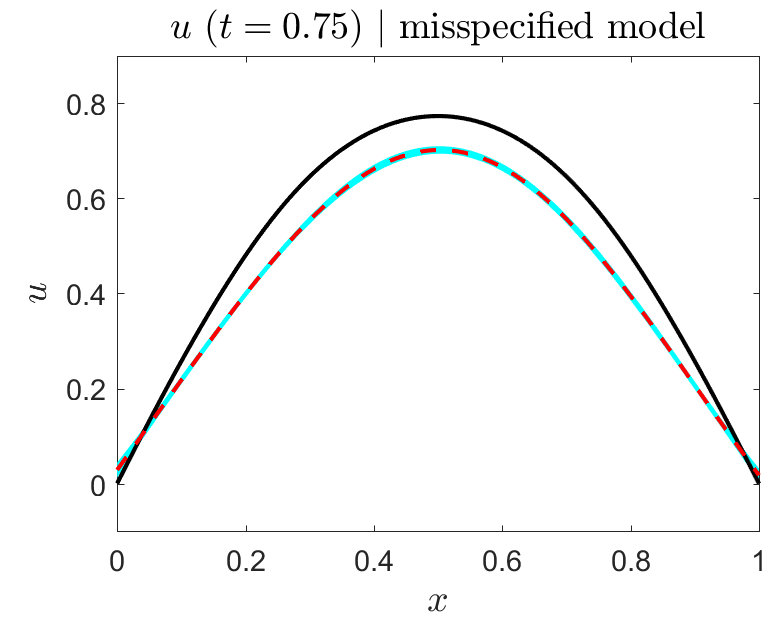}
        \includegraphics[width=0.3\textwidth]{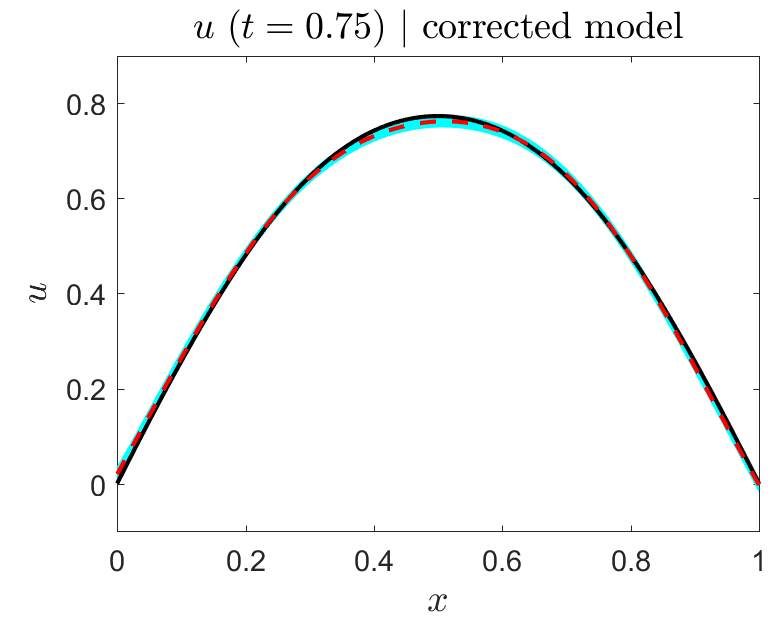}
    }
    \subfigure[Predictions for $\phi$.]
    {
         \includegraphics[width=0.3\textwidth]
        {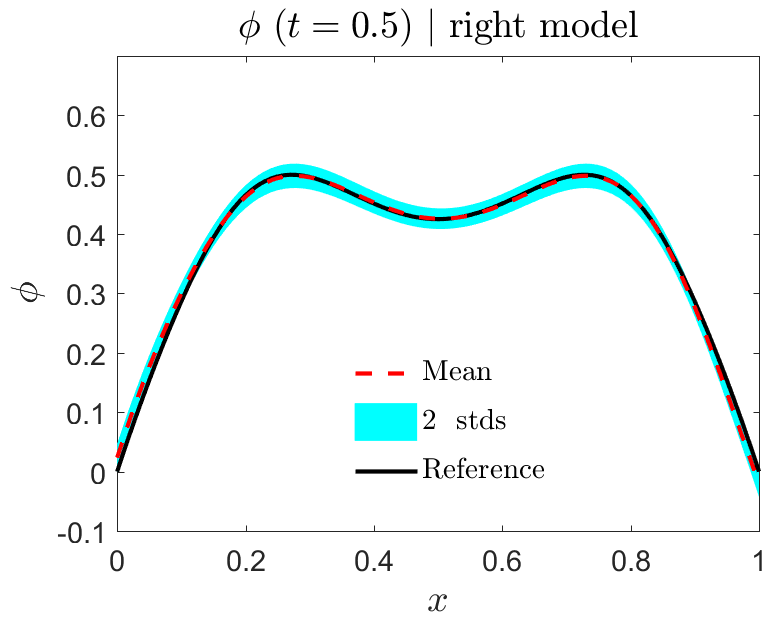}
        \includegraphics[width=0.3\textwidth]{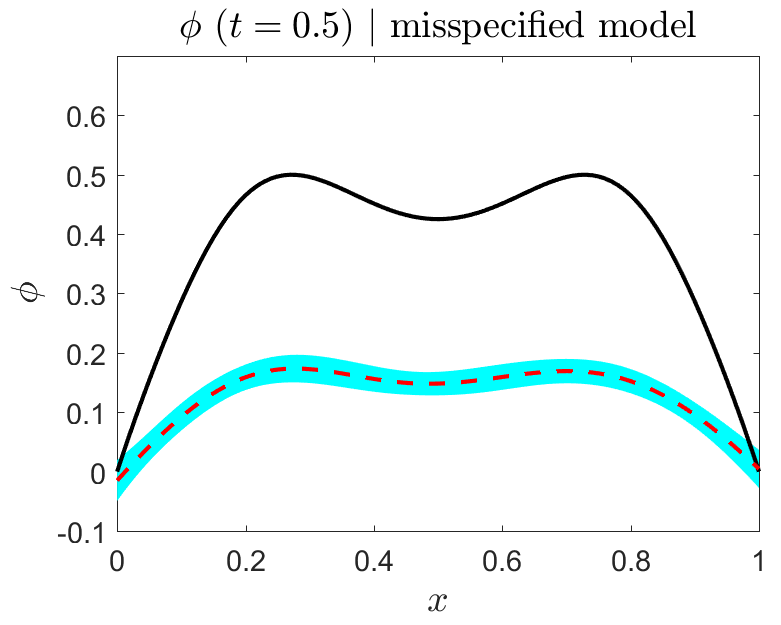}
        \includegraphics[width=0.3\textwidth]{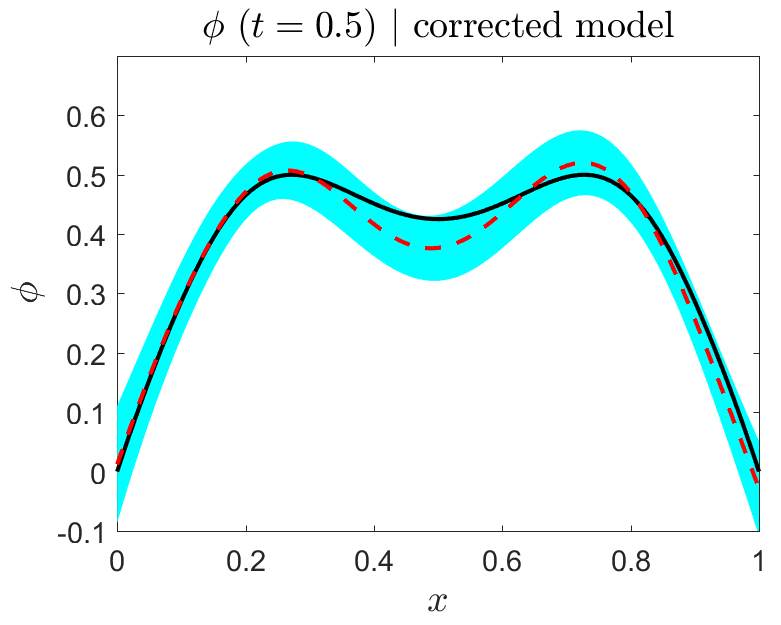}
    }
    \caption{
    Time-dependent reaction-diffusion system in slices ($t=0.5,0.75$): Predicted uncertainty for $u$ and $\phi$ from B-PINNs with noisy data of $u$ and $f$ and differently specified reaction models. Representative predicted uncertainties for (a) $u$, and (b) $\phi$. Red dashed line: Predicted mean from B-PINNs; Black solid line: Reference solution. Our approach is able to improve the the predicted mean and the predicted uncertainty significantly. The decrease in the accuracy of predicted mean and increase in the predicted uncertainty of our approach are caused by the lack of knowledge over the reaction model.
    }
    \label{fig:example_2_2}
\end{figure}

Here we only consider a general scenario, i.e., the training data are noisy and gappy. In this case, we employ B-PINNs to identify $\phi$ with uncertainties. We assume that we have 121 measurements on $u$ and $195$ measurements of $f$, which are evenly distributed in the temporal-spatial domain $(x, t)\in [0, 1]\times[0, 1]$. In addition, the measurement errors for both $u$ and $f$ are assumed to be Gaussian,  and the noise scales are $0.02$ and $0.05$ for $u$ and $f$, respectively. Further, the reference  solution is obtained by solving Eq.~\eqref{eq:reaction_diffusion} with $g(u)=u(1-u)$, $\lambda=2$, and $f(x, t)= 0$. Similar as in Sec. \ref{subsec:ode}, we use one DNN to model the sought solution $u$ and another DNN to approximate the discrepancy $s$ in Case (C). Details of the B-PINN method, e.g. hyperparameters of NNs and HMC, can be found in \ref{subsec:hyperparameter}.

As show in Fig. \ref{fig:example_2_1}, we observe that: (1) the predicted means for $u$ and $\phi$ agree the best with the reference solutions as the reaction model is correctly specified (Case (A), second column in Fig. \ref{fig:example_2_1}); (2)
the predictions for $u$ and $\phi$ show significant discrepancy as compared to the reference solution when the reaction model is misspecified (Case (B), third column in Fig. \ref{fig:example_2_1}); (3) the predicted mean for $u$ in Case (C) is similar as that in Case (A), in which the reaction model is correctly specified; and (4) the predicted mean for $\phi$ in Case (C) is not as good as the one in Case (A), but it is much better than the result in Case (B) where the reaction model is misspecified. All the results demonstrate that the added DNNs are able to correct the misspecified model which in turn enhance the computational accuracy in PINNs.

We further illustrate the predicted uncertainties for both $u$ and $\phi$ in Fig. \ref{fig:example_2_2}. In particular, we only depict the uncertainties at one representative time for simplicity. As we can see, (1) the computational errors for both $u$ and $\phi$ are generally bounded by the predicted uncertainties in Case (A) (first column in Fig. \ref{fig:example_2_2}) and Case (C) (third column in Fig. \ref{fig:example_2_2}), while they cannot be bounded by the predicted uncertainties in the case with misspecified reaction model i.e. Case (B) (second column in Fig. \ref{fig:example_2_2}), and (2) the predicted uncertainties for $u$ and $\phi$ are generally larger in Case (C) (third column in Fig. \ref{fig:example_2_2}) than those in Case (A) (first column in Fig. \ref{fig:example_2_2}), which are similar as the results in Sec. \ref{subsec:ode}.  Again, we attribute the difference between the predicted uncertainties for Case (A) and (C) to the model uncertainty.

\subsection{2D non-Newtonian flows}\label{subsec:non_newtonian_flow}
In this section, we test the capability of the present approach in non-Newtonian flows where we misspecify the constitutive relation as the Newtonian one, and then we employ DNNs to correct the misspecified models. In particular, we consider the steady flows in a 2D channel and a 2D cavity, and the employed non-Newtonian fluids are  power-law ones. The constitutive relation  in a power-law fluid is expressed as $\mu = \mu_0 |\mathbf{S}|^{n-1}$, where $\mu$ is the dynamic viscosity of a fluid, $\mu_0$ is a constant, $\mathbf{S} = \nabla \mathbf{u} + (\nabla\mathbf{u})^T$ denotes the shear rate ($|\mathbf{S}| = \sqrt{(\mathbf{S}:\mathbf{S})/2}$), $\bm{u}$ is the fluid velocify, and $n$ is the power law index. Without loss of generality, the fluids are shear-thinning (pseudoplastic, $n < 1$) and shear-thickening (dilatant, $n > 1$) in the channel and cavity flows, respectively.

\subsubsection{Channel flow}
\label{subsec:channel_flow}

The following 2D non-Newtonian channel flow is considered here and the corresponding governing equation is expressed as  \cite{wang2015localized}:
\begin{equation}\label{eq:channel_flow:1}
\begin{aligned}
    &\frac{\partial}{\partial y} (\mu(y) \frac{\partial u}{\partial y}) - \frac{\partial P}{\partial x} = f(y), ~y\in[-H/2, H/2],\\
    &u(-H/2) = u(H/2) = 0, 
\end{aligned}
\end{equation}
where $H=1$, $\frac{\partial P}{\partial x} = c$ is a constant, and $f(y) = 0$. The analytical solution reads as
\begin{equation}\label{eq:power_law_fluid}
    u(y) = \frac{n}{n+1} (-\frac{1}{\mu_0}\frac{\partial P}{\partial x})^{1/n}[(\frac{H}{2})^{1+1/n} - |y|^{1+1/n}].
\end{equation}
For this specific case, we set $n=0.25, c=-1, H=1, \mu_0=0.5$ to generate the reference solution as well as training data. 

In the first test case, we assume that we have $30$ clean measurements of $u$ and $51$ clean measurements of $f$, randomly and uniformly sampled from $[-H/2, H/2]$, respectively. We assume the fluid is a Newtonian one and the viscosity is an unknown constant.  We then employ the PINN method to identify the unknown viscosity $\mu_1$ in the following equation:
\begin{equation}\label{eq:channel_flow:2}
    \frac{\partial}{\partial y} (\mu_1\frac{\partial u}{\partial y}) - \frac{\partial P}{\partial x} = f(y),
\end{equation}
which describes Newtonian flows. Specifically, the boundary condition is hard-encoded in the equation for $u$ so that we only have two terms in the PINN loss function \eqref{eq:loss_PINN}, i.e., $\mathcal{L}_{\mathcal{D}_u}$ for the data of $u$ and $\mathcal{L}_{PDE}$ for the equation. As shown in Fig.~\ref{fig:1}(b), we cannot minimize $\mathcal{L}_{\mathcal{D}_u}$ and $\mathcal{L}_{PDE}$ simultaneously in PINNs, regardless of the choice of belief weights for those two terms, indicating that the NN surrogate model $u_\theta$ is not able to fit the non-Newtonian data and satisfy the Newtonian physical law at the same time.

\begin{figure}[ht!]
    \centering
    \subfigure[$\mu_1 \frac{\partial^2 u}{\partial y^2} - \frac{\partial P}{\partial x} + s(y) = f(y)$]{
    \includegraphics[width=0.23\textwidth]{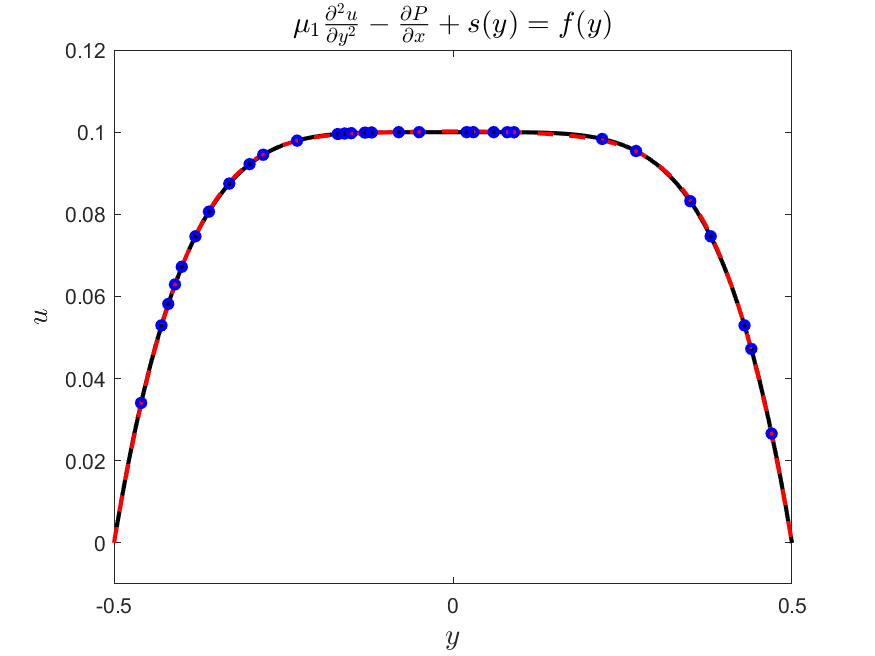}
    \includegraphics[width=0.23\textwidth]{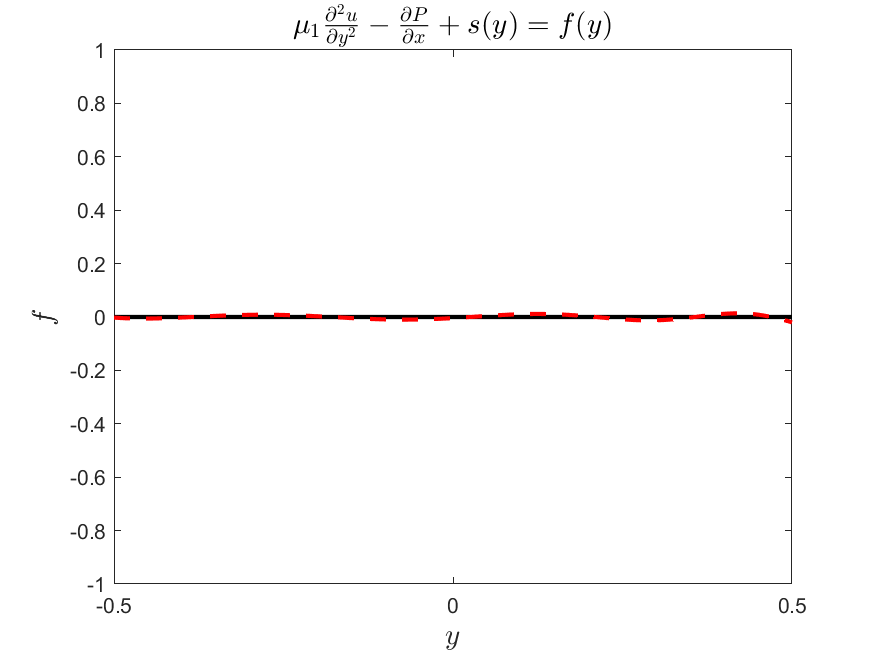}
    }
     \subfigure[$\frac{\partial}{\partial y}(\mu(y)\frac{\partial u}{\partial y}) - \frac{\partial P}{\partial x} = f(y)$]{
        \includegraphics[width=0.23\textwidth]{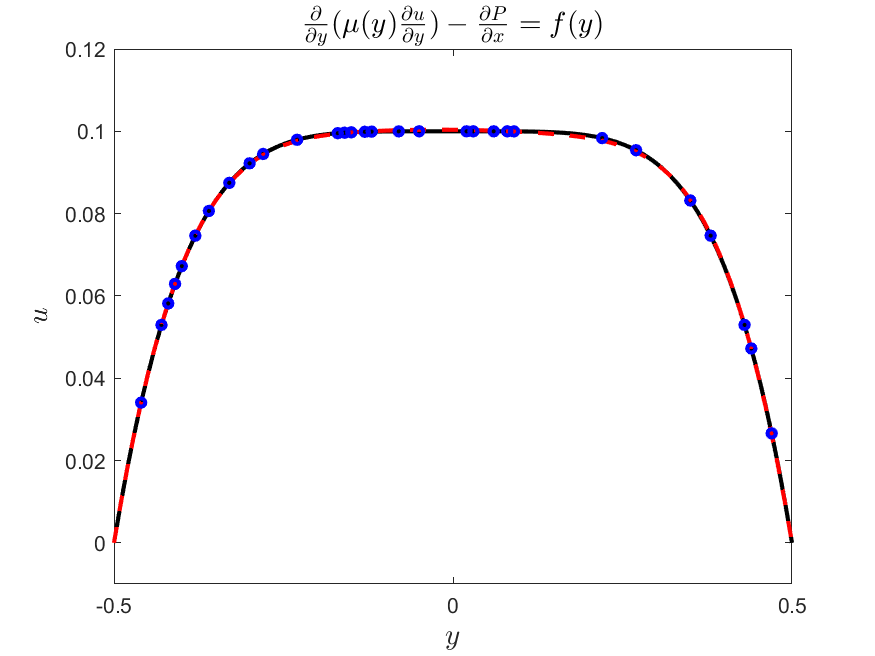}
        \includegraphics[width=0.23\textwidth]{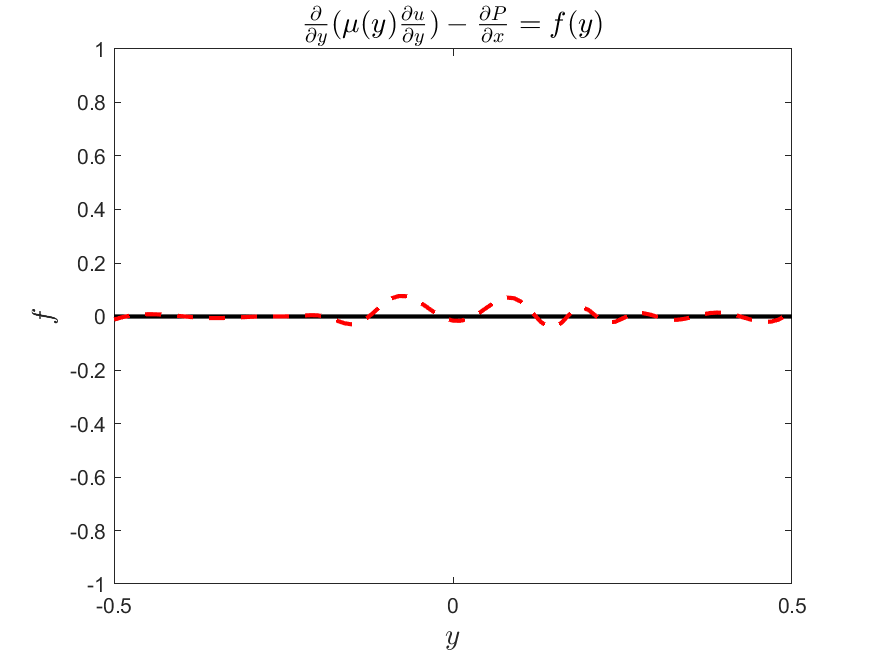}
    }
    \subfigure[Loss history during training]{
        \includegraphics[width=0.3\textwidth]{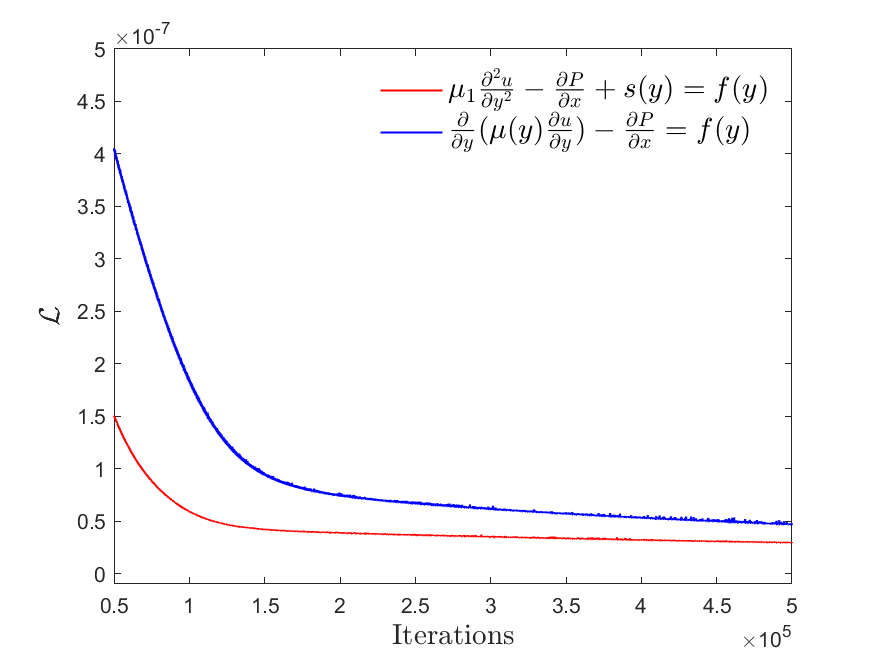}
        \includegraphics[width=0.3\textwidth]{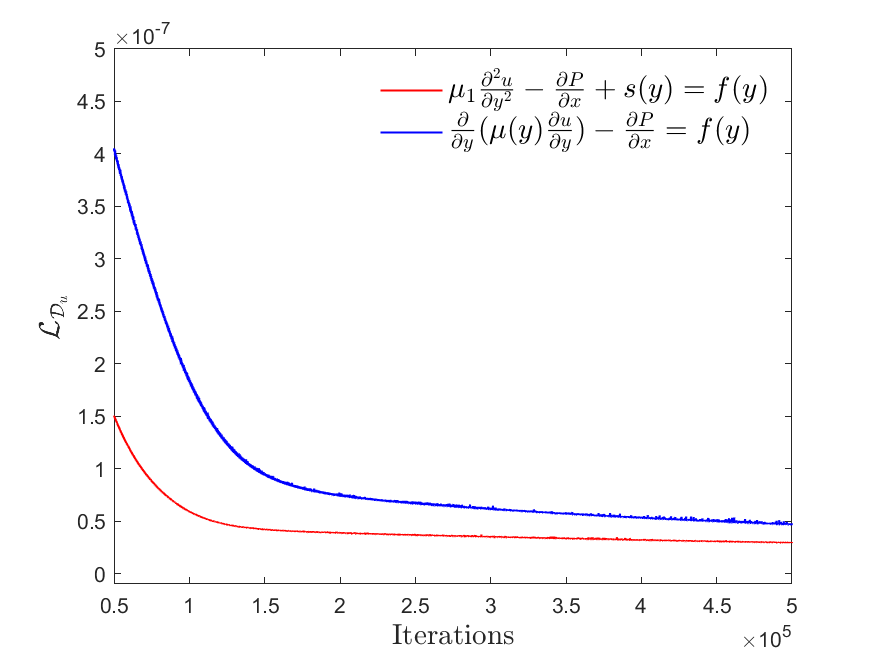}
        \includegraphics[width=0.3\textwidth]{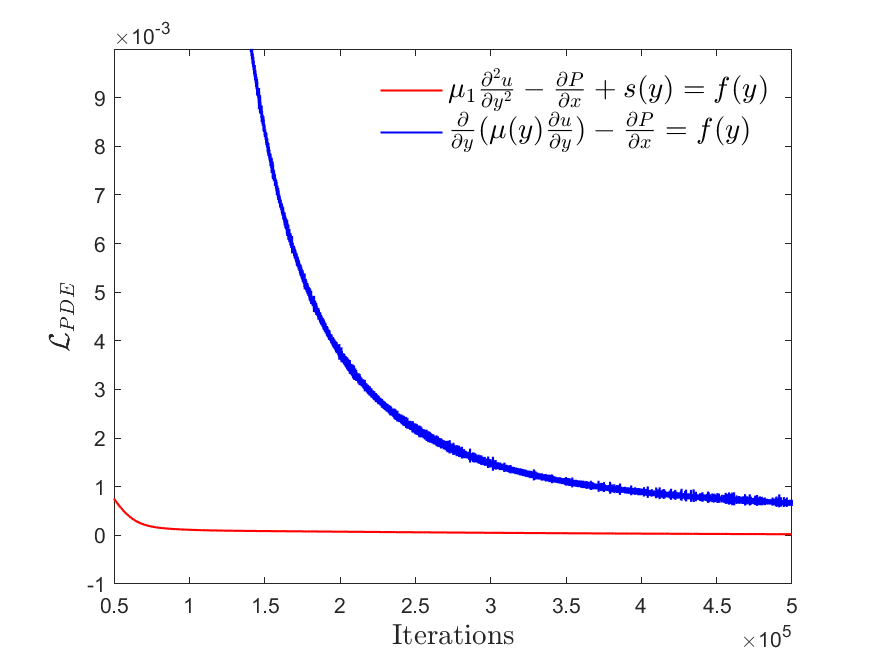}
    }
      
    \caption{
    2D non-Newtonian channel flow: Correcting the misspecified physics with two different approaches, i.e. (a) modeling the discrepancy in the equation $s(y)$  with a DNN, and (b) modeling the unknown space-dependent viscosity $\mu(y)$ with a DNN. In (c), the PINN training losses are presented and they converge to smaller values for both $\mathcal{L}_{\mathcal{D}_u}$ and $\mathcal{L}_{PDE}$ in the case where we add a DNN as correction as compared to the one in which we use a DNN to approximate the viscosity.
    }
    \label{fig:example_3_2}
\end{figure}

Similar as in previous two examples in this section, we can add a DNN as the correction in Eq. \eqref{eq:channel_flow:2} to address the model misspecification issue in PINNs. The equation with the correction term can now be expressed as follows:
\begin{equation}\label{eq:channel_flow:3}
    \frac{\partial}{\partial y} (\mu_1 \frac{\partial u}{\partial y}) - \frac{\partial P}{\partial x} + s(y) = f(y),
\end{equation}
where $s$ denotes the correction term. 
Here we treat $\mu_1$ as known and set $\mu_1=0.1$ since we have a DNN to model the discrepancy. We present the results from the PINNs with a correction in Fig. \ref{fig:example_3_2}(a), and we can see that the current predictions for $u$ and $f$ agree with the reference solutions much better than in the case where the model is misspecified.

\begin{figure}[ht!]
    \centering
         \subfigure[B-PINN with misspecified model.]{
        \includegraphics[width=0.3\textwidth]{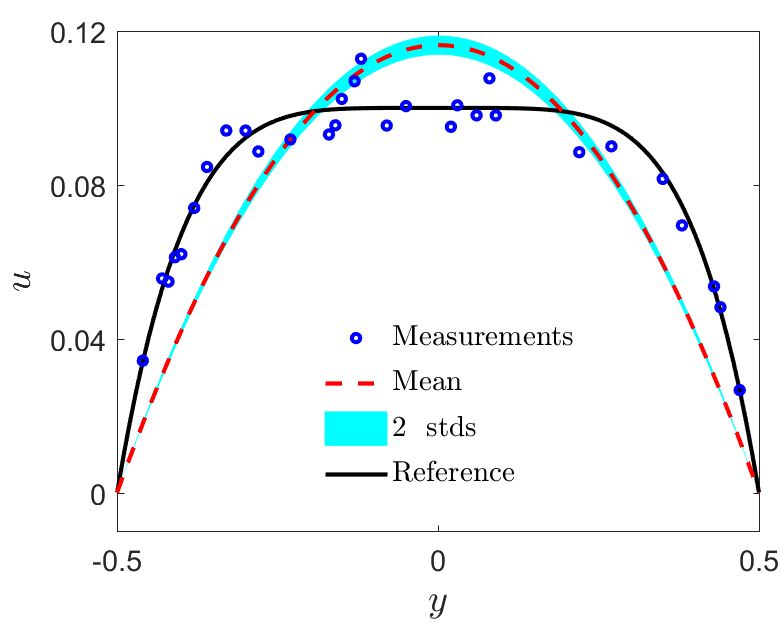}
        \includegraphics[width=0.3\textwidth]{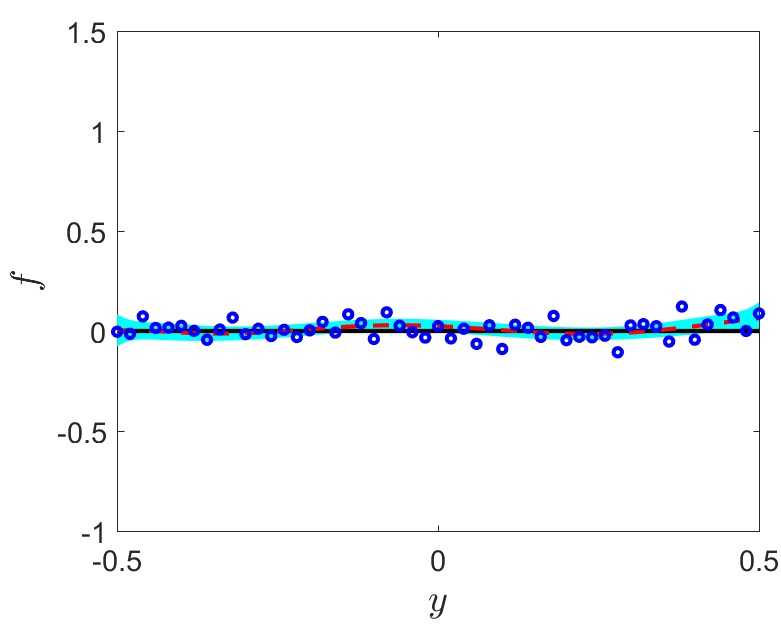}
    }
    \subfigure[Correcting the misspecified model with a DNN in B-PINNs.]{
        \includegraphics[width=0.3\textwidth]{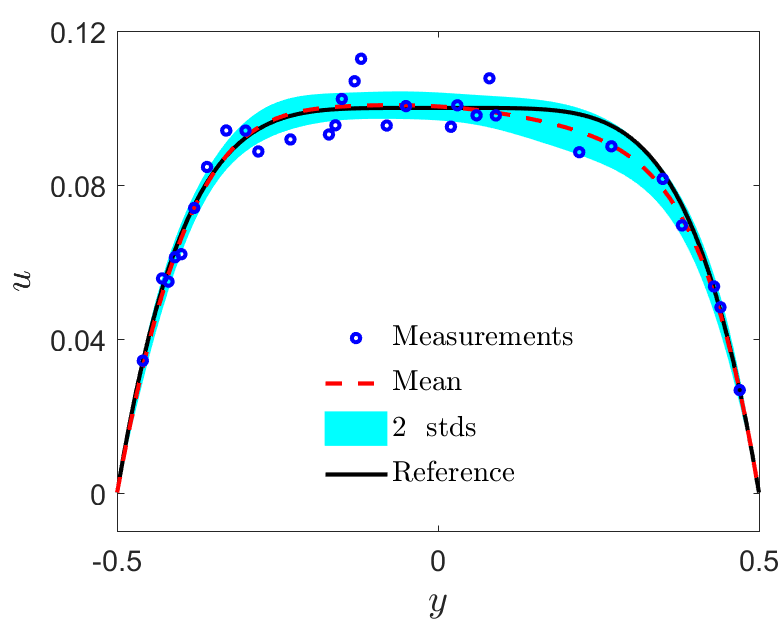}
        \includegraphics[width=0.3\textwidth]{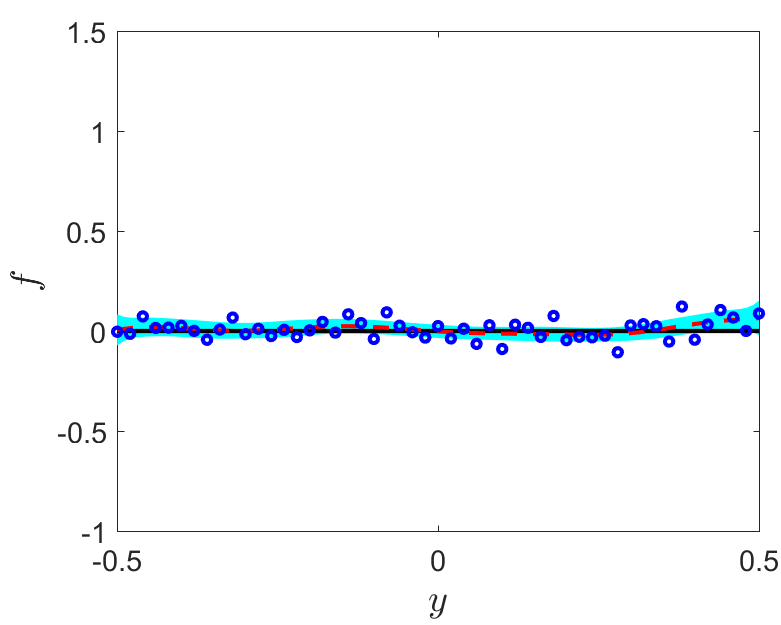}
        \includegraphics[width=0.3\textwidth]{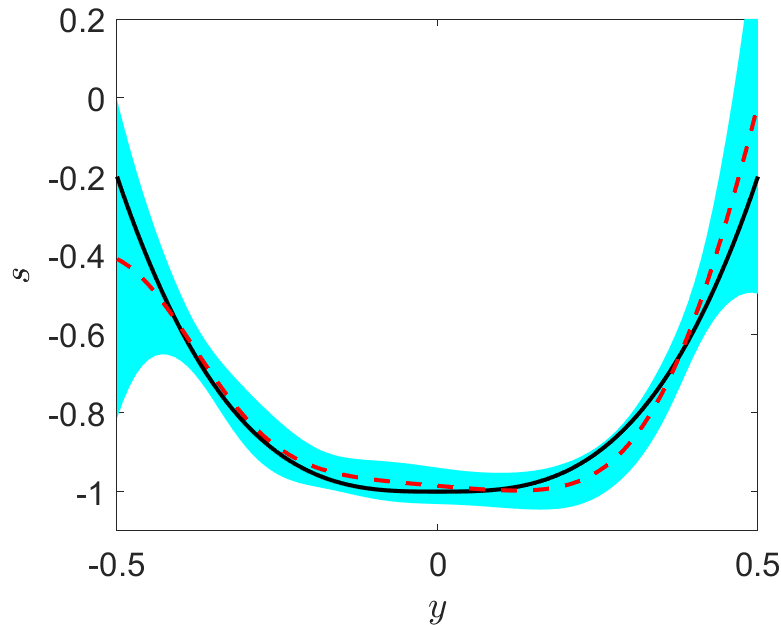}
    }
    \caption{
    2D non-Newtonian channel flow: Predictions for $u$, $f$ and the discrepancy $s$ from B-PINNs with noisy data. B-PINNs with (a) mispecified model, and (b) mispecified model plus a correction. 
    }
    \label{fig:example_3}
\end{figure}

We note that in this specific case we can also employ other approaches to address the issue that we cannot fit the data and satisfy the equation simultaneously in PINNs if the physical model is misspecified. A natural way is to treat the viscosity in Eq. \eqref{eq:channel_flow:2} as an unknown function and employ a DNN to directly learn the constitutive relation. We present the corresponding results in Fig. \ref{fig:example_3_2}(b), in which we employ the same setup, e.g., training data, NN architecture in PINNs, as used in Fig. \ref{fig:1} as well as Fig. \ref{fig:example_3_2}(a). As shown, PINNs with a DNN for learning the viscosity now can fit $u$ and $f$ much better than the results in Fig. \ref{fig:1}. 
We then compare the these results with the ones in Fig. \ref{fig:example_3_2}(a). As shown, the predictions for $u$ are similar, but the PINN with a DNN as correction is able to provide more accurate predictions for $f$, i.e., less fluctuations are observed in Fig. \ref{fig:example_3_2}(a) than in Fig. \ref{fig:example_3_2}(b). Furthermore, we illustrate the loss histories for training PINNs in these two approaches in Fig. \ref{fig:example_3_2}(c). As we can see, the training loss decreases faster and converges to smaller values in the case where we add a DNN as correction to the misspecified model, as compared to the approach in which we model the viscosity as an unknown function.  
We conjecture that the superior performance of the proposed approach can be attributed to the following reason: modeling the viscosity as an unknown function via a DNN requires the accurate estimate of the derivatives in Eq. \eqref{eq:channel_flow:2},  which is quite challenging since $n < 1$ results in very sharp gradients for the viscosity around $y = 0$. However, the correction term, i.e., $s$ in Eq. \eqref{eq:channel_flow:2}, is a quite smooth function, as shown in Fig. \ref{fig:example_3}. Hence, we can expect the easier training as well as better accuracy in the proposed method. 

We now move to the noisy-data case and employ B-PINNs to quantify the model uncertainty. 
The same $30$ noisy measurements of $u$ and $51$ noisy measurements of $f$ are sampled but now they are corrupted by additive Gaussian noise with $0.005$ and $0.05$ noise scales, respectively. 
In B-PINNs, we test two specific cases, i.e., (1) we assume the fluid is Newtonian and encode Eq. \eqref{eq:channel_flow:2} in B-PINNs. The objective is then to infer the unknown viscosity given data, similar as the case in Fig.~\ref{fig:1}; and (2) we are not sure if the fluid is Newtonian or not, we then add a DNN as the correction to the possible misspecified model, i.e., Eq. \eqref{eq:channel_flow:3}.  Also, we note that in this specific case $s$ can be computed analytically by plugging the analytic solution of the power-law fluid into Eq.~\eqref{eq:channel_flow:3}: $s(y) = {4\mu_1}c^4|y|^3/{\mu_0^4} + c$, which serves as the reference in what follows.

The results from B-PINNs are illustrated in Fig. \ref{fig:example_3}. As observed, the NN surrogate $u_\theta$ is not able to fit the data and satisfy the misspecified physical model at the same time (Fig. \ref{fig:example_3}(a)). In addition, with the additional DNN for correcting the misspecified physical model in the previous case, $u_\theta$ can now fit the data and satiafy the corrected physical model simultaneously, as shown in Fig. \ref{fig:example_3}(b). Also, we can see that  the computational errors between the  predicted mean and the reference solution for $s$ are bounded by the predicted uncertainties in the entire computational domain.

\subsubsection{Cavity flow}\label{subsec:cavity_flow}
Next, we consider steady non-Newtonian flow in a two-dimensional lid-driven square cavity (i.e., $x, y \in [0, 1]$), which can be described by the incompressible Navier-Stokes equations as
\begin{subequations}
\begin{align}\label{eq:cavity}
    \nabla \cdot \bm{u} &= 0,\\ 
     \bm{u} \cdot \nabla \bm{u} &= -\nabla P +  \nabla \cdot [\nu (\nabla \bm{u} + (\nabla \bm{u})^T)]  + \bm{f}, 
\end{align}
\end{subequations}
where $\bm{u} = (u, v)$ denotes the velocity in $x-$ and $y-$ directions, respectively; $P$ is the pressure; and $\nu$ is the kinematic viscosity.  In particular, the boundary condition for the upper wall is expressed as
\begin{equation}
     u = U\left(1 - \frac{\cosh[r(x - \frac{L}{2})]}{\cosh(\frac{rL}{2})}\right), ~ v = 0,
\end{equation}
where $U$, $r$ and $L$ are constant. Specifically, $r = 10$, $L = 1$ is the length of the cavity. In addition, the remaining walls are stationary. Here, the viscosity model is the same as the one used in channel flow, i.e., $\mu = \rho \nu = \mu_0 |\bm{S}|^{n-1}$. The objective is to learn the governing equations, which explain the data given measurements on $\bm{u}$, $P$, and $\bm{f}$.
In particular, the training data as well as the reference solution in this case are generated by solving Eq. \eqref{eq:cavity} using the lattice Boltzmann equation model (LBE) in \cite{wang2015localized} with $n = 1.5$. Details for the computations in LBE can be found in \ref{subsec:lbe}.

Similar as in Sec. \ref{subsec:channel_flow}, suppose we misspecify the governing equations as the ones for Newtonian flows, and we then employ the present approach to correct the model misspecification as follows:
\begin{subequations}
\begin{align}
    \nabla \cdot \bm{u} &= 0,  \label{eq:cavity_1a} \\
    \bm{u} \cdot \nabla \bm{u} &= -\nabla P + \nu_0 \nabla^2 \bm{u} + \bm{f} + \bm{s} , \label{eq:cavity_1b}
\end{align}
\end{subequations}
where  $\bm{s} = (s_x, s_y)$ are the two components correcting the momentum equations in $x-$ and $y-$ directions, respectively, and $f_x(x, y) = f_y(x, y) = 0$ for $x, y \in [0, 1]$. The continuity equation Eq. \eqref{eq:cavity_1a} is correctly specified and hence does not need correction.

We assume that we have $400$ random measurements of $u$, $400$ random measurements of $v$, and $81\times81$ uniformly distributed measurements of $P$ from the reference solution.  Here we only consider the case with clean data and the ensemble PINNs are employed for quantifying the model uncertainty. We would like to mention that we do not use B-PINNs here since we need to utilize several hundred training data to achieve meaningful results, which is computationally expensive as the HMC is used for posterior estimation. Similarly, we test two specific cases, i.e., (1) the viscosity in Eq. \eqref{eq:cavity_1b} is an unknown constant and we do not use the correction $\bm{s}$ in PINNs, and (2) we have an initial guess for the viscosity in Eq. \eqref{eq:cavity_1b} and we add $\bm{s}$ in PINNs for the correction for the possible misspecification.  Specifically, we use one DNN to model the velocity $\bm{u}$, one DNN to approximate the pressure $p$, and one DNN to model the discrepancy in Eq.~\eqref{eq:cavity_1b}, i.e. $\bm{s}$. In particular, we set $\nu_0=0.0001$ as the initial guess in Case (2) which will not affect the computational accuracy in PINNs since the model discrepancy is already leveraged by $\bm{s}$.

\begin{table}[ht!]
    \footnotesize
    \centering
    \begin{tabular}{c|c|c|c|c}
    \hline\hline
    & Error of $u$ & Error of $v$ & Error of $f_x$ & Error of $f_y$ \\
    \hline
    Misspecified model & $7.79\%$ & $11.28\%$ & $3.9153\times 10^{-4}$ & $3.1466\times 10^{-4}$\\
    \hline
    Misspecified model with  correction & $1.33\%$ & $2.29\%$ & $7.8389\times 10^{-7}$ & $8.2283\times 10^{-7}$\\
    \hline\hline
    \end{tabular}
    \caption{2D non-Newtonian Cavity flow: Errors of the predicted means from ensemble PINNs. The error of $u$/$v$ is defined as the relative $L_2$ error, while the error of $f_x$/$f_y$ the mean squared error since the reference solutions are $f_x = 0$ and $f_y = 0$.}
    \label{tab:cavity}.
\end{table}

\begin{figure}[ht!]
    \centering
    \subfigure[PINNs with misspecified physical model]{
        \includegraphics[width=0.23\textwidth]{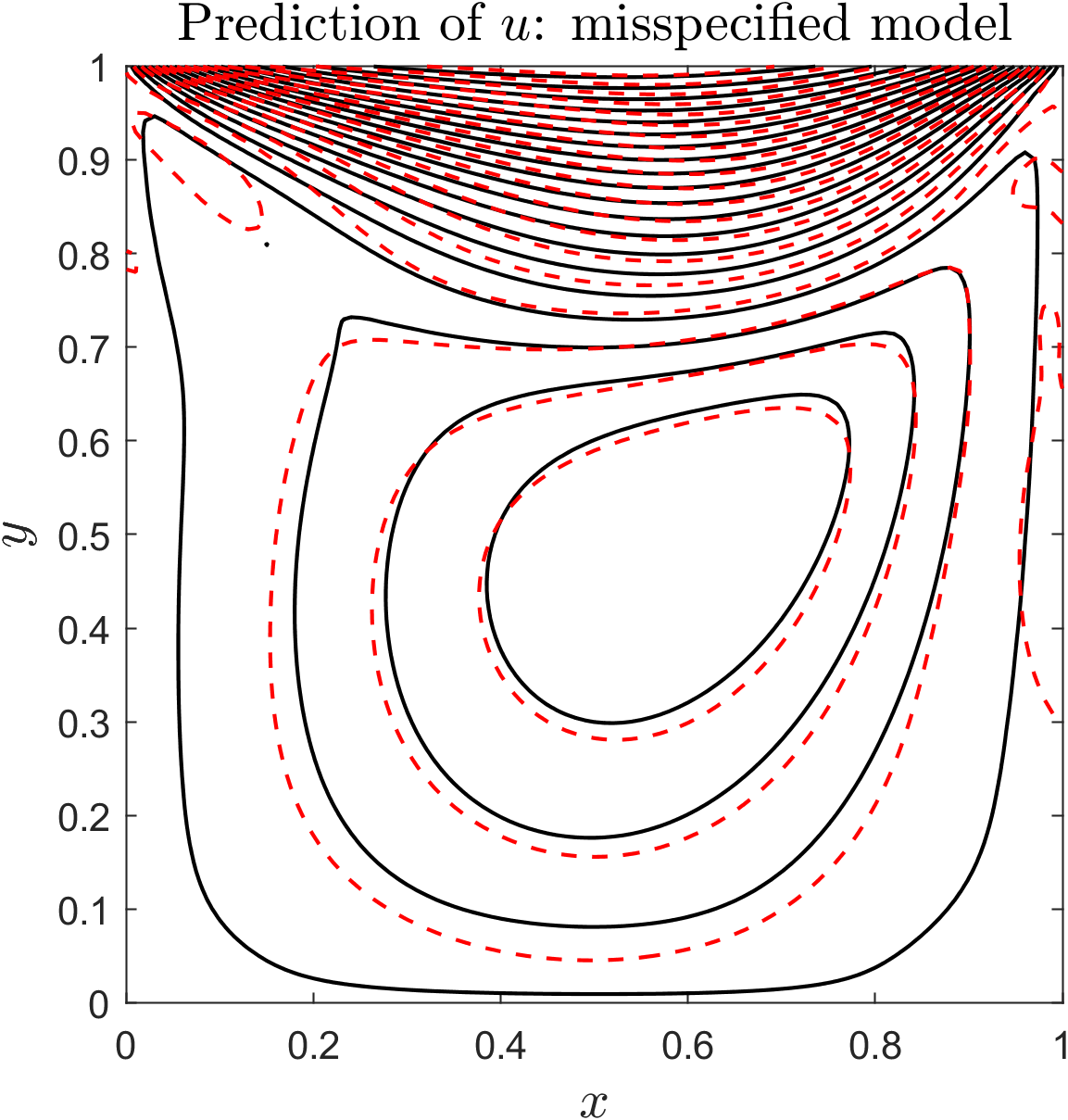}
        \includegraphics[width=0.23\textwidth]{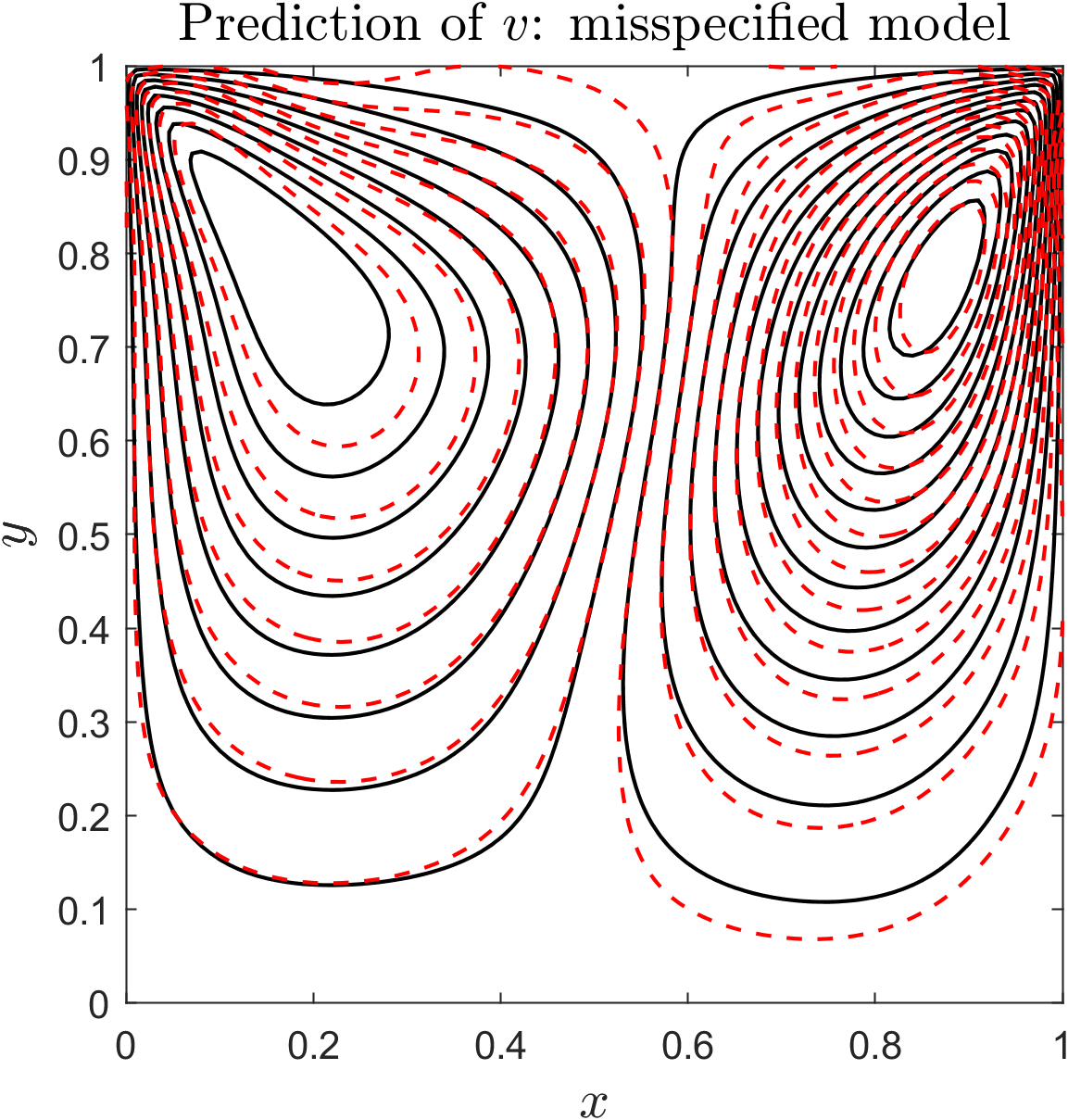}
    }
    \subfigure[Correcting the misspecified physical model.]{
        \includegraphics[width=0.23\textwidth]{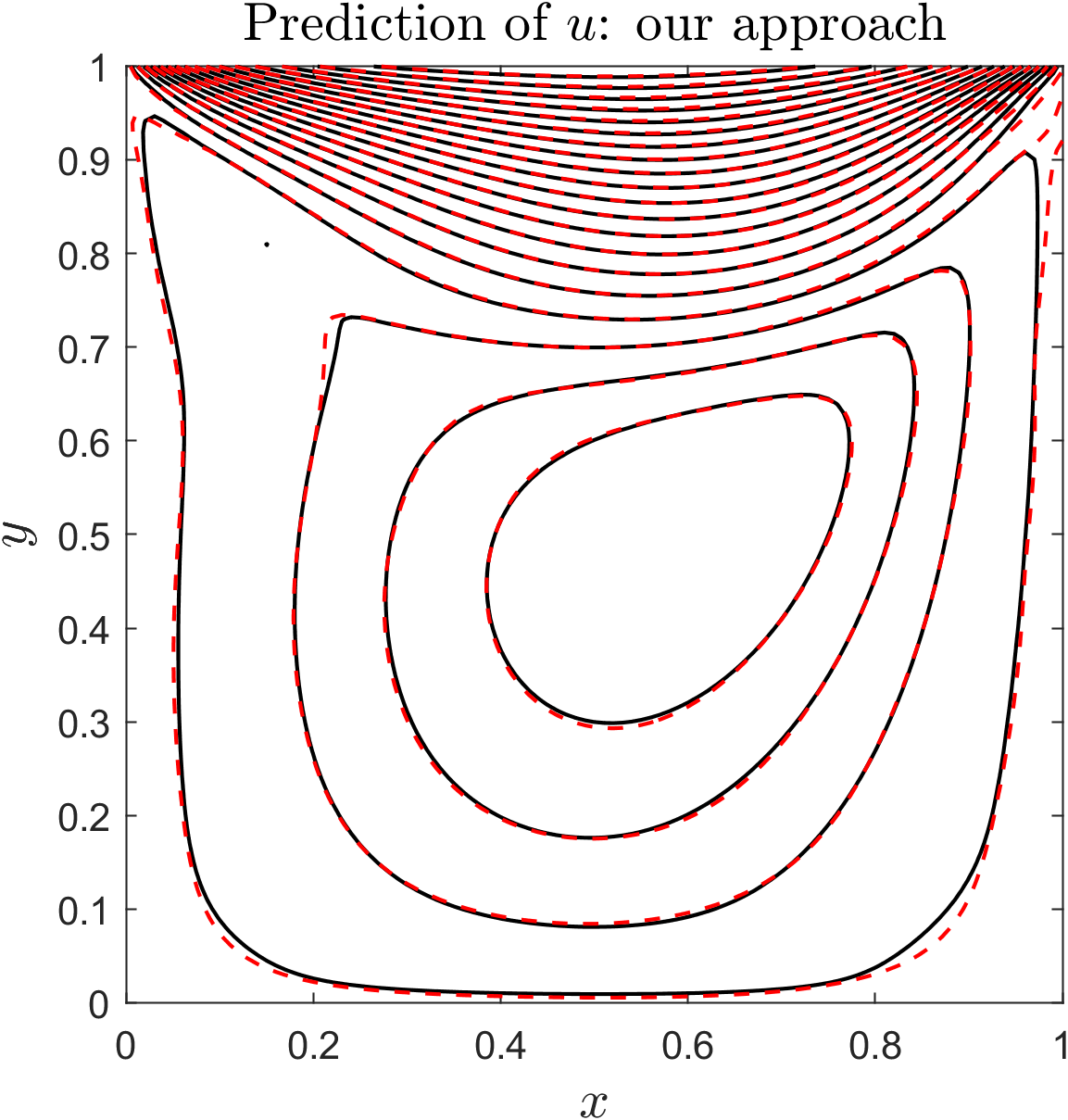}
        \includegraphics[width=0.23\textwidth]{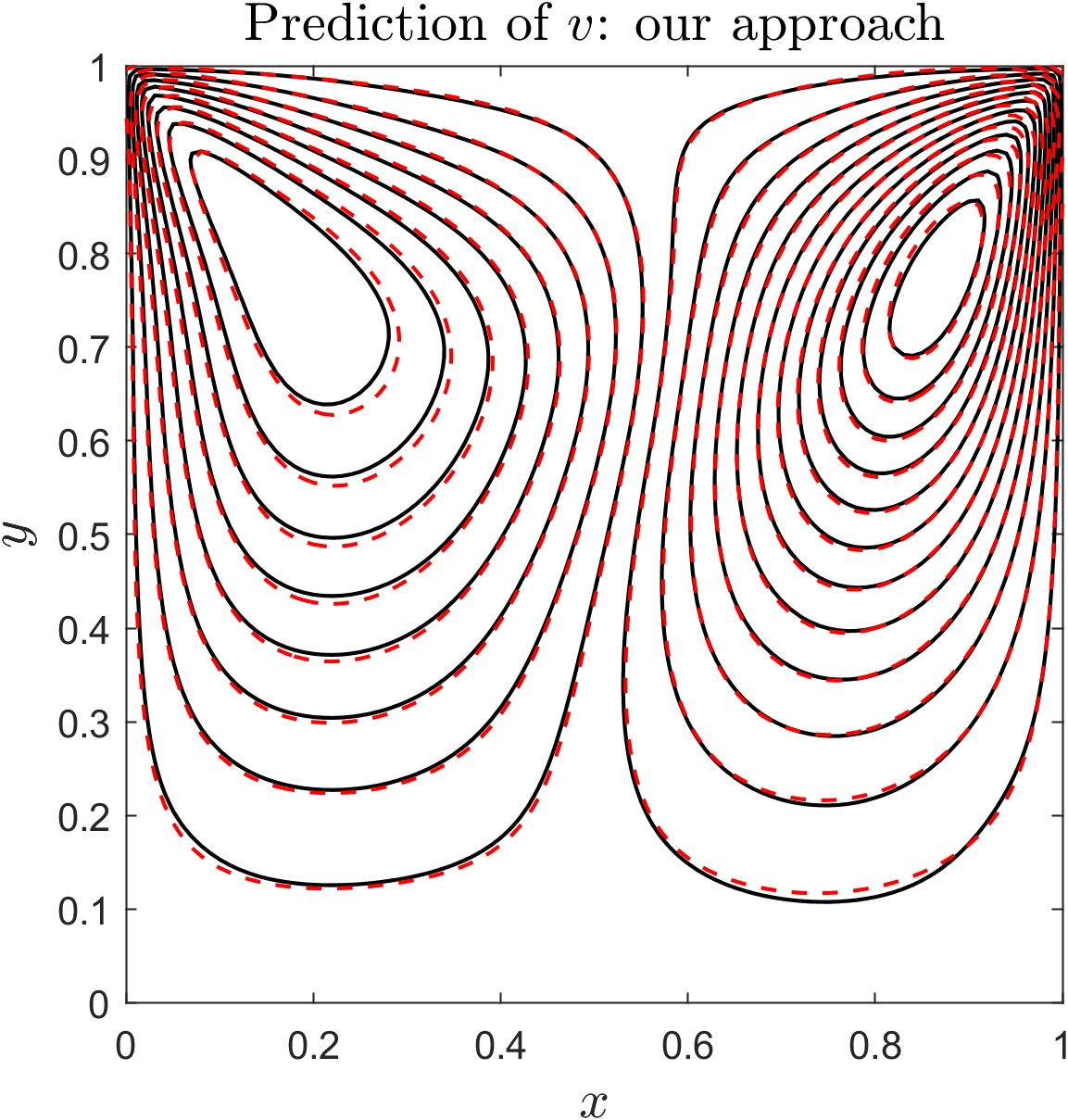}
    }
    \caption{
    Non-Newtonian cavity flow: Predictions from ensemble PINNs for $u$ and $v$ in contours. Red dashed line: Predicted mean; Black solid line: Reference solution.
    }
    \label{fig:example_4_2}
\end{figure}

\begin{figure}[ht!]
    \centering
    \subfigure[PINNs with misspecified physical model]{
        \includegraphics[width=0.23\textwidth]{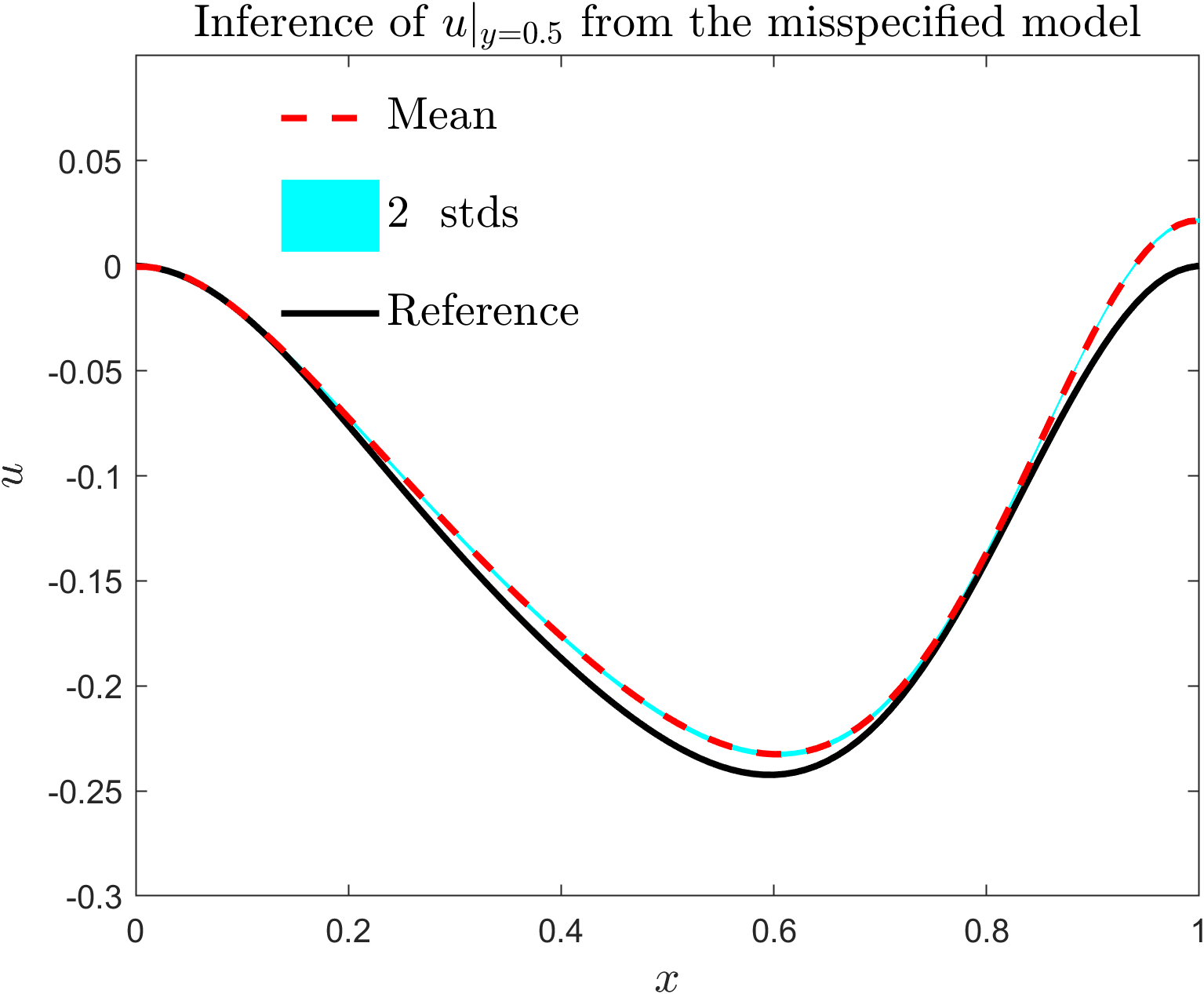}
        \includegraphics[width=0.23\textwidth]{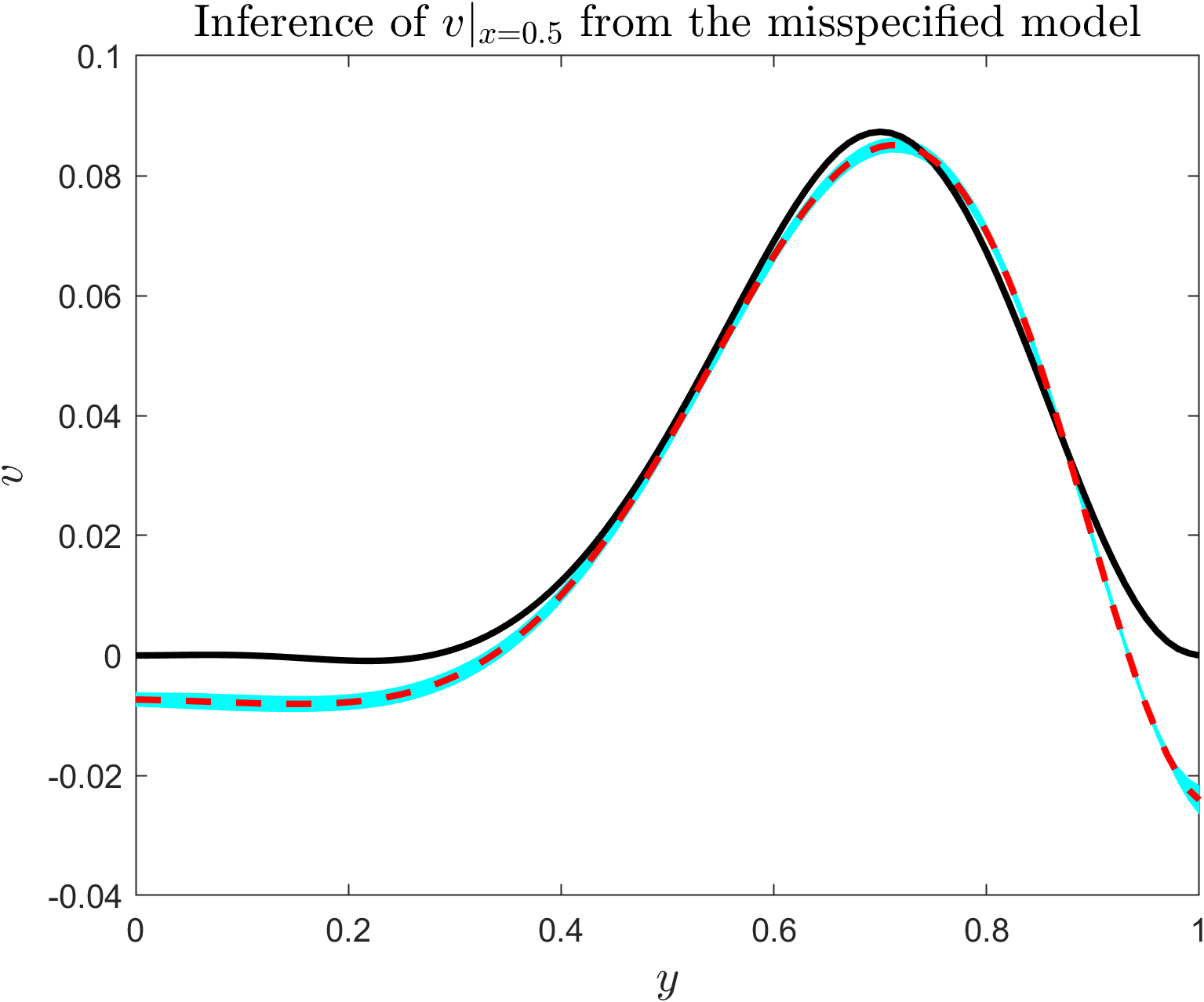}
        \includegraphics[width=0.23\textwidth]{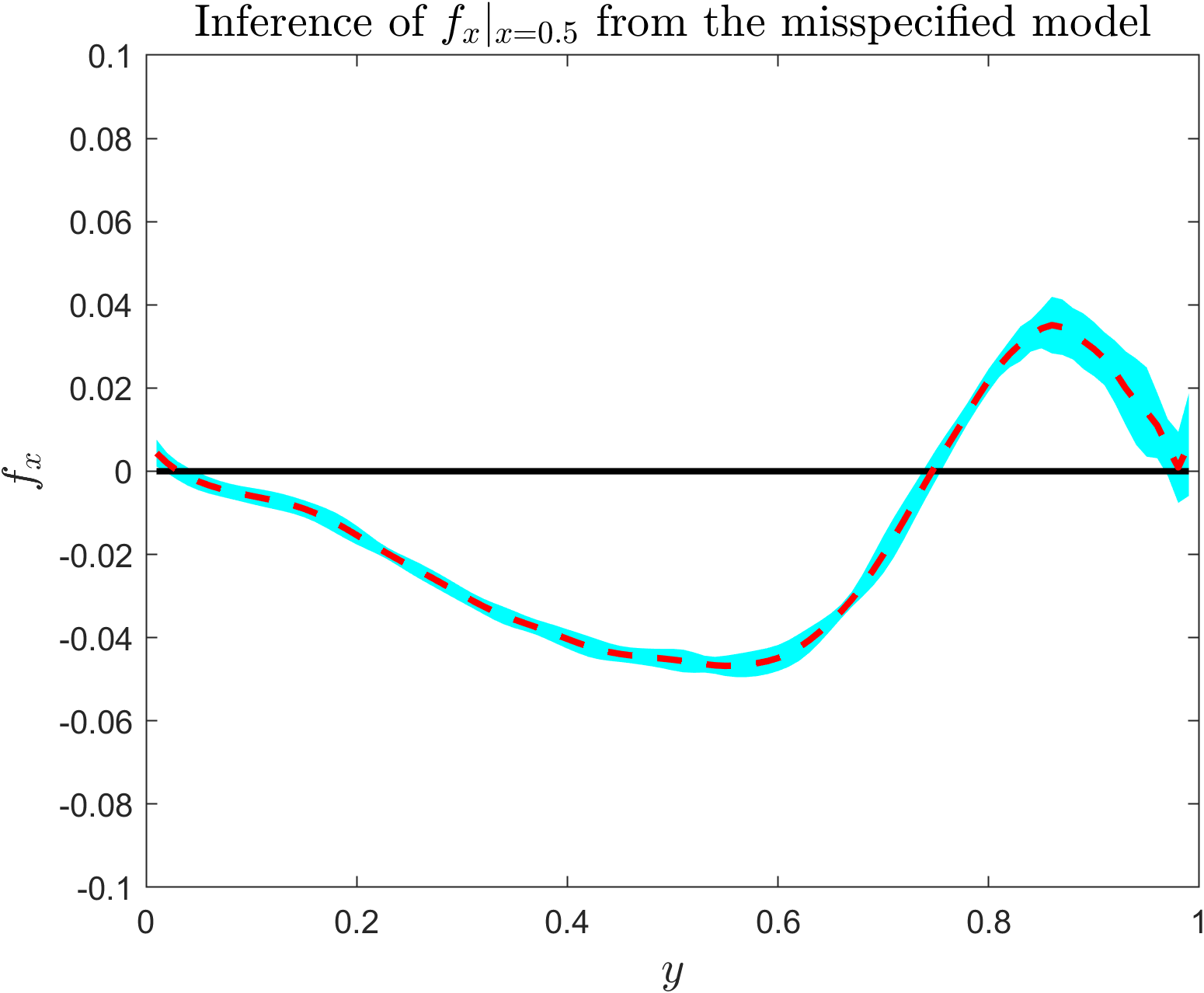}
        \includegraphics[width=0.23\textwidth]{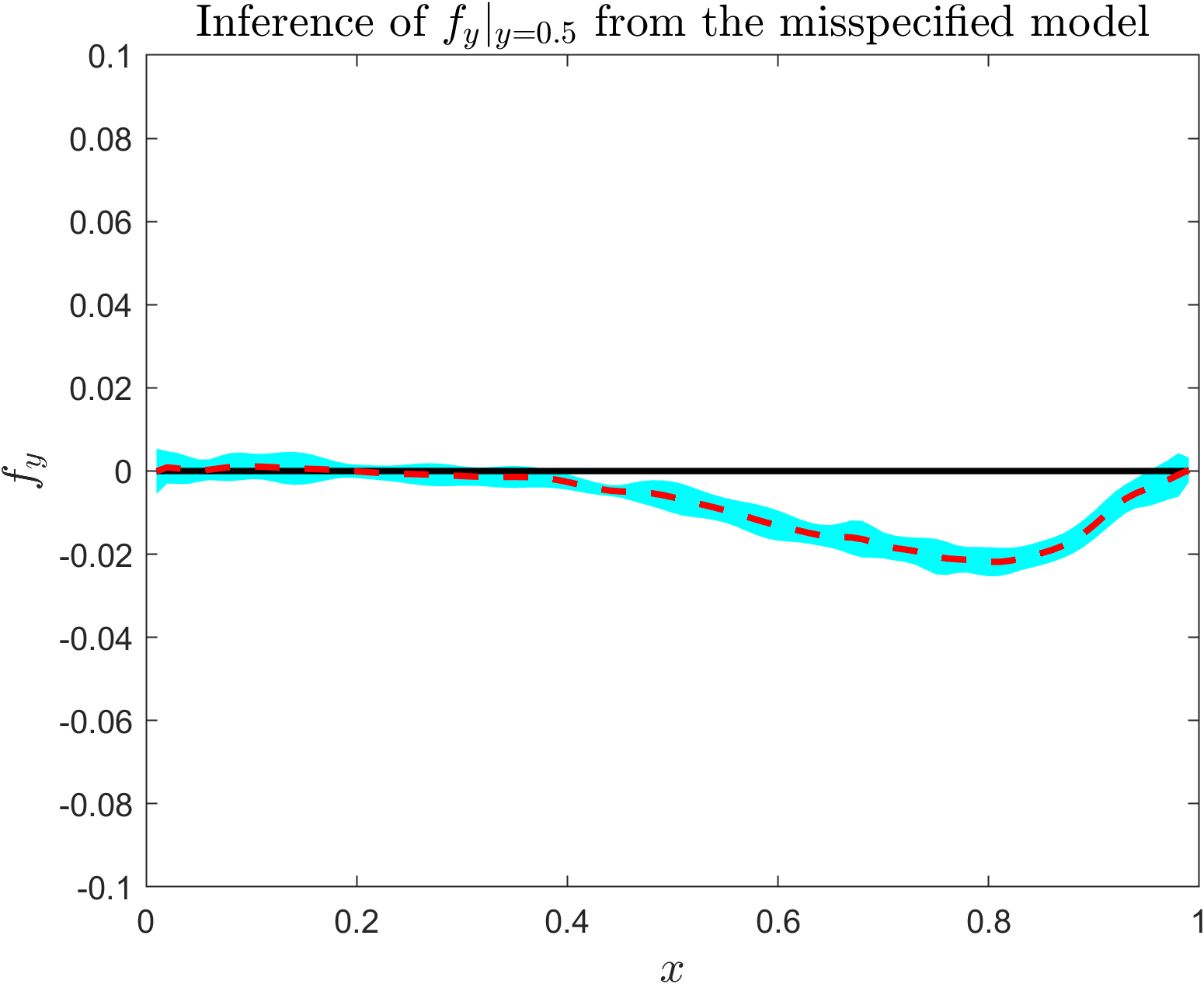}
    }
    \subfigure[Correcting the misspecified physical model with a DNN in PINNs]{
        \includegraphics[width=0.23\textwidth]{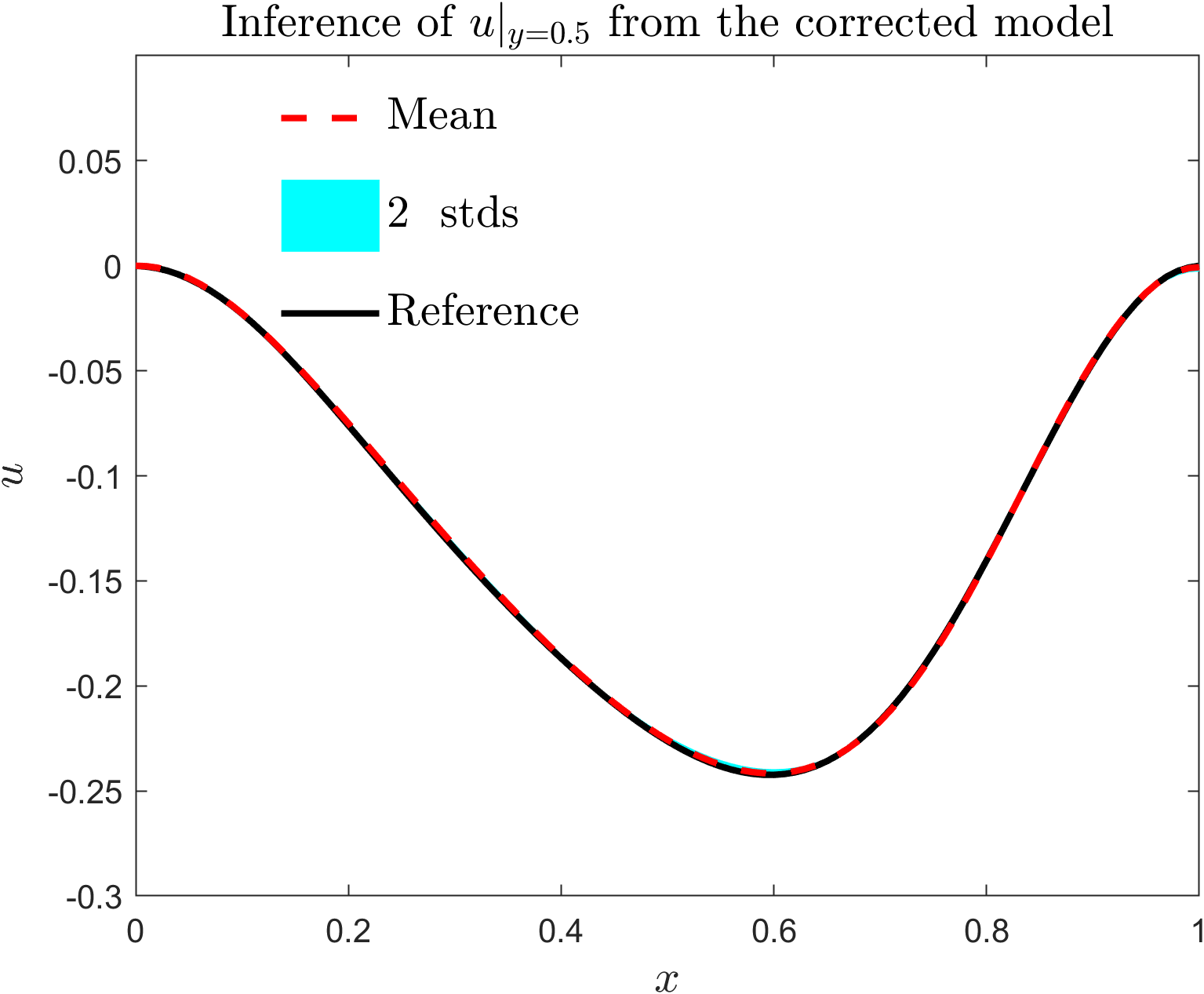}
        \includegraphics[width=0.23\textwidth]{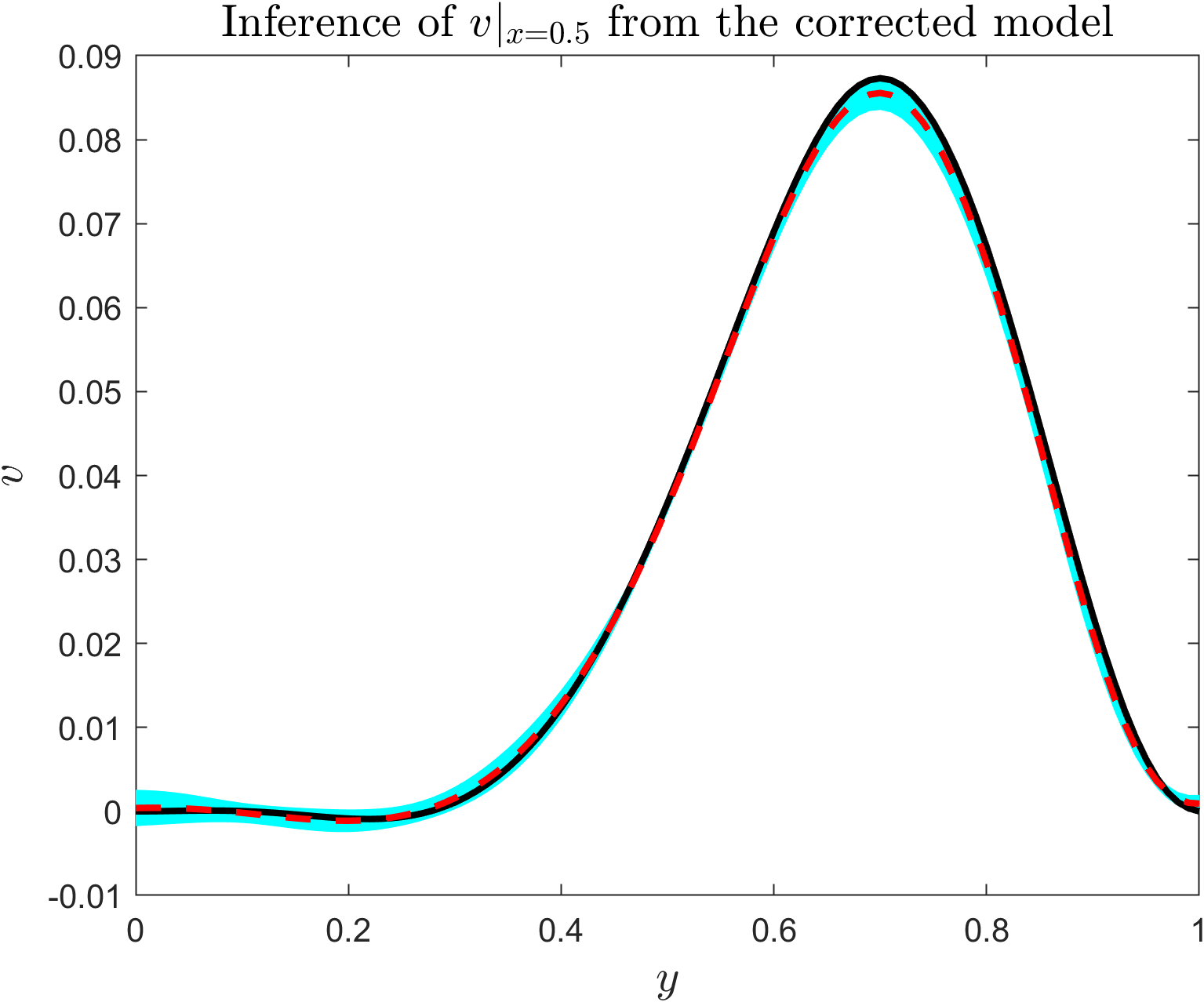}
        \includegraphics[width=0.23\textwidth]{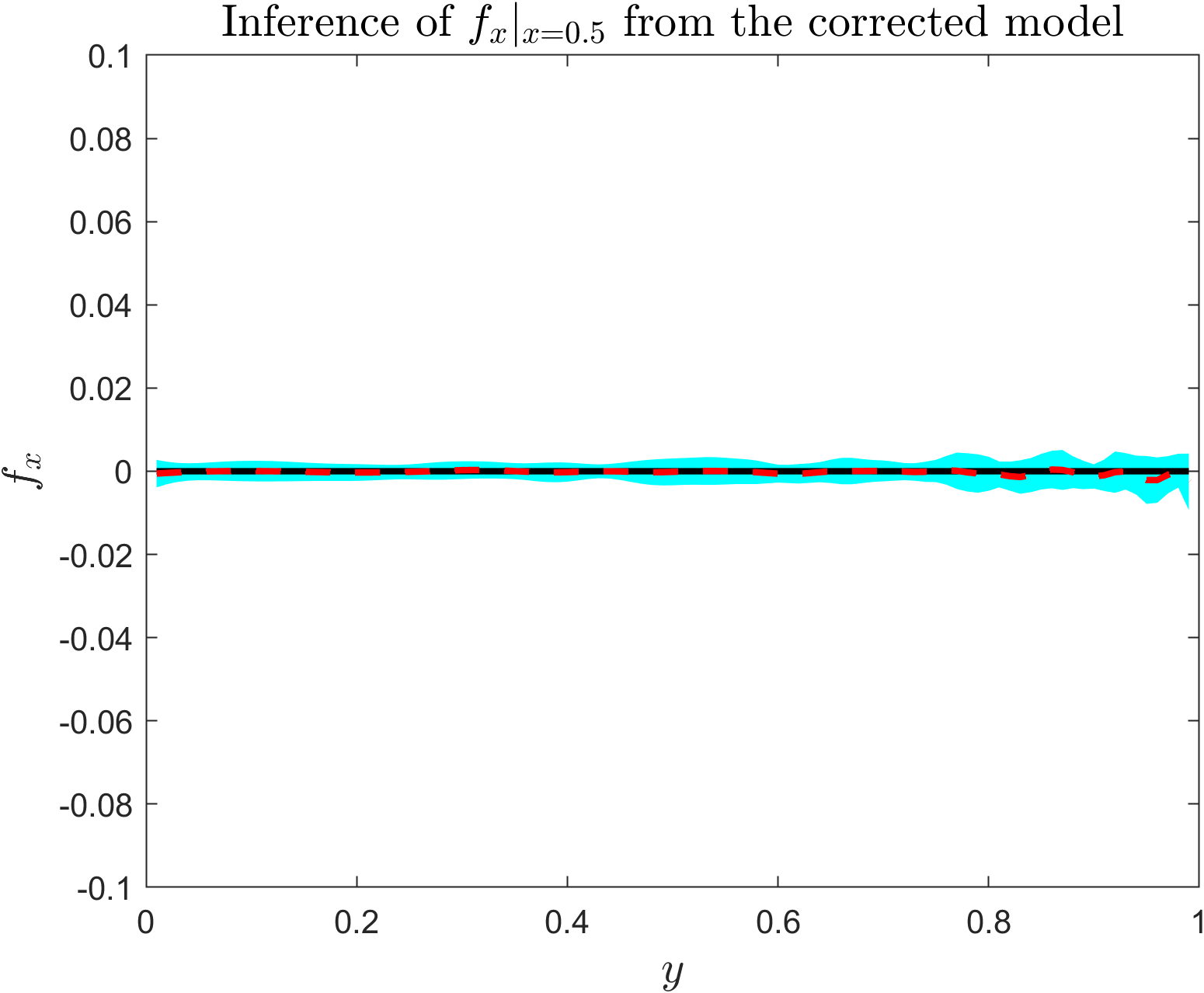}
        \includegraphics[width=0.23\textwidth]{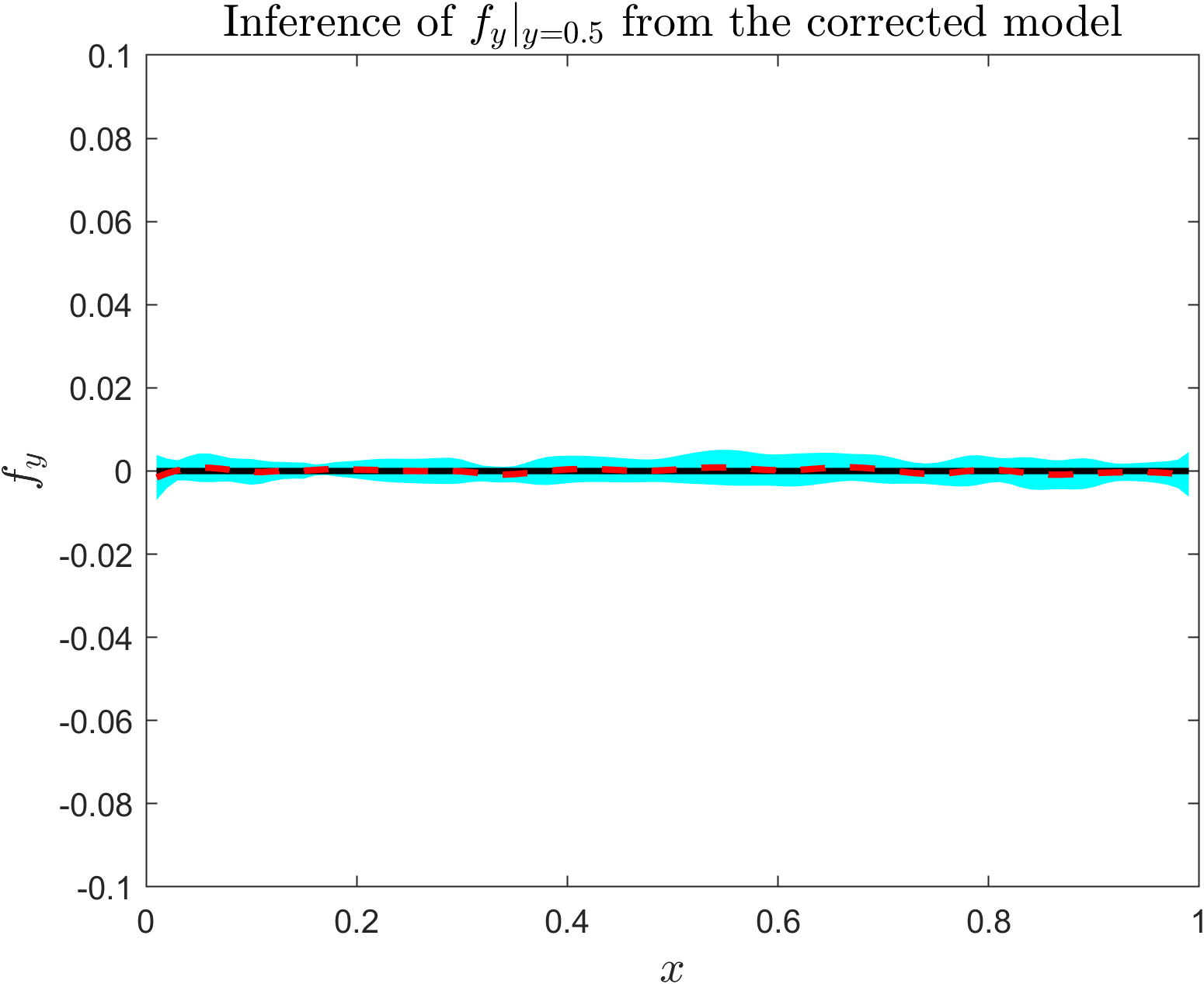}
    }
    \caption{
    Non-Newtonian cavity flow: Predictions for $u, v, f_x, f_y$ in 1D slices. Red dashed line: predicted mean; black solid line: reference solution; uncertainty bound: predicted uncertainty.
    }
    \label{fig:example_4_1}
\end{figure}

We present the predictions for $\bm{u}$ in Fig.~\ref{fig:example_4_2}. 
As we can see, PINNs with misspecified model lead to inaccurate inference for both $u$ and $v$ (Fig. \ref{fig:example_4_2}(a)). The proposed approach, however, achieves much better results by adding a DNN to correct the model misspecification. We further illustrate the computational errors for both cases in Table ~\ref{tab:cavity}, which clearly shows the effectiveness as well as superiority of the present method. In addition,  we show some representative 1D slices for the predicted $u, v, f_x$, and $f_y$ with uncertainties in Fig.~\ref{fig:example_4_1}. Similarly,  the PINNs with misspecified model cannot fit the data and satisfy the physical model at the same time (Fig.~\ref{fig:example_4_1}(a)), while this issue can be handled well as we add a DNN to correct the misspecified model, as shown in Fig.~\ref{fig:example_4_1}(b).

\section{Summary}\label{sec:4}
We presented a general approach to improve the accuracy of physics-informed neural networks (PINNs) for discovering governing equations from data when the physical model is misspecified. In particular, we first assume the mathematical models for the specific problem we considered and encode them in PINNs via automatic differentiation. We note that the assumed  models may be misspecified in certain complex systems since not all the physical processes are fully understood here. We then utilize another deep neural networks (DNN) to model the discrepancy between the imperfect models and the observational data in order to correct the misspecified model. In addition, to quantify uncertainties arising from noisy and/or gappy data, we employ two typical uncertainty-induced PINNs methods, i.e., the Bayesian physics-informed neural networks (B-PINNs) and ensemble PINNs. A series of numerical examples are conducted to show the effectiveness of the proposed method. Specifically, we considered (1) a reaction-diffusion system where we misspecified the reaction model, and (2) non-Newtonian flows in two-dimensional channel and cavity but misspecified the fluid viscosity. The results show that the present approach can achieve good accuracy since the added DNN is able to correct the misspecified physical models encoded in PINNs. Also, the B-PINNs and/or ensemble PINNs can provide reasonable uncertainty bounds for the discovered  physical models, which reflects the uncertainties in physical models, i.e., \emph{model uncertainty}. Furthermore, we showcase that we can seamlessly combine the proposed method with the symbolic regression to obtain explicit governing equations based on the trained NNs. Finally, we expect the proposed approach to be a promising tool in a wider class of applications for solving problems where the physical processes are not fully understood.

\section*{Acknowledgement}
This work was supported by the MURI/AFOSR project (FA9550-20-1-0358), the DOE-MMICS SEA-CROGS project (DE-SC0023191), the NIH grant Neural Operator Learning to Predict Aneurysmal Growth and Outcomes (R01HL168473), and the NIH grant CRCNS: Waste-clearance flows in the brain inferred using physics-informed neural networks (R01AT012312).
Z.Z. thanks Dr. Khemraj Shukla from Brown University for helpful discussion.

\bibliographystyle{elsarticle-num} 
\bibliography{references}

\appendix
\appendix

\section{Details of hyperparameters in numerical experiments}\label{subsec:hyperparameter}

In Sec.~\ref{subsec:ode}, DNNs used to model $u$/$s$ have two hidden layers, each of which has $50$ neurons and is equipped with hyperbolic tangent as the activation function, and take as input $t$ and output $u$/$s$. In Sec.~\ref{subsec:reaction_diffusion}, DNNs have two hidden layers and each layer has $50$ neurons and a hyperbolic tangent activation function. They take as input $x, t$ and output $u$ or $s$. In Sec.~\ref{subsec:channel_flow}, DNNs have two hidden layers ($50$ neurons and a hyperbolic tangent activation function) and take as input $y$ and output $u$ or $s$. 
In Sec.~\ref{subsec:cavity_flow}, a DNN with three hidden layers ($100$ neurons and a hyperbolic tangent activation function) is used to model $u, v$. That is, this DNN takes as input $x, y$ and outputs $u, v$. Another DNN with three hidden layers ($100$ neurons and a hyperbolic tangent activation function) is used to model $p$. Besides, an additional DNN with three hidden layers ($100$ neurons and a hyperbolic tangent activation function) is used to correct two misspecified equations and it takes as input $x, y$ and outputs $s_2, s_3$.

The ensemble PINN method \cite{psaros2023uncertainty, zou2022neuraluq} is used for clean data to obtain predictions with UQ. In Sec.~\ref{subsec:ode}, $20$ PINNs are independently trained with Adam optimizer  \cite{kingma2014adam} ($10^{-3}$ learning rate) for $100,000$ iterations. In Sec.~\ref{subsec:cavity_flow}, $10$ PINNs are independently trained with Adam optimizer ($10^{-3}$ learning rate) for $200,000$ iterations. To obtain results shown in Fig.~\ref{fig:example_3_2}, PINNs are trained with Adam optimizer for $500,000$ iterations. The learning rate decays from $10^{-3}$ to $10^{-5}$ for stable training. 

The B-PINN method \cite{yang2021b} is used for noisy data to obtain predictions with UQ. In all related examples, parameters of DNNs are assumed to be identically independent distributed (i.i.d.) Gaussian as the prior distribution with mean zero and standard deviation one. The likelihood distribution is also assumed to be independent Gaussian where the standard deviation is the known noise scale for additive Gaussian noise.
In addition, Hamiltonian Monte Carlo (HMC) with adaptive step size \cite{neal2011mcmc} is used to obtain posterior samples, in which the number of posterior samples is set to $1,000$ and the number of steps for the leap-frog scheme is set to $50$.
The number of burn-in samples and the initial step are tuned such that the acceptance rate lies between $50\%$ and $70\%$. A open-sourced Python library for UQ in scientific machine learning, termed NeuralUQ \cite{zou2022neuraluq}, is used for fast and convenient implementations.

\section{Additional results for the ODE system}\label{sec:appendix_ode}

\subsection{Clean and gappy data}\label{sec:appendix_ode:1}

\begin{figure}[ht!]
    \centering
    \subfigure[PINN with correct physical model.]{
        \includegraphics[width=0.3\textwidth]{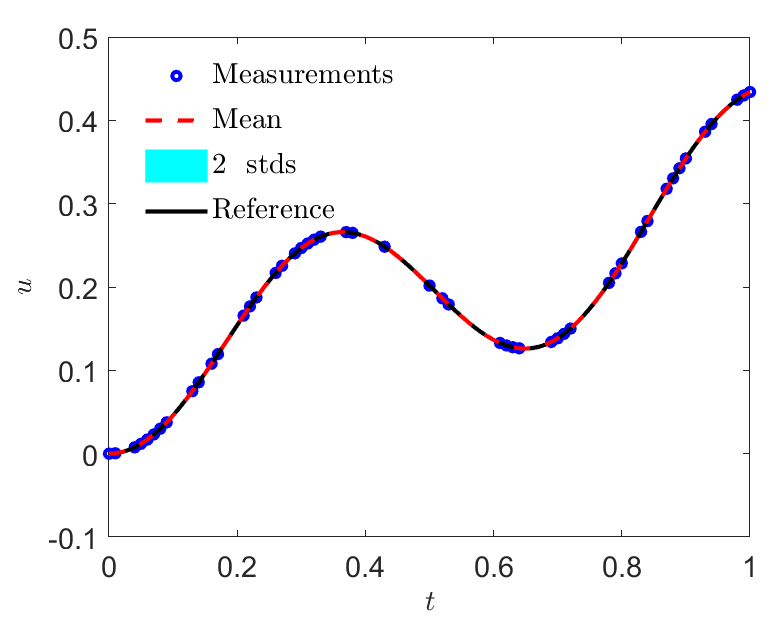}
        \includegraphics[width=0.3\textwidth]{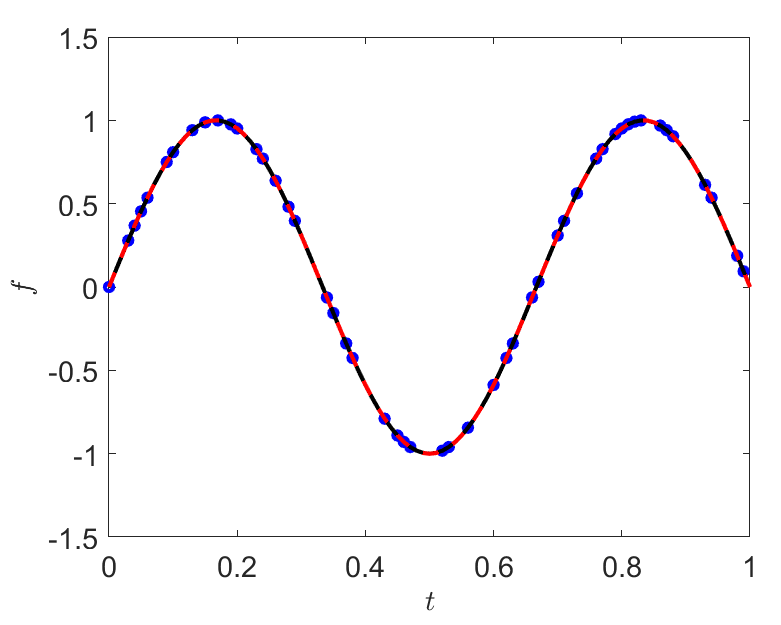}
        \includegraphics[width=0.3\textwidth]{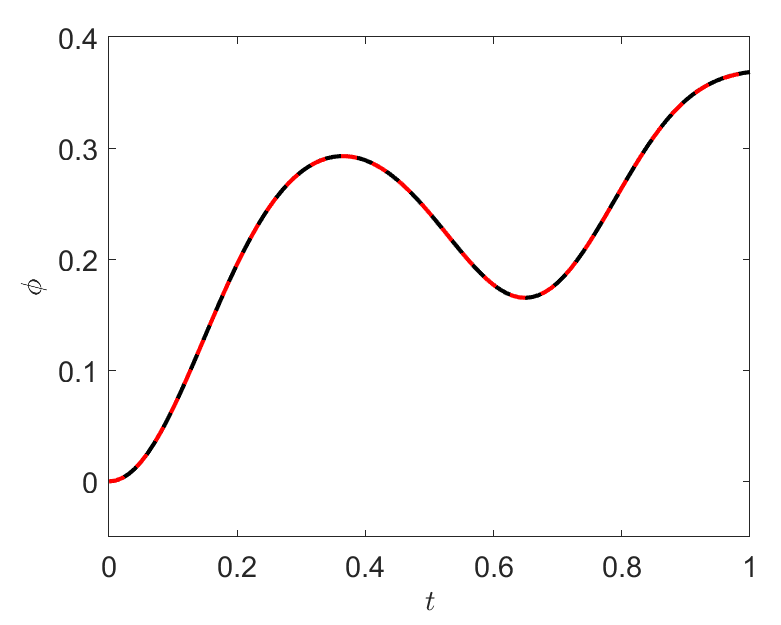}
    }
    \subfigure[PINN with misspecified physical model.]{
        \includegraphics[width=0.3\textwidth]{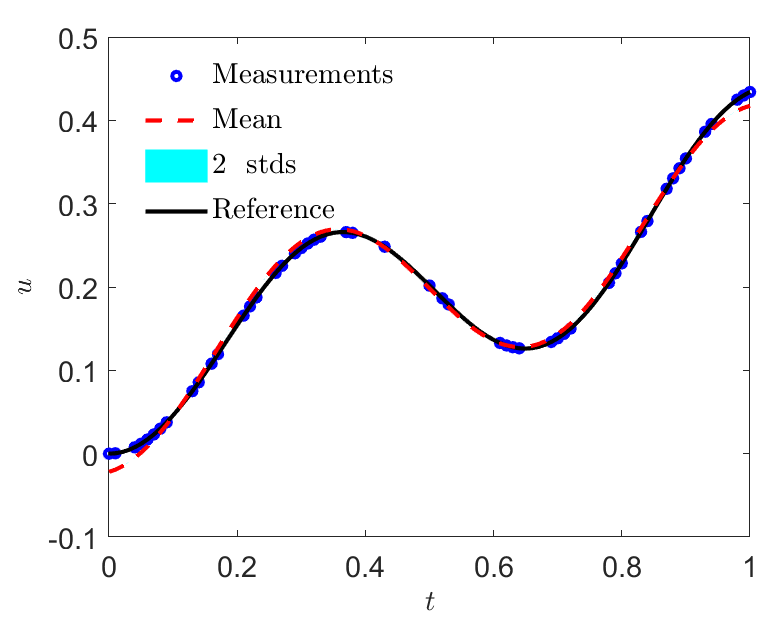}
        \includegraphics[width=0.3\textwidth]{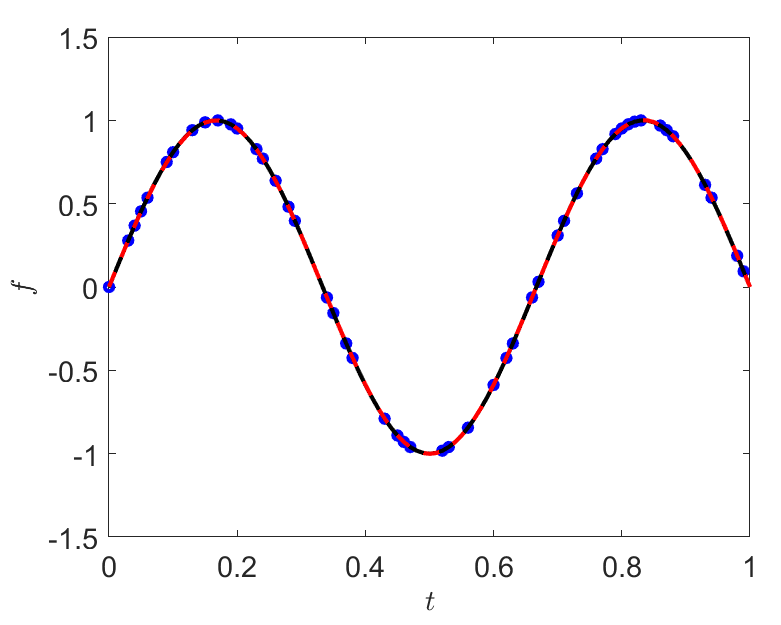}
        \includegraphics[width=0.3\textwidth]{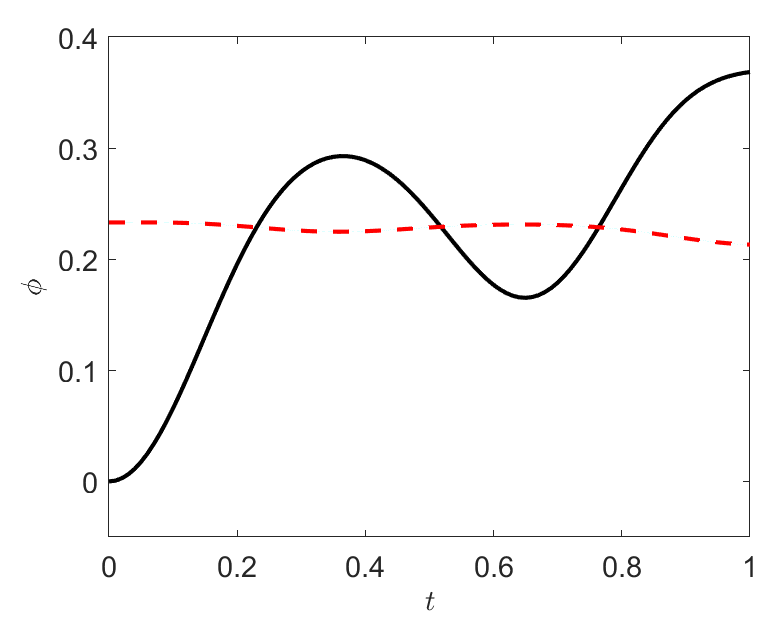}
    }
    \subfigure[PINN with misspecified physical model plus a correction.]{
        \includegraphics[width=0.3\textwidth]{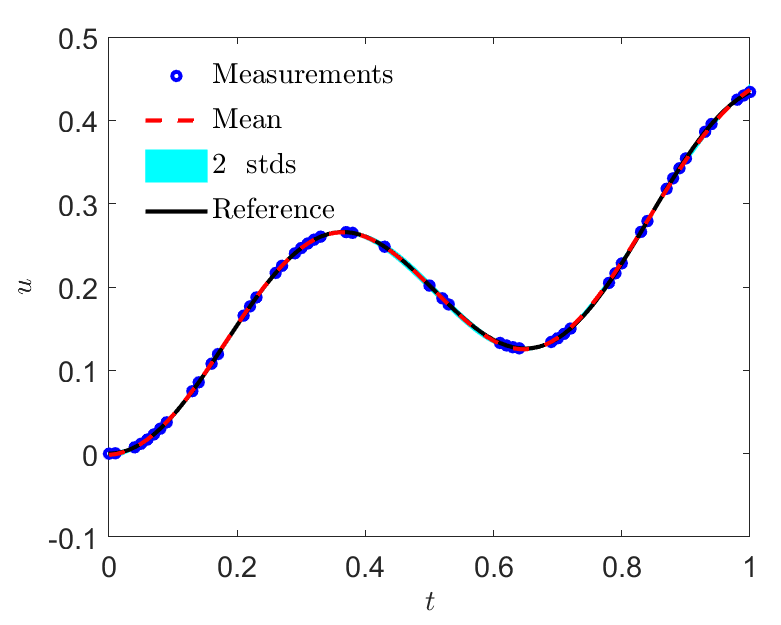}
        \includegraphics[width=0.3\textwidth]{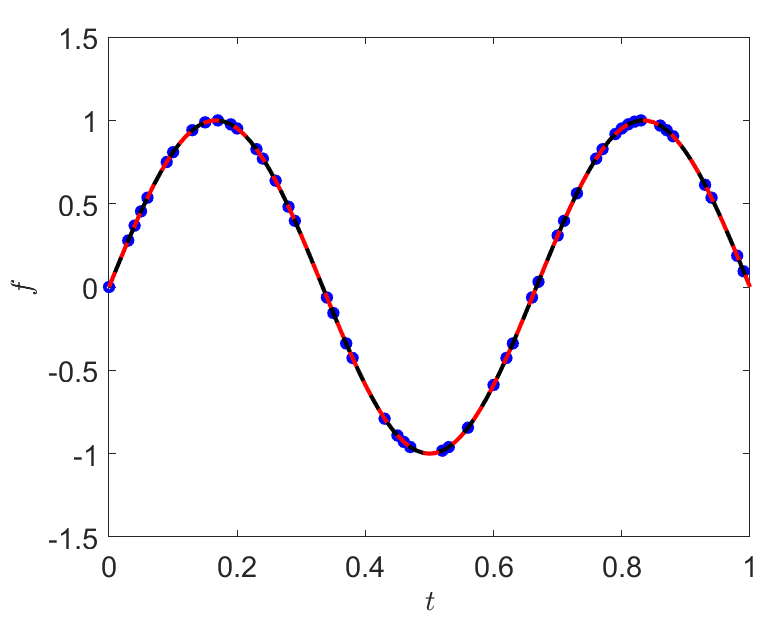}
        \includegraphics[width=0.3\textwidth]{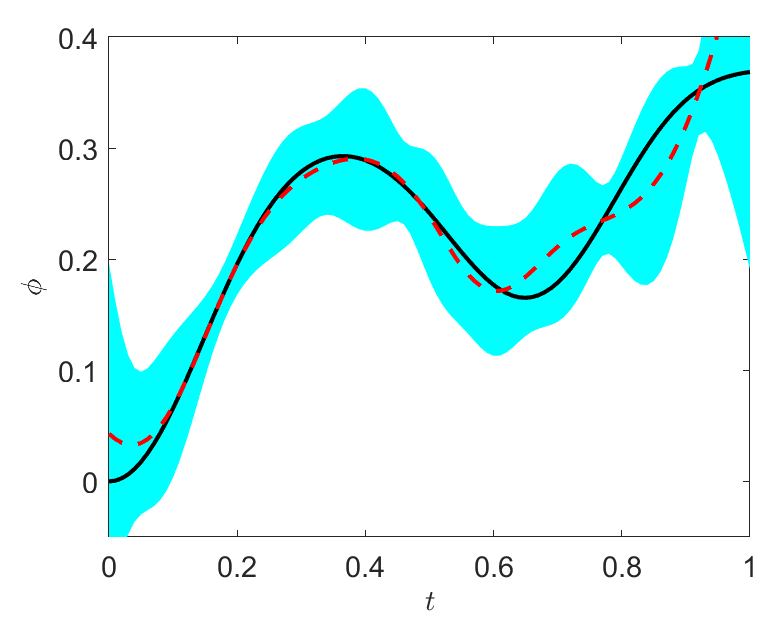}
    }
    \caption{
    ODE system: Predictions for $u$, $f$ and $\phi$ from ensemble PINNs with clean and gappy data. Ensemble PINNs with (a) the correct physical model, (b) a misspecified physical model, and (c) the misspecified physical model plus a correction. Blue circles: Representative measurements for $u$ and $f$; Red dashed line: Predicted mean; Black solid line: Reference solution.
    }
    \label{fig:appendix_1}
\end{figure}

\begin{table}[ht]
    \footnotesize
    \centering
    \begin{tabular}{c|c|c|c|c|c}
    \hline\hline
    & $\tilde{\lambda}$ & Error of $\phi$ & Error of $u$ & Error of $f$ & Error of $\tilde{u}$\\
    \hline
    Case (A): Known model & $1.4997\pm 0.0003$ & $0.02\%$& $0.01\%$ & $0.09\%$ & $0.03\%$\\
    \hline
    Case (B): Misspecified model & $0.2331\pm0.0000$ & $41.59\%$ & $3.14\%$ & $0.16\%$ & $10.20\%$\\
    \hline
    Case (C): Corrected model & $0.2$ & $12.03\%$ & $0.43\%$ & $0.04\%$ & $0.75\%$\\
    \hline\hline
    \end{tabular}
    \caption{Results of PINNs for the ODE system with clean and gappy data; see the caption of Table \ref{tab:example_1_1} for details.}
    \label{tab:appendix}
\end{table}

In Sec. \ref{subsec:ode}, we considered both clean and noisy data cases. In clean data cases, we used sufficient clean data to demonstrate the effect of model misspecifications. Specifically, data are collected from values of $u$ and $f$ on $101$ sampling points uniformly distributed on $t\in[0, 1]$. An alternative approach to estimate the reaction model $\phi$ is using finite difference scheme to approximate ${du}/{dt}$ and then compute $\phi$ as $\phi = {du}/{dt} - f$. The shortcoming of this approach is that it is not applicable to more general dataset, e.g. data of $u$ and $f$ are not sampled on uniform grids and/or are not sampled from the same sensors.
To showcase the generality of the proposed method, here we further consider a case where the data is clean but randomly sampled from $t\in[0, 1]$.

As shown in Fig.~\ref{fig:appendix_1}(a) and Table \ref{tab:appendix}, PINNs with correct physical model (Case (A)) yields the best results in predicting $\phi$ and identifying the governing equations. The results of PINNs with misspecified physical model (Case (B)) and our approach deliver consistent results with all the other cases in Sec.\ref{subsec:ode}.
As illustrated in Fig.~\ref{fig:appendix_1}(c) as well as Table~\ref{tab:appendix}, our approach produces accurate inference of $\phi$ and the surrogate model $u_\theta$ is able to satisfy the ODE and fit the data at the same time, indicating that the generality of the present approach in handling model misspecification with clean but gappy data. We also notice that, compared to the case where data are clean and sufficient, here the predicted uncertainty and the error of $\phi$ from our approach are much larger (shown in Fig.~\ref{fig:appendix_1}(c) and Table \ref{tab:appendix}). We note that when compared with Case (A) where the physical model is known with the same dataset, the increase of the predicted uncertainty and the error is caused by the model uncertainty; when compared with the clean-and-sufficient-data case, it is instead caused by the gappy data.

\subsection{Symbolic regression}\label{sec:appendix_ode:2}

\begin{figure}[ht!]
    \centering
    \subfigure[Clean and sufficient data]{\includegraphics[width=0.3\textwidth]{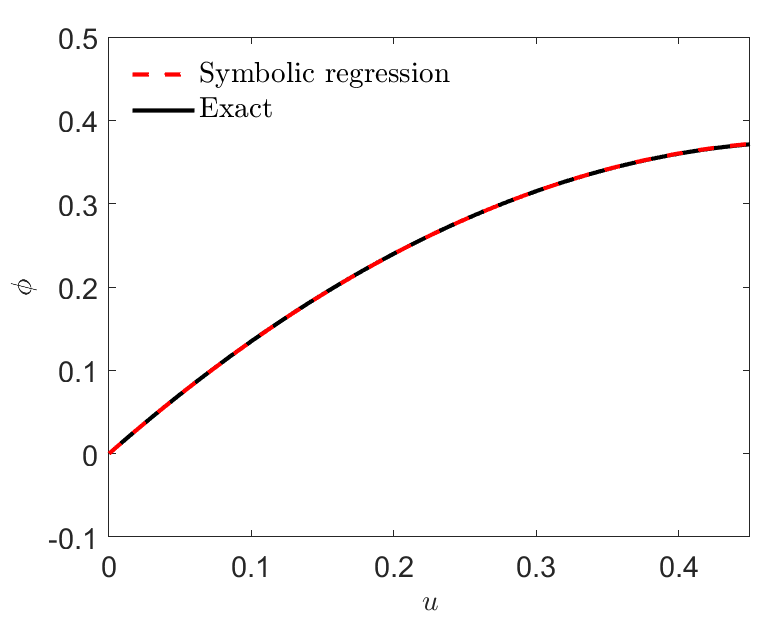}}
    \subfigure[Noisy and gappy data]{\includegraphics[width=0.3\textwidth]{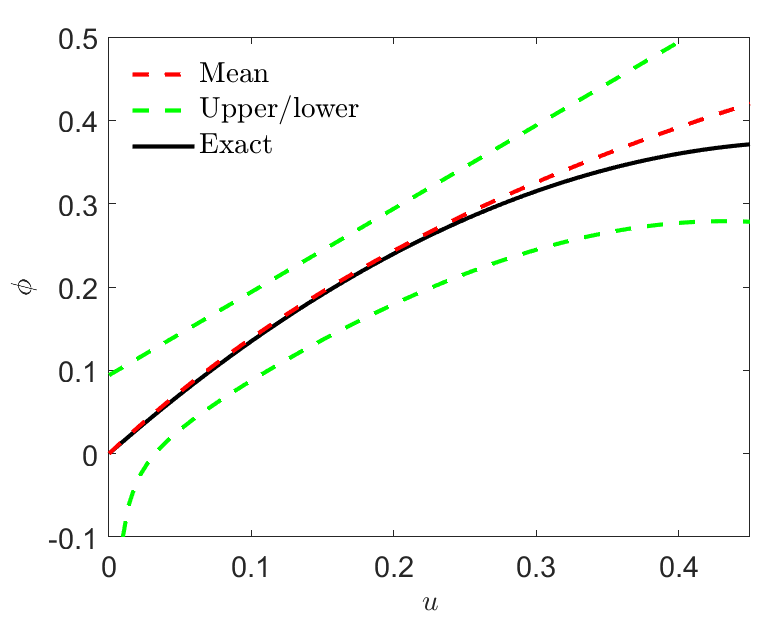}}
    \subfigure[Noisy and sufficient data]{\includegraphics[width=0.3\textwidth]{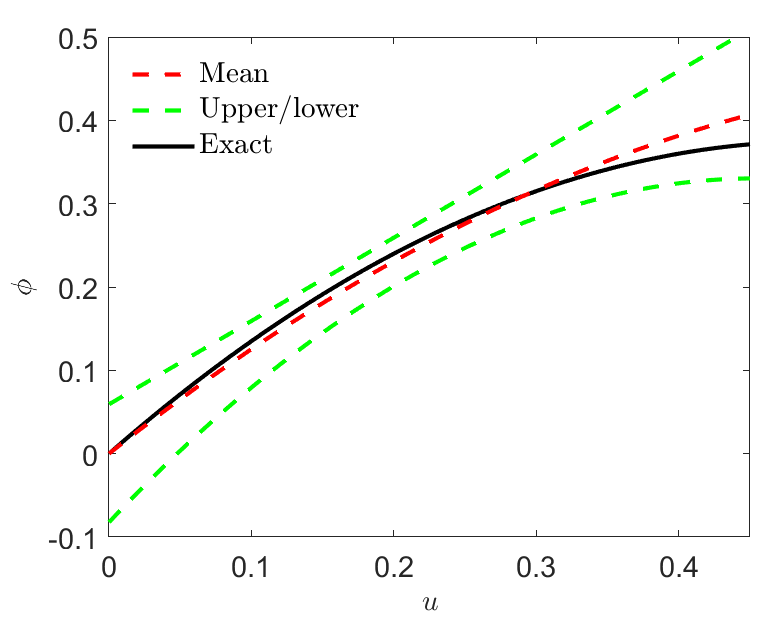}}
    \caption{
    ODE system: Plots of the discovered equation $\phi(u)$ using symbolic regression, following our approach in correcting the model misspecification; see Table~\ref{tab:example_1_3} for the mathematical expressions. In (a), the symbolic regression is performed on the predicted mean from ensemble PINNs. In (b) and (c), the symbolic regression is performed on the predicted mean as well as the upper and lower bounds of the uncertainty band (mean$\pm2~\text{stds}$) from B-PINNs. We note that the symbolic regression is performed on $\phi(u)$ instead of $\phi(t)$.
    }
    \label{fig:appendix_2}
\end{figure}

In Fig.~\ref{fig:appendix_2}, we present plots of the discovered equation $\phi(u)$ using symbolic regression in various cases; see Sec.~\ref{subsec:ode} for details and Table~\ref{tab:example_1_3} for the discovered mathematical expressions. Recall that in noisy-data cases, we employed B-PINNs and obtained uncertainties in predicting $u$ and $\phi$. By forming the data for symbolic regression as $\{u_\theta(t_i), s_\psi(t_i) + \tilde{\lambda}\cos(u_\theta(t_i))\}_{i=1}^N$, we in fact obtain uncertainty in $\phi(u)$ as well, which is leveraged in the following symbolic regression. As shown in Fig.~\ref{fig:appendix_2}, for noisy-data cases, we also perform symbolic regression on the upper and lower bounds of the uncertainty band in predicting $\phi(u)$. Although the discovered mathematical expressions in Table~\ref{tab:example_1_3} from noisy-data cases are not as accurate, the error in function values is bounded by the upper and lower bounds. Besides, due to the increase of data, the predicted uncertainty from the noisy-and-sufficient-data case is significantly smaller.

\section{Details for the computations in lattice Boltzman equation model}\label{subsec:lbe}

In this section, we provide details for generating data of the non-Newtonian Cavity flow in Sec.~\ref{subsec:cavity_flow} using the lattice Boltzman equation model (LBE) in \cite{wang2015localized}. We set Re $= U^{2-n}L/\nu_0 = 100, L=1, U=0.01, \rho=1, n=1.5$ and use a uniform mesh of size $101\times101$ on $[0, 1]^2$  to perform the simulations.

\end{document}